\documentclass[11pt]{article}
\usepackage[utf8]{inputenc}
\usepackage[T1]{fontenc}
\usepackage{amsmath,amssymb}
\usepackage{booktabs}
\usepackage{array}
\usepackage{tikz}
\usetikzlibrary{arrows.meta,positioning,fit}
\usepackage[hyphens]{url}
\usepackage[hidelinks]{hyperref}

\usepackage[margin=1in]{geometry}
\usepackage{mathptmx}
\DeclareMathAlphabet{\mathcal}{OMS}{cmsy}{m}{n}
\usepackage{longtable}
\newenvironment{reflist}{\begin{list}{}{\setlength{\leftmargin}{1.5em}\setlength{\itemindent}{-1.5em}\setlength{\parsep}{2pt}\setlength{\itemsep}{2pt}}}{\end{list}}

\title{Constitutive Priors for Machine Intelligence: A Legitimacy Theory of the Artificial Physical World}
\author{Jiang Jiang\thanks{Corresponding author: jiangjiang@persagy.com} , Yifu Sun, Qi Shen\\[2pt]
\small Persagy Science and Technology Co., Beijing, China\\
\small jiangjiang@persagy.com, sunyifu@persagy.com, shenqi@persagy.com}
\date{August 2026}

\begin{document}
\maketitle

\begin{abstract}
Machine intelligence's push into the physical world is stuck on a gap: deployment scenarios demand auditable judgments from the first day of operation, fault samples are scarce or absent, and the norms that define "what counts as a fault" live in design documents, not in operational data. This paper argues that the gap is structural, and locates where it can be legitimately closed. We divide the worlds machine intelligence faces into four (\allowbreak{}phenomenal, basic physical, artificial physical, artificial symbolic) along a single axis of constraint strength, and we give the Promulgation Criterion: extracting a prior framework from a world is legitimate if and only if the world is intentionally constituted (C1) and has left a readable generative archive (C2). On the criterion's own two gradient axes, the corner that is high on both holds exactly one world: the artificial physical world of buildings, factories, and infrastructure, whose norms are promulgated before their instances; the legitimate path to machine intelligence for this world is to extract the framework from its constitutive archive, not to induce it from data. We then characterize the shape such a framework must take: four construction goals (stability, openness, abstraction, concreteness) force four mutually incompatible carriers, so a legitimate prior framework must divide into at least four layers --- syntax, concepts, knowledge, instances (\allowbreak{}Structural Characterization 5.1); on a closed concept layer, fault localization is decidable in polynomial time (\allowbreak{}Proposition 5.2), and every judgment becomes interrogable, traceable clause by clause to a promulgated norm. The same criterion fixes the division of labor with large language models at runtime: promulgatable duties go to rule engines, on-site judgments beyond promulgation go to LLMs, and every generation sandwiched by promulgated clauses is auditable. The theory is falsifiable: we put four bets on the table (P1--P4) --- among them, that the next large-scale AI breakthrough occurs in the artificial physical world, ahead of the basic physical and phenomenal worlds --- plus two structural corollaries, each with explicit falsification conditions and named components at stake. The paper's theses are supported by three kinds of evidence: formal proofs (Appendix A, in the supplementary material); two constructive cases (Appendix B: an operating cooling plant, and the Curiosity rover's Sol 1536 anomaly); and four-layer reverse readings of eight independently evolved lineages, engineering and mathematical, from BACnet to RDF/OWL (Appendix C).

\end{abstract}


\section{Chapter 1. Introduction}
\subsection{1.1 A judgment made without posterior data}
On the morning of July 15, 2026, a fault occurred in the cooling plant of a data center. At 08:06, the outlet flow of chilled-water pump \#1 was still within its normal range. At 08:11, the ratio of flow to rated value fell below 0.8; at 08:12, with the crossing sustained for more than 60 s, the monitoring system raised an alarm. At 08:14, the reasoning engine judged the state as "flow loss \textperiodcentered{} critical" and immediately began cause-tracing along a causal graph: three candidates were ruled out first, then the chain "grease degradation $\to$ bearing wear $\to$ impeller efficiency drop $\to$ flow loss" was hit. At 08:16, the system-level conclusion "cooling function of the central cooling system failed" was activated together with the corresponding handling procedure. The bearing was replaced by hand that morning; by 12:08, all alarm states had returned to zero. (This example is a compressed narration of the demonstration scenario constructed in Appendix B.1 (\allowbreak{}supplementary material) --- the knowledge-base artifact is real, the fault event is staged; for all reasoning steps, the knowledge entries they rest on, and the boundary statements, see that section.)

This operations story is unremarkable, except for one fact: the system that made all the judgments above had never seen any historical fault data of any pump. None of its steps was fitted from data; each was made against an archive. The threshold 0.8 and the 60 s debounce time were promulgated; the nodes and edges of the causal graph were promulgated; the handling procedures were promulgated; even the definition of the quantity "ratio of flow to rated value" was itself promulgated. Data were responsible only for telling the system what was happening now; the answers to "does this count as a fault, of which category, and what shall be done" had been written into the archive before the system was ever switched on.

In the terms of diagnosis theory, the causal graph and vocabulary on which this cause-tracing relied are exactly a system description (SD). Since Reiter (1987), the MBD tradition has run its diagnostic computation with the SD as a given premise; its internal inquiries into the SD concentrate on engineering provenance (how models are composed, abstracted, and reused (\allowbreak{}Falkenhainer \& Forbus, 1991; Struss, 1992)) while the legitimacy provenance remains presupposed and unelaborated: by what right does an SD constrain instances prior to instances, and why is it promulgated rather than fitted? This paper answers the latter (Section 2.4, R.8; Chapter 4).

Such an archive is what we call a \emph{prior framework}; the way its entries hold, we call \emph{promulgation}: issued intentionally by an authorized body, prior to instances, prescribing how instances shall be. Standards, drawings, and operating procedures are promulgation's engineering forms; this paper's work is to build its theoretical form.

\subsection{1.2 The problem: a gap prior to posterior data}
The basic paradigm of machine learning is posterior: first induce regularities from data, then put them to use. Chapter 2 rereads the timeline of that victory. But more and more deployment scenarios sit on its opposite side: equipment must deliver auditable judgments from its first day of operation; fault samples are scarce or even zero; and the answer to "what counts as a fault" is not hidden in the data --- it exists in design documents, operating procedures, and industry standards, prior to any observation. Judgment must precede data, yet data cannot supply the norms on which judgment rests. We call this situation the \emph{cold-start deadlock}.

This gap has long gone untreated as an independent object of research. Our diagnosis: a tool for distinguishing learning-task spaces was missing. Chapter 3 provides such a tool: a one-dimensional framework on a "constraint--openness" axis, dividing the worlds machine intelligence faces into four: the basic physical world, the phenomenal world, the artificial symbolic world, and the subject of this paper, the \emph{artificial physical world}: a physical world designed by humans, whose norms are promulgated prior to instances, and whose instances owe their existence and identity to compliance with those norms. The cold-start deadlock is this world's home ground.

\subsection{1.3 Contributions of this paper}
This paper establishes a legitimacy theory of constitutive priors for machine intelligence in the artificial physical world. Five contributions:

\begin{enumerate}
\item \textbf{The criterion (Chapter 4).} We give the Promulgation Criterion for constitutive priors: the norms must be promulgated (C1), and the constitution has left a readable generative archive (C2). The criterion also yields a two-way boundary: when the criterion is not satisfied, the problem is referred to model defect rather than world violation.
\item \textbf{The architecture and its lower bound (Chapter 5).} We give the four-layer carrier structure of a prior framework (syntax, concepts, knowledge, instances) and two structural results (\allowbreak{}Structural Characterization 5.1, Proposition 5.2; the normative/logical proportions of their proof loads are honestly registered after each): four layers are a lower bound rather than a design preference, and fault reduction on this structure is decidable. From this follows the property Chapter 9 closes on: a system built on this framework is interrogable --- every judgment can be traced up the promulgation chain into a reason (Corollary 5.1).
\item \textbf{The division-of-labor interface with large language models (Chapter 6).} Prior frameworks and large language models are not competitors: the framework is the semantic anchorage, providing a constitutive, instance-bound landing ground for what the model reads in; runtime duties are divided by the Promulgation Criterion.
\item \textbf{Falsifiable predictions (Chapter 7).} All empirical premises of this paper are organized into four falsifiable bets and two structural corollaries, among them: the machine-intelligence breakthrough in the artificial physical world will precede those in the basic physical world and the phenomenal world; and the prior framework retains its advantage over purely posterior routes in domains with readable archives.
\item \textbf{The dialogue with the MBD tradition (Section 2.4, R.8; Section 5.5).} We give an account of the legitimacy provenance of the system description (SD) (the MBD tradition has elaborated its engineering provenance; the legitimacy provenance remains presupposed and unelaborated --- Section 2.4, R.8) and a complexity divergence: reduction over a promulgated closed concept layer is $O(d\cdot|C|)$ (\allowbreak{}Proposition 5.2), as distinct from the NP-hardness of minimal-hitting-set diagnosis.
\end{enumerate}
\subsection{1.4 Claim types and evidence policy}
This paper's claims are structural propositions and impossibility propositions, not performance propositions. Correspondingly, its evidence set consists of three kinds, one per appendix: the formal proofs of Appendix A (\allowbreak{}supplementary material); the two constructive cases of Appendix B, demonstrating that the framework can actually be built and run end to end (the existence side); and the eight external witnesses of Appendix C (BACnet, LonWorks, OPC, Modbus, autonomous-driving ODD, COIN, Brick/Haystack, RDF/OWL), seven engineering lineages evolved independently over decades plus one purely mathematical construction, inventoried at their landing points when reverse-read through this paper's framework (the natural-experiment side). Comparative experiments are not in this paper's evidence set. This is no omission; it is determined by the claim type: for propositions about "how systems of this kind shall be constructed and what their structural lower bound is," no control benchmark exists on a laboratory bench. The discipline this paper adopts is: premises may be judgment; inference must be logic. All premises must therefore be laid out in the open and be falsifiable --- this is why the stake structure of Chapter 7 exists. Correspondingly, this paper does not sell itself on the number or difficulty of formal results: The discipline of Appendix A (\allowbreak{}supplementary material) is to honestly register the load-bearing distribution of each result, rather than to package normative premises as mathematical theorems. A notation table is provided at the head of Appendix A (\allowbreak{}supplementary material).

\subsection{1.5 The structure of this paper}
Chapter 2 rereads the history of AI from the learner's perspective and settles accounts with the nearest traditions in the form of related work. Chapter 3 gives the four-world partition tool and the formal definition of the artificial physical world. Chapter 4 establishes the Promulgation Criterion. Chapter 5 gives the four-layer architecture and the lower-bound argument. Chapter 6 discusses the division of labor with large language models. Chapter 7 registers the falsifiable predictions and the stake structure. Chapter 8 replies to five objections, one by one. Chapter 9 concludes by returning to the theme of interrogable intelligence. Appendices A, B, and C carry the three kinds of evidence (proofs, constructions, and natural experiments) respectively.

\subsection{1.6 Boundary statement}
This paper proposes no route toward general intelligence. The four-world framework is an analytical tool motivated by human individual cognitive development; it promises neither completeness nor sufficiency. The paper answers exactly one question: in what form the constitutive priors needed by the artificial physical world exist, and by what right they are legitimate. Chapter 9 closes on a plainer claim: where norms precede data, the trustworthiness of intelligence comes from the interrogability of the archive, not the scale of the model.

\section{Chapter 2. Learning and Promulgation: A Rereading of AI History and Related Work}
\begin{quote}
This chapter does two things. The first three sections reread the history of AI from the learner's perspective: the rule era, the statistical era, and the present predicament each answer one question --- why handwritten rules failed, what the order of posterior victories leaks, and where the push into the physical world is stuck. Section 2.4 is the related work, settling boundaries with the neighbors that must be delineated, one by one. All attributions in this chapter are first registered in timeline form; the diagnosis in criterion form is given in Chapter 4.
\end{quote}
\subsection{2.1 The rule era: a misread failure}
Rule-based NLP tried to write grammar down as rules first, and then let machines parse and generate by the rules (Chomsky, 1957). The engineering bottleneck was not the inference engine; it was that the rules kept multiplying while the exceptions grew faster, and maintenance costs eventually exceeded the returns. Expert systems met another form of the same bottleneck: the rules an expert can articulate are far fewer than the knowledge the expert actually uses in judgment --- the knowledge-acquisition bottleneck (\allowbreak{}Feigenbaum, 1977). Cyc took this road to its farthest point: roughly a person-century of investment and about one million hand-entered axioms (Lenat, 1995). What it proved was not that "common sense cannot be written," but that the \emph{writing} of common sense has no termination condition.

The common attribution of these failures is neither "the rules were wrong" nor a blanket "priors failed." It is this: \emph{the object world contains no design rules prior to instances}. Language has no promulgating authority, and common sense has no design documents; what the rule writer can rely on is always only after-the-fact observation of the world, so handwritten rules are descriptions of descriptions. The same attribution replays in the opposite direction in protein-structure prediction (Section 4.3): the evolutionary archive happens to be readable, but what it records is "what survived," not any design intention. The lesson of the rule era was accordingly summarized by the statistical era as "priors are useless." This paper reads it differently. Handwriting content rules in a world without design rules is overreach; failing to use them in a world where design rules exist is waste. What distinguishes the two worlds is the criterion given in Chapter 4.

\subsection{2.2 The statistical era: the hidden message behind the posterior victories}
The statistical era was the across-the-board victory of posterior methods: deep learning handwrites no content rules and lets everything be learned from data. Behind this victory hides a piece of information that matters greatly to this paper --- the order in which the victories occurred.

Counted by large-scale commercial deployment, the order is: board games first (Silver et al., 2016), then code and natural-language systems; image generation has to date achieved only shadow-level success in the phenomenal direction (\allowbreak{}generation that samples surface statistics without causal mastery), and the cognition of and interaction with the real world still lags. The coordinate-system explanation of this order, Moravec's paradox (Moravec, 1988) included, and the literature registration are given in Section 3.4.

\subsection{2.3 The present: three cracks and one temptation}
After the statistical era's success in the symbolic world, three cracks have appeared in machine intelligence's push into the physical world.

\textbf{Symbol grounding remains unsolved.} The "meaning" of a large language model is a network of relations among symbols, not a relation between symbols and the world (Harnad, 1990): it can talk about water but has never interacted with water; its physical common sense is inherited from the compression embodied in human language --- second-hand.

\textbf{Physical data are scarce.} Physical instance data are expensive, scarce, and mostly private; there is no free corpus of plumbing, HVAC, and electrical systems, and failure data are especially rare --- the cold-start deadlock of Section 1. The zones where data are poorest are precisely the asset-intensive facilities where value density is highest.

\textbf{The world models' disorientation.} The two main routes pushing into the physical world (physical simulation, such as MuJoCo (Todorov et al., 2012) and NVIDIA Isaac; and learned world models, such as World Models (Ha \& Schmidhuber, 2018), Genie (Bruce et al., 2024), and NVIDIA Cosmos) aim at the same target: learning predictive representations of world dynamics from video and interaction data. The direction dispute of this route has already gone public within the field (e.g., LeCun's criticism of the pixel-level generation route and his JEPA proposal of predicting in representation space (LeCun, 2022; Assran et al., 2023)). That "what world models should learn, and how" cannot reach consensus is certainly a disorientation; but this paper holds that the deeper cause of the disorientation lies beyond the architecture dispute: world models learn how states evolve, whereas deployment scenarios require answering something else --- which states count as normal, which count as violations, and what defines the correct state.

Alongside the three cracks stands a real industry temptation: large language models have finished reading the knowledge of the human symbolic world, world models are acquiring the ability to interact with the real world, and the two routes seem about to converge at the summit, closing the dam of AGI. Yet the three cracks tell us clearly that this convergence is not so simple. We cannot answer "how the dam will close," but Chapters 3, 4, and 5 will answer the question the three cracks jointly point to: the promulgated-norm gap: its form of existence and its conditions of legitimacy. (The strongest form of this temptation, and the full reply, are in Chapter 8, O1; the runtime division of labor between prior frameworks and LLMs, i.e., large language models, is given in Chapter 6.)

In 2026, two industry voices issued judgments in the same direction: Su Hao at WAIC 2026 proposed a six-level ladder of physical knowledge (the lower three levels belong to the objective world; the upper three "come into being because of me"), and pointed out that "language is only a projection of the world," hallucination stemming from knowledge having no anchor in reality (Su, 2026); Richard Sutton judges that AI is moving from the "era of human data" into the "era of experience": large models have no reward signal and cannot tell the true and good from the false and bad among their own behaviors (Silver \& Sutton, 2025; Sutton, 2026). Their arguments differ; their judgment is the same: convergence is not simple.

\subsection{2.4 Boundaries and comparisons}
\textbf{R.1 Knowledge representation and description logic.} The KR tradition has provided an institutional home for explicit semantic structure for fifty years, but its founding definition stays neutral on the epistemic status of specifications: Gruber's "specification of a conceptualization" (Gruber, 1993) admits at one price both "laws promulgated to the world" and "summaries extracted from the world." This paper's criterion is the question pressed down by that neutrality: whether the object domain has left a readable archive that precedes and constitutes instances. Description logic provides the closest formal counterpart to the four-layer skeleton:

\begin{table}[htbp]
\centering
\small
\begin{tabular}{@{}>{\raggedright\arraybackslash}p{0.473\textwidth}>{\raggedright\arraybackslash}p{0.473\textwidth}@{}}
\toprule
This paper & The DL tradition \\
\midrule
syntax layer & signature (\allowbreak{}vocabulary) \\
concept layer & TBox (\allowbreak{}terminological axioms) \\
knowledge layer & rule base (rule extensions) \\
instance layer & ABox, extended with streaming data \\
\bottomrule
\end{tabular}
\vspace{3pt}
\noindent{\small\textbf{Table 2-1.} The four-layer skeleton set against description logic.\par}
\end{table}
In the DL literature, the TBox/ABox separation is a convention of representation practice rather than a theorem, and the choice of expressive power correlates with reasoning complexity (Baader et al., 2003; W3C, 2012). This paper's layering lower bound (\allowbreak{}Structural Characterization 5.1) argues: for a framework to achieve the two goal pairs (stability with openness, abstraction with concreteness) the crossing of the two axes produces four carriers (stable $\times$ abstract, open $\times$ abstract, stable $\times$ concrete, open $\times$ concrete), each residing in its own layer, and the four separations are therefore a lower bound rather than a design preference. The two bodies of work are complementary: description logic answers how expressive power affects decidability; this paper answers why a framework that "both grows and keeps" needs at least four separations. Knowledge graphs are the industrial descendants of this tradition: they sediment the concept layer and the instance layer, but their content is extracted rather than promulgated, and the knowledge-layer norms remain absent throughout (Table 2-2, Section 5.6).

\textbf{R.2 Normative systems.} Deontic logic first asked about the logical form of "ought" (von Wright, 1951); normative multi-agent systems engineered it: norms have a life cycle of creation, promulgation, observance, violation, and sanction, and obligations are promulgated by legislators rather than averaged from behavior (Esteva et al., 2001; Dastani, 2008; Andrighetto et al., 2013). One terminological overlap needs clarification: Boella and van der Torre distinguish regulative from constitutive norms (Boella \& van der Torre, 2004) --- the same word as this paper's, from the same Searlean lineage, differing in object: their constitutive norms underwrite institutional facts in agent societies, while this paper's constitutive norms delimit the normal states of the artificial physical world. Structural isomorphism, different objects. The two traditions are complementary: normative MAS presupposes the promulgating authority, and this paper's criterion answers when that presupposition holds; their formal machinery (deontic operators, norm life cycles, enforcement models) in turn provides this paper's constitutive norms with a ready-made formal language.

\textbf{R.3 Bayesian priors.} A Bayesian prior is a quantitative belief (Bernardo \& Smith, 1994): its content is a distribution, its source is the modeler, and when data conflict with the prior, the prior is updated. A constitutive prior is a promulgated norm: its content is types, rules, and clauses, its source is the design archive, and when an instance deviates from a norm, the fault lies with the instance. The two fail differently: a misspecified Bayesian prior is diluted by the likelihood and silently degrades, while a violated constitutive norm raises an explicit alarm at a named clause. Formally, the two do not live in the same mathematical space: one in the space of probability measures, the other in the model-theoretic satisfaction relation ($\models$). This is the formal echo of the is--ought gap: a distribution over "what happens" cannot express "what ought to happen." The instance layer may legitimately hold subjective priors over quantities the archive does not prescribe, but no amount of Bayesian updating can turn a distribution into a norm.

\textbf{R.4 Neurosymbolic AI.} This community has long held that statistical learning alone is insufficient for robust and interpretable intelligence (d'Avila Garcez \& Lamb, 2023), and its practical gains (e.g., sample efficiency in concept learning and differentiable logical constraints (Mao et al., 2019; Badreddine et al., 2022)) are indirect evidence that prior structure works. The division of labor with this paper is clear: neurosymbolic research answers the architecture question (how symbols and networks couple); this paper answers the question one step earlier (where symbols come from and by what right they constrain networks). That tradition lacks a defense of the epistemic status of its symbolic layer; this paper's criterion and layering lower bound (\allowbreak{}Structural Characterization 5.1) supply that defense.

\textbf{R.5 Agent scaffolding.} The engineering evolution of the LLM industry has gradually filled in human-promulgated components: system prompts, tool specifications (Yao et al., 2023; Schick et al., 2023), persistent memory (Packer et al., 2023), multi-agent orchestration rules (Wu et al., 2023). These components are not descriptions distilled from model behavior; they are norms written down to constrain model behavior. They are written-to-be-read, and they occupy an extreme cell on the archive-readability gradient --- among the very few documents whose first reader is a machine (Section 4.3). By this paper's criterion, scaffolding is prior; the posterior tradition's own engineering practice thereby constitutes practical evidence for the necessity of promulgation. But what scaffolding promulgates is only conventions over model behavior, and it does not answer three things: where the norms come from and by what right they are legitimate (Chapter 4); into how many layers the promulgated content should be divided (Chapter 5); and how to anchor instances of the physical world (Appendix B, supplementary material). These three are this paper's additions relative to scaffolding.

\textbf{R.6 Physics-informed neural networks (PINNs).} PINNs (Raissi et al., 2019) are the closest neighbor to this paper in surface features (explicit structure, distrust of pure black boxes, respect for the domain's laws) and therefore the neighbor that must be separated with precision. PINNs and this paper both advocate explicit structure; the entire difference lies in one variable: their explicit content is descriptive (laws extracted from observations) while this paper's is constitutive --- norms promulgated as to how the world shall be maintained. Physical laws describe how the world runs; they do not define what the world shall be maintained as: "the supply-return water temperature difference shall remain stable" is not a subset of any physical law. PINNs' norms have no failure semantics and stay silent in the face of deviation. The relation is complementary rather than competitive: the physics engine supplies constraint boundaries (what is physically impossible); this paper's prior framework supplies normative coordinates (what counts as normal and what counts as a fault).

\textbf{R.7 Existing layer-count claims.} On the number of layers, five notable positions are documented in the literature, each with its own provenance. \textbf{MOF} (OMG, 2016): its four layers are uniquely determined by the closure of instantiation regression. \textbf{Marr's three levels} (Marr, 1982) are epistemological (what is needed to understand an existing system), whereas this paper's four layers are ontological (what is needed to promulgate and maintain a normative world). \textbf{Palantir's Ontology layering} is determined by its construction goals, and its per-customer forward-deployment cost is exactly the price of the criterion being half-satisfied (Section 4.3). \textbf{Rasmussen's abstraction hierarchy} (Rasmussen, 1985) answers how an operator understands an existing plant. \textbf{Su Hao's six-level ladder} (Su, 2026; already described in Section 2.3) answers what a learner must know. The four layers of this framework answer another question: where the framework's content resides and under what disciplines it is revised. Each position answers its own question; only this paper's "four" grows out of the definition of promulgation.

\textbf{R.8 Model-based diagnosis (MBD/DX).} MBD is the founding tradition of fault diagnosis in AI: Reiter gave the theory of diagnosis from first principles --- given a system description SD, component behavior models, and observations, a diagnosis is a minimal hitting set of minimal conflict sets (Reiter, 1987); GDE gave the engineered implementation for multiple faults (de Kleer \& Williams, 1987); for systematization see (de Kleer et al., 1992), and for a genealogy of the logical definitions see (Console \& Torasso, 1991). The relation to MBD has three layers. \textbf{First, the questions differ}: MBD answers "given a system description, how to localize faults logically"; this paper asks the question one step earlier: where the legitimacy of that system description comes from: by what right it constrains instances prior to instances, and why it is promulgated rather than fitted. The DX tradition's inquiries into the provenance of SD concentrate on the engineering side: compositional modeling and model libraries (\allowbreak{}Falkenhainer \& Forbus, 1991; Struss, 1992) and structured system descriptions (Darwiche, 1998) answer "how an SD is built, abstracted, and reused at low cost"; but the legitimacy provenance of SD's priority and authority remains presupposed and unelaborated, and this paper's criterion (\allowbreak{}Proposition 4.1) answers that layer. \textbf{Second, the structures are adjacent but the problem scales differ}: this paper's knowledge layer corresponds to the failure-mode side of behavior descriptions in an SD, and the reduction of Algorithm 1 corresponds to diagnostic computation; but the two complexity figures are not directly comparable. MBD's diagnostic space is the full combinatorics of component-fault hypotheses (minimal hitting sets are NP-hard (de Kleer et al., 1992)), whereas this paper's reduction is a level-by-level scan over a promulgated type list, $O(d\cdot|C|)$ (\allowbreak{}Proposition 5.2), with unanticipated fault combinations absorbed by the augmentation passage of Section 5.2 rather than by the reduction algorithm. The divergence lies in who supplies the diagnostic space, not in algorithmic speed. \textbf{Third, complementary rather than competitive}: MBD's inference engines can live inside this paper's knowledge layer; what this paper seeks to supply is the other question: the legitimacy provenance and maintenance discipline of the SD. The knowledge-acquisition bottleneck of handwritten SDs is a self-acknowledged pain point of the DX literature (of the same form as the expert systems of Section 2.1), and its engineering value has in-orbit confirmation: Remote Agent's Livingstone component autonomously performed model-based diagnosis and recovery on Deep Space One (\allowbreak{}Muscettola et al., 1998). A handwritten SD can fly; the question lies only in its provenance and its maintenance discipline. One final registration against misreading: this paper's "fault reduction is decidable" does not conflict with MBD's complexity results: the decidability comes from the promulgated antecedent of a closed concept layer; where that antecedent is absent (\allowbreak{}archiveless worlds), the boundary of Proposition A.13 applies to MBD's model acquisition as well.

\textbf{R.9 Closing.} The communities and routes above can be uniformly measured by this paper's criterion. Several mutually unaffiliated communities (knowledge representation, normative systems, neurosymbolic AI, industrial modeling, the diagnosis tradition, the LLM industry) driven by different pain points, have independently converged on the same structure: an explicit, promulgated, layered semantic layer; and one independent mathematical construction, COIN, also lands on the same four layers, inventoried in Appendix C.6 (\allowbreak{}supplementary material). The control group (R.6) isolates the decisive variable: what matters is not explicitness but promulgation. One same-name anchor is registered last: Newell's knowledge level (Newell, 1982) is a level in the descriptive hierarchy of intelligent systems, situated above the symbol level, and its abstraction deliberately leaves untouched the carriers in which knowledge resides and the disciplines by which it is maintained; this paper's "knowledge layer" in Chapter 5 shares the name but not the thing --- it is one of the four carriers, and it answers exactly the question Newell set aside: once knowledge is attributed to a system, on what carrier does it reside, and under what discipline is it revised? The two contemporary main routes in Table 2-2 (world models and large language models) are not given separate items in this section: the former was registered as the disorientation in Section 2.3, and for the latter the encyclopedia objection is developed in Chapter 8, O1, and the runtime division of labor is given in Chapter 6.

\begin{table}[htbp]
\centering
\small
\begin{tabular}{@{}>{\raggedright\arraybackslash}p{0.173\textwidth}>{\raggedright\arraybackslash}p{0.173\textwidth}>{\raggedright\arraybackslash}p{0.173\textwidth}>{\raggedright\arraybackslash}p{0.173\textwidth}>{\raggedright\arraybackslash}p{0.173\textwidth}@{}}
\toprule
Route & Data source & Direction of fit & Failure semantics & Zero-shot operable \\
\midrule
Deep learning & instances & description & silent drift & no \\
World models & video, interaction & description (\allowbreak{}prediction) & silent & no \\
Large language models & human symbolic corpora & description & no promulgated clause to check against (\allowbreak{}hallucination) & "yes" within the symbolic domain \\
PINNs & observations + laws & description (explicit) & prediction error, not violation & no \\
Knowledge graphs & extracted concepts & description & none & partial \\
MBD/DX & system description (\allowbreak{}promulgated by the modeler) & promulgation (the promulgator is the modeler, not the world's design archive) & conflict sets & yes \\
Rule era (Cyc, expert systems) & handwritten rules & promulgation, wrong domain & loud but brittle & claims "yes"; overreach where no design rules exist \\
\textbf{Constitutive priors (this paper)} & \textbf{design archives} & \textbf{promulgation} & \textbf{loud violation} & \textbf{yes} \\
\bottomrule
\end{tabular}
\vspace{3pt}
\noindent{\small\textbf{Table 2-2.} The routes toward the physical world, measured with the criterion as the instrument.\par}
\end{table}
\subsection{Chapter summary}
This chapter has cleared the ground for the chapters that follow, from the sides of history and related work. Historically, the rule era's failure is attributed to handwriting content rules in a world without design rules; behind the statistical era's sweeping victory hides one piece of information --- the order of the breakthroughs, which forms the ordering clue of Chapter 3; and the three cracks in the present push toward the physical world point to one and the same gap: the place of promulgated norms. In related work, several mutually unaffiliated communities and one independent mathematical construction converge on promulgation-based semantic layers, and the criterion measures the differences among all routes with a single instrument (Table 2-2). The following chapters give, in order: a coordinate system for learning tasks (Chapter 3), the criterion (Chapter 4), and the necessary structure of the framework (Chapter 5).

\section{Chapter 3. Four Worlds: An Analytical Framework for Learning Tasks}
\subsection{3.1 An analytical framework}
This chapter introduces an analytical framework that characterizes systematic differences among object worlds in their \emph{learning conditions}. Two things must be fixed first: how the framework poses its question, and where its claims stop.

The framework does not answer the metaphysical question of what the world itself is made of. It answers an instrumental one: \emph{for the learning tasks faced by artificial intelligence systems, how should object worlds be partitioned on the basis of differences in learning conditions?} This way of posing the question has a clear methodological lineage. Uexküll's notion of Umwelt shows that each species' surrounding world is carved out by its sensory and motor apparatus, so that the partitioning of the world depends on the observer's equipment and tasks (Uexküll, 1934). Dennett's theory of stances shows that the properties an object presents switch with the stance the observer adopts (Dennett, 1987). The ontology-engineering tradition of Gruber and Guarino treats a conceptualization as an engineered artifact in the service of a particular task, its value adjudicated by the task it serves (Gruber, 1993; Guarino, 1998). We inherit that instrumental stance: the legitimacy of the world partition proposed below is decided not by metaphysics but by its explanatory and predictive power over learning problems.

Concretely, the framework is a one-dimensional analytical tool built on a single axis of constraint strength. Three qualitative dimensions first characterize the learning conditions of each world (Section 3.3); they then converge into an ordering axis along which object worlds range from closed to open (Section 3.4). The one-dimensionality is a deliberate simplification: differences among object worlds certainly exist that this axis cannot capture (Section 3.8 gives an important instance). We do not claim the framework is complete (it does not promise to exhaust the space of learning tasks) nor that it is sufficient --- it does not promise that the intelligences of the four worlds add up to general intelligence. The claim of the framework is functional: it \emph{explains} the historical order of AI capability breakthroughs (Section 3.4) and yields a \emph{falsifiable prediction} (end of Section 3.4; P1 in Chapter 7). If someone proposes a fifth, identifiable world, the criterion of Chapter 4 applies to it as well; that would be an extension of the framework, not a falsification of it.

\subsection{3.2 The four worlds: working definitions}
Guided by the sequence of human individual cognitive development, we introduce four object worlds as our analytical units. We give working definitions first; Section 3.3 supplies the three dimensions that characterize their differences, and Section 3.5 gives the precise definition of the world that matters most here.

\begin{itemize}
\item \textbf{Phenomenal world}: the continuous stream of sensorily accessible phenomena --- cold and warm, soft and hard, light and shadow, sound. It presents itself as continuous, high-dimensional sensory signals with no ready-made discrete units.
\item \textbf{Basic physical world}: the world of moving, colliding, and supporting objects, of fluids and energy: the mechanistic layer governed by physical law. Its regularities are not directly presented to the senses; they are revealed through instruments and experiments in interaction with nature.
\item \textbf{Artificial physical world}: designed and built physical artifacts and their operation: buildings, machines, facilities, infrastructure. It exists in two forms at once: physical instances and design documents.
\item \textbf{Artificial symbolic world}: discrete symbol systems produced for human reading: writing, books, programs, data records.
\end{itemize}
The four worlds are \emph{motivated} by the ontogenetic sequence. An infant first grasps sensory phenomena directly; it then constructs physical schemas through the interplay of action and symbol, forming expectations of object permanence and intuitive physics within months (Piaget, 1954; Spelke \& Kinzler, 2007); around the toddler stage it develops a teleological understanding of artifacts, treating them as things "made for a purpose" (Kelemen, 2004); and it finally acquires symbol systems (Deacon, 1997). One point deserves emphasis: the developmental sequence supplies the framework's motivation and its candidate units, not its derivation. The developmental order of the human individual is a given of human biology; nothing follows from it about whether a machine learner's task space falls into exactly four classes. The framework's warrant is ultimately adjudicated by the explanation and prediction of Section 3.4, not guaranteed by its provenance.

\subsection{3.3 Three characterizing dimensions}
The differences among the four worlds in learning conditions are characterized along three dimensions. We state up front: the three dimensions are for \emph{qualitative description} and do not constitute a metric.

\textbf{Dimension 1: data interface.} In what form, and at what cost, does the world present data to the learner --- a continuous sensory stream or discrete symbols, written-to-be-read or to be extracted by the learner itself?

\textbf{Dimension 2: constraint structure.} How compressible are the regularities the learner actually faces? This is measured from the learner's accessible standpoint: we ask not about the nomological strength of the world's laws in themselves, but about the compressibility of regularities within the data available to the learner.

\textbf{Dimension 3: rule source.} What is the generative relation between rules and instances? This dimension distinguishes two situations: (i) rules are \emph{promulgated} prior to instances, constituting and regulating them; (ii) rules are \emph{induced} posterior to instances, describing them. Chapter 4 refines this dimension into the two conjuncts of the legitimacy criterion; the present chapter only registers whether design rules prior to instances exist. The conditions under which such rules legitimate the extraction of a prior framework are answered in Chapter 4.

The valuations of the three dimensions for the four worlds are as follows:

\begin{table}[htbp]
\centering
\small
\begin{tabular}{@{}>{\raggedright\arraybackslash}p{0.173\textwidth}>{\raggedright\arraybackslash}p{0.173\textwidth}>{\raggedright\arraybackslash}p{0.173\textwidth}>{\raggedright\arraybackslash}p{0.173\textwidth}>{\raggedright\arraybackslash}p{0.173\textwidth}@{}}
\toprule
World & Examples & Data interface & Constraint structure & Rule source \\
\midrule
Phenomenal world & cold weather, soft cloth, a cool breeze & continuous high-dimensional sensory stream; no ready-made segmentation & weakest (open) & none \\
Basic physical world & pushing a box, tiring when lifting & obtained in interaction with nature & the laws themselves are highly compressible, but the behavior of composite systems is effectively open to the learner & none; laws are discovered descriptions, not promulgated rules \\
Artificial physical world & lamps, air conditioners, buildings, machines & dual channel: physical instances plus design archives & strong (dual constraint of physics and intention) & present (rules promulgated in design archives) \\
Artificial symbolic world & writing, books, programs & discrete symbols, written-to-be-read & strongest (closed) & split: code precedes its instances; language follows its instances (see Section 3.7) \\
\bottomrule
\end{tabular}
\end{table}
Two remarks. First, on the classification of organisms. A reader may ask why life is not listed as a world of its own, since the genome, too, is a generative record that precedes the individual. The reason is that evolution has no intention: the genome fails condition (ii) of Definition 3.1 (\allowbreak{}intentional constitution), so organisms fall under the basic physical world. The practical consequence of this boundary (a readable generative trace is not an extractable norm) is developed in the protein case of Section 4.3. Second, the four worlds are not four equivalent perspectives on a single intelligence. Their differing learning conditions mean different learning problems; the full argument for this point is given for the artificial physical world in Section 3.6.

\subsection{3.4 The constraint axis: explaining history and predicting}
Ordering the four worlds by constraint strength yields a continuous axis: \emph{the further toward the artificial-symbolic end, the stronger the constraint and the more bounded the semantic space; the further toward the phenomenal end, the closer to an open world.} We call it the constraint axis.

The constraint axis agrees with the historical order of AI capability breakthroughs. Board games, whose rules are fully closed, fell first (Silver et al., 2016). In the large-model era, code systems entered large-scale commercial deployment earliest (Chen et al., 2021; OpenAI, 2023): their training corpora are pre-governed by programming-language norms (code that fails syntax and type checks does not enter the repository) and correctness can be mechanically certified by compilers and tests. Natural-language systems then reached large-scale commercial use as well. Latent-space diffusion models drove rapid progress in high-resolution image and video generation (Rombach et al., 2022); yet it must be noted that generative success samples the surface statistics of the phenomenal world rather than mastering its causal structure. Along the phenomenal direction, only such shadow-level success has been achieved to date, and understanding and manipulation tasks in open physical environments (robotics being the representative case) still lag. By and large, breakthroughs have proceeded from stronger to weaker constraint. The success of natural language has one further, often overlooked condition: linguistic data are written-to-be-read. Humans have preprocessed language into a discrete, standardized, low-noise symbol stream; the physical world offers no such free data interface. Moravec's paradox (Moravec, 1988) is explained in the same coordinate system: sensorimotor skill sits at the phenomenal end, where the semantic space is most open and no data interface exists. Human ability there is precompiled by evolution and therefore looks easy; AI must face the largest learning space head-on.

This agreement is not the paper's main claim; it is a testable corollary of the framework. It directly yields a prediction: \emph{the next large-scale AI breakthrough will occur in the artificial physical world, ahead of the basic physical world and the phenomenal world.} Two reasons support it: the artificial physical world is more strongly constrained than either natural world (the dual constraint), and it possesses a data interface the other physical worlds lack: the design archive. The full statement of the prediction, with its falsification condition, is registered as P1 in Chapter 7: if the general world-model route, trained primarily on open phenomenal video, achieves scalably replicable success in physical manipulation before any archive-driven route does, the explanatory power of the constraint axis is severely damaged.

\subsection{3.5 The artificial physical world: a precise definition}
Among the four worlds, the artificial physical world is the object on which this paper's theory operates; it requires a characterization more precise than a working definition.

\textbf{Definition 3.1 (\allowbreak{}artificial physical world).} A family of systems belongs to the artificial physical world if and only if:

\begin{itemize}
\item (i) its instances are \emph{physical}: matter, energy, and processes in space-time;
\item (ii) each instance is constituted by a purposive \emph{design} that exists prior to it, a design that specifies what the instance is and what counts as its failure (condition C1);
\item (iii) the design is recorded in a \emph{readable generative archive}: drawings, manuals, operating procedures, standards (condition C2).
\end{itemize}
Conditions (ii) and (iii) correspond respectively to the constitution relation (\allowbreak{}Definition A.7) and the readable generative archive (\allowbreak{}Definition A.8) formalized in Appendix A (\allowbreak{}supplementary material); the legitimacy criterion of Chapter 4 (Criterion A.9) takes C1 and C2 as its two conjuncts. Conditions (i) and (ii) correspond to the two poles of the "dual nature" thesis for technical artifacts --- physical structure and intentional function (Kroes \& Meijers, 2006). Condition (iii) is the supplement that thesis lacks and learning theory requires: the dual nature answers what an artifact is, but not whether its norms are accessible to a learner. "Artificial" here is taken in Simon's sense: constituted by design rather than by nature, irrespective of the designer's species (Simon, 1969). The dependence of artifact kinds on intention receives systematic treatment in the ontology of artifacts (Thomasson, 2003); this paper does not enter that metaphysical dispute and takes only the consensus core compatible with the criterion: that the kind depends on design intention.

Each conjunct of the definition excludes one class of objects. Condition (ii) excludes every system without design intention, including all natural worlds, even though some of them (such as the genome) leave readable generative traces. Condition (iii) excludes artifacts whose design knowledge has been lost --- Stradivari violins, Roman concrete: intentionally constituted, but their constitutive norms are no longer extractable today. Chapter 4 will argue that the conjunction of these two conditions delimits exactly the full domain over which the extraction of prior frameworks is legitimate.

\subsection{3.6 Irreducibility: why this is an independent learning-task space}
Carving the artificial physical world out as a separate task space is necessary not merely because of the developmental motivation, but because of a structural fact: \emph{the core content of this world (its norms) is obtained neither automatically from learning over physical interaction nor inherited automatically from reading human corpora.} We argue the two directions of irreducibility in turn.

\textbf{Not reducible bottom-up: norms are not in interaction data.} Data from the physical world (telemetry, video, sensor streams) are descriptive. They can support learning how states evolve, but no "ought" and no "error" appears in the data stream. Norms are invisible in observation: every event in the data stream, failure events included, is a positive instance of "having happened." Error is not an observational category; an event is classified as a violation only under a norm. Inducing normative structure from interaction data is therefore an identification problem with positive examples only, and such problems have theoretical limits: two worlds can generate the same observation stream while differing in normative boundary --- this is exactly the content of Hypothesis H-N in Appendix A (\allowbreak{}supplementary material); the precise statement is the non-identifiability boundary of Proposition A.13, whose first tier is a direct application of Gold's theorem (Gold, 1967). On top of this stands an independent difficulty of magnitude: the most informative failure data are rare, expensive, and privatized. Hence the general world-model route, which takes open phenomena as its primary training object, learns the physics of this world at any scale but not its norms: a world model can be entirely right about what happens next and remain silent on whether it should happen.

\textbf{Not reducible top-down: norms are not in the statistical digestion of corpora.} Design archives exist as text and have therefore entered the training corpora of large language models. But statistical language modeling flattens the "should" in "the supply water temperature should be 7 $^\circ$C" into "people usually write this way": the model can recite the norm, yet no mechanism makes it bound by the norm; its output is plausible text, not an auditable state. The full development of this argument, including the treatment of whether structures emergent inside LLMs are equivalent to priors, appears as O1 in Chapter 8. It suffices to register here: the corpus provides a symbolic shadow of this world, not a constitutive grasp of it.

Taken together, the two arguments give the actual content of this chapter's partition: learning the artificial physical world must proceed along a dedicated path --- \emph{reading the design archive by way of promulgation, and anchoring the extracted norms to concrete instances.} This path differs from general-purpose learning from interaction as much as from general-purpose learning from corpora; Chapters 4 and 5 give its criterion and its structure. A restrained corollary follows. If this chapter's partition holds, then different worlds require different learning mechanisms, and the hypothesis that a single model learns all worlds needs to be justified world by world rather than serving as the default premise; the programmatic development of this corollary appears in Chapter 9.

\subsection{3.7 The true dividing line: whether a world comes with design rules}
Definition 3.1 yields a dividing line sharper than "artificial versus natural." The real divide is not whether something is artificial; it is \emph{whether there exist readable rules that precede instances and constitute and regulate them}. This line not only separates the four worlds; it cuts through the artificial symbolic world itself.

On one side of the line, the artificial physical world (via design archives) and the formal part of the symbolic world (code) share the same property: rules precede instances. The program exists first, the process runs afterward, and correctness can be mechanically adjudicated by compilers and tests. This is precisely why programming became one of the first large-model applications to enter large-scale commercial deployment (Section 3.4). On the other side of the line, the phenomenal world, the basic physical world, and language all lack rules prior to instances: language has no promulgating authority, and grammar is a posterior compression of pre-existing usage (Section 4.1). The side that comes with rules permits the learning direction "framework first, instances later"; the other side permits only bottom-up induction from instances.

One misreading must be blocked here: \emph{permitting is not possessing.} Most of today's coding models do not explicitly carry a prior framework; yet their success depends on the rule-first character of the code world. The training corpus is pre-governed by programming-language norms, and erroneous code can hardly enter the repository. The prior resides upstream of the data rather than inside the model. Chapter 4 gives the full argument for this dividing line: what the criterion certifies is precisely a domain that sits on the correct side of the line and whose archives are accessible.

\subsection{3.8 Range limits and boundary-setting}
\textbf{Range limit: the world of human organizations is off the axis.} This framework does not list the world of human organizations (\allowbreak{}institutions, firms, markets) as a fifth world. The reason is not that it is unimportant; it is that the one-dimensional axis cannot characterize it. The constraint axis measures the learning conditions a single agent faces: size of the semantic space, data interface, rule source. The defining feature of the organizational world is multi-agent games: other agents change their behavior in response to the learner's actions, and the learning environment itself is non-stationary. This is not a matter of stronger or weaker constraint; the \emph{type} of the learning problem has changed. A one-dimensional instrument cannot measure it, just as a ruler cannot measure weight. The organizational world therefore has no position on the constraint axis, and none of this paper's ordering predictions, P1 included, covers it. It should also be noted that the legitimacy criterion of Chapter 4 is domain-independent: it applies to the organizational domain as well and yields a precise graded diagnosis (the gradient analysis of Section 4.3; for the structural corollary concerning the organizational domain, see structural corollary II at the end of Chapter 7). The world partition has a limited range; the criterion's range is not so limited. This division of labor between a one-dimensional tool and a domain-independent criterion is what is at work here.

\textbf{Boundaries against three close relatives.} Section 2.4, R.7, has already drawn the boundary for "number-of-levels" claims; here we draw boundaries only for the world partition. The four-worlds idea has several precursors; we set the boundaries one by one to fix this paper's scholarly coordinates.

\emph{Popper's three worlds.} Popper divided existence into the physical world (W1), the world of subjective experience (W2), and the world of objective knowledge (W3) (Popper, 1972); the artifact as a unified whole constituted by design has no independent place in that trichotomy. The present framework gives artifacts a home: the artificial physical world is the institutionalized interface between W1 and W3, where knowledge not merely "concerns" objects but "constitutes" them. This interface is the domain over which the criterion of Chapter 4 operates.

\emph{Hartmann's strata of emergence.} Hartmann divides existence into emergent strata of matter, life, psyche, and spirit (Hartmann, 1940). We honor the layered intuition while drawing the standard clearly: his strata are strata of being; this chapter's are strata of learning: the partition is based on the data conditions the learner faces, not the ontological status of objects.

\emph{Simon's sciences of the artificial.} Simon's proposal to build a science around artifacts (Simon, 1969) is the closest precursor: he asks how artifacts are possible and how they are designed; this paper asks how artifacts are learnable, and the present work can be read as supplying a learning-theoretic chapter for the sciences of the artificial.

\emph{The parallel-systems approach.} Fei-Yue Wang's parallel-systems method constructs artificial systems corresponding to real complex systems, supporting management and control through computational experiments and real-virtual interaction (Wang, 2004). What needs to be set straight: its "artificial world" is an artificially made \emph{virtual} world (a digital sandbox for computational experiments) whereas this paper's artificial physical world is an artificially made \emph{real} world, physical artifacts plus archives written-to-be-read. The two approaches stand in contrast, and their objects are different things.

\textbf{Chapter summary.} This chapter has proposed a one-dimensional analytical framework: four object worlds declared with the developmental sequence as motivation; three qualitative dimensions characterizing their differences in learning conditions; and a constraint axis that explains the historical order of AI breakthroughs and yields a falsifiable prediction (P1). The framework thereby converges on the one physical world that satisfies both "rules prior to instances" and "readable archives", the artificial physical world (\allowbreak{}Definition 3.1), and argues that its learning task can be reduced neither to generic interaction learning nor to generic corpus learning. The next chapter answers the question this raises: under what conditions is it legitimate to extract a prior framework from a world?

\section{Chapter 4. The Legitimacy Criterion for Prior Frameworks: When May the Framework Come First and Learning Second?}
This chapter's criterion also answers a question that the diagnosis-theory tradition has presupposed: Reiter's (1987) MBD runs its diagnostic computation on the antecedent that the system description (SD) is prior and authoritative; Proposition 4.1 of this chapter gives exactly the conditions under which that antecedent holds --- when an SD legitimately precedes instances and constrains them. (For the full boundary against MBD, see Section 2.4, R.8.)

\subsection{4.1 Three asymmetries}
The criterion is motivated by three facts: whether a symbolic structure precedes or follows the world it describes has three different answers in three situations.

\textbf{Grammar follows language.} Natural language served for long ages before the first grammar was written down. Every written grammar is a posterior compression of pre-existing usage: a statistical summary of what speakers were already doing, rather than the generative cause of those doings. This is the uncontested core of usage-based linguistics (Tomasello, 2003; Bybee, 2010). Its engineering corollary was delivered by the history of rule-based NLP: handwritten grammars cannot cover the variation and ambiguity of real text (Manning \& Schütze, 1999; Church, 2011). A rule system distilled from a corpus cannot turn around and serve as the machine that produced the corpus.

\textbf{Laws follow phenomena.} Newton's laws stand to the motion of bodies as grammar stands to speech: an elegant, late-arriving compression of regularities that had been running for eons before they were stated. Physics is the most successful descriptive framework humanity has built, yet physical laws are \emph{discovered}, not \emph{promulgated}: the phenomenal world does not consult its own laws; it is the laws that read the world. A learner who wants the framework before the data therefore has no choice but to reverse-engineer the framework out of the data --- which is what machine learning does at scale, and what no handwritten rule system ever managed. Injecting physical laws explicitly into a neural network's loss function (the PINNs family) does not change this asymmetry; its boundary is given in Section 4.3.2, and the full comparison is in Section 2.4, R.6.

\textbf{Drawings precede artifacts.} Now consider a building, an airplane, a power plant. Here the symbolic structure (drawings, specifications, bills of materials, control logic, industry standards) is not distilled from the artifact after the fact; it is promulgated prior to the artifact. It governs the artifact's generation: what to build, how to build it, what the finished thing must be like to pass. A drawing does not describe what a building \emph{is}; it prescribes what a building \emph{shall be}. For the construction industry, drawings are not descriptive tools but conditions of existence: burn a building's drawings and the building still stands; but if the practice of drawing itself disappeared, the construction industry (the entire institution of design, construction, acceptance, and maintenance) could not exist in its present form. The third situation is no marginal variant of the first two; it is a different causal topology: the symbolic structure is no longer a post-hoc product of instances, but a precondition for instances to come into being at all.

That the three situations coexist means this: "framework first" fails in some worlds and is daily routine in others. The chapter's question can now be posed: \emph{under what conditions is it legitimate for a machine learning system to extract a prior framework first and let that framework govern subsequent learning?} Section 4.2 gives the criterion, Section 4.3 develops its gradient, Section 4.4 argues its sufficiency, and Section 4.5 states its boundaries.

\subsection{4.2 The criterion}
\subsubsection{4.2.1 Why a temporal criterion fails}
Theoretical reflection naturally begins with a temporal formulation: a prior framework is legitimate in any world that is \emph{designed first and born second}. On this criterion, artifacts qualify and language does not. But it admits an object that should not qualify: the organism. The gene precedes the protein, and evolution precedes every individual; in sheer temporal order, the gene stands to the body in the same "prior" position as the drawing stands to the building. Intuition tells us the two should not be treated as one case: erecting priors from drawings is engineering routine, while erecting a prior framework from genes has never succeeded. A competent criterion should exclude the gene. The temporal criterion lets in the very case it meant to exclude, which proves that temporal precedence is a surface phenomenon and cannot serve as the criterion.

Return to the three asymmetries of Section 4.1. Grammar only summarizes; laws only interpret; drawings do a third thing: they prescribe. The discriminating variable is therefore not \emph{when} the symbolic structure exists but \emph{what it does}. A design document is promulgated with a purpose: it was written down precisely in order to prescribe what the artifact shall be. Promulgated norms thereby accomplish what no gene and no grammar can: they \emph{constitute} their object and \emph{define} the object's norms. The difference is visible in a linguistic fact: deviation from a drawing is called a \emph{fault}; deviation from a gene is called a \emph{mutation}. Evolution contains no errors, only outcomes; the category "bad" exists only in artificial worlds. Condensation on a roof is, for the phenomenal world, merely a natural phenomenon; for a building it may be a fault. "Bad" exists only relative to norms that can be satisfied or violated. Purpose, function, failure (the whole normative vocabulary on which engineering and operations run) is defined relative to intentional design, relative to norms (Kroes \& Meijers, 2006; Searle, 1995 on constitutive rules; Dennett, 1987 on the design stance). We call such norms \emph{constitutive norms}: the design precedes the artifact not only in time but in the order of reasons; it defines what the artifact is supposed to be.

This distinction also separates a prior framework from any "model that has read all human documents" (developed fully in Chapter 8, O1).

\subsubsection{4.2.2 From an ontological condition to an epistemic condition}
Constitutive normativity is an ontological property, and a criterion meant to guide \emph{learners} cannot be ontological alone. A framework can exist yet be out of reach: Stradivari violins and Roman concrete are indisputably intentional constitutions, yet the knowledge of the making that constituted them can no longer be fully extracted today: reproducing the recipe of Roman concrete remains an active research frontier (Seymour et al., 2023). However definite the original purpose, it cannot be extracted now. The ontological question, "was it intentionally constituted?", answers what is true; a criterion in the learning-theoretic sense must also answer what is \emph{accessible}. Since the stance of this research is instrumental, the criterion is forced to carry an epistemic hemisphere.

The two hemispheres join into one conjunction:

\begin{quote}
\textbf{The Legitimacy Criterion.} Extracting a prior framework from a world W is legitimate if and only if: (i) W is \emph{intentionally constituted}: there exists a purposive design, prior to its instances, that constitutes the instances and defines their norms; (ii) this constitutive process has left behind a \emph{readable generative archive}.
\end{quote}
The domain the criterion certifies is exactly the artificial physical world of Definition 3.1. Each conjunct kills a class of counterexamples the other cannot kill. Worlds "designed but with archives lost" (violins, concrete, every artifact whose archive has burned) die by (ii): the priors exist in principle but cannot be obtained in practice. Worlds "readable but without purpose" (genomes, multiple sequence alignments, the fossil record) die by (i): the archive can be read, but what is read out is structure without normativity. Section 4.3.2 will argue that this is why AlphaFold can recover a protein's shape but cannot recover what the protein \emph{is for}. Normativity travels only with intention.

The conjunction pays one further dividend. Conjunct (i) supplies philosophical completeness: the priors of a designed world hold \emph{authority} over its instances, not mere correlation with them. Conjunct (ii) supplies the practical boundary: legitimate prior extraction is confined to artifacts that have archives and are institutionally maintained. The theoretical boundary and the deployable application boundary thus coincide, drawn by the same two conditions; the theoretical positioning is developed in Section 4.3.5, the engineering boundary in Section 4.5.

\subsubsection{4.2.3 The formal skeleton of the criterion}
We now give the semi-formal statement of the criterion.

\textbf{Definition 4.1 (object domain and constitution).} An \emph{object domain} D is a triple $\langle$I, N, A$\rangle$: I is a set of instances (the observable, identifiable physical objects and processes of the domain, i.e., matter, energy, and processes in space-time, as in condition (i) of Definition 3.1); N is a set of norms (\allowbreak{}statements about how the instances in I \emph{ought to be}); A is an archive (a machine-readable record in symbolic form). We say the norm set N \emph{constitutes} the instance set I, written N $\lhd$ I, if and only if: the existence and identity of every instance in I depend on some set of norms in N being complied with, and the fixing of N precedes the production of the corresponding instances in I. Two semantic boundaries against misreading: $\lhd$ is not the necessity operator of modal logic (it says nothing about possible worlds), nor is it the concept inclusion of description logic (what it relates is not concepts to concepts but norms to instances). The semantic anchor of $\lhd$ is the act of \emph{promulgation}: an intentional promulgation that prescribes what instances shall be. Its lineage is the branch of the normative-systems tradition formalized long before this paper (Section 2.4, R.2).

\textbf{Definition 4.2 (\allowbreak{}promulgation, description, and readable generative archive).} For a norm n $\in$ N and an instance i $\in$ I: if an i that violates n is adjudged a \emph{defect of the world} (i is judged faulty, in violation, to be corrected) then n bears the direction of fit of \textbf{promulgation} toward i; if an i that violates n is adjudged a \emph{defect of the model} (n is judged mismatched, to be revised) then the direction of fit is \textbf{description}. (The directions are directions of fit in the Anscombe--Searle sense (Anscombe, 1957; Searle, 1983): promulgation = world-to-word; description = word-to-world.) An archive A is a \emph{readable generative archive} of D if and only if: (i) A records the promulgated content of (part or all of) N; (ii) A exists prior to the corresponding instances; (iii) A exists in a machine-parseable symbolic form.

\textbf{Proposition 4.1 (the Constitution Criterion).} For an object domain D = $\langle$I, N, A$\rangle$, extracting a prior cognitive framework over D is legitimate if and only if: (C1) N $\lhd$ I; (C2) there exists a readable generative archive A of D. (Alias registration: this criterion is the "Legitimacy Criterion" of this chapter's title and the "Promulgation Criterion" of the abstract and Chapters 1 and 6; the three names refer to the same criterion, and the text takes each up as context dictates; the necessity direction is argued below, and the sufficiency direction is adopted as this paper's normative commitment --- see the note on the logical relation below and Criterion A.9.)

\textbf{Argument (necessity).} Suppose C1 fails: instances are not constituted by prior norms. Then the content of any "prior framework" can only come from induction over pre-existing instances; its direction of fit is description, and the "framework" is in fact a product of posterior learning --- prior in name only (Section 4.3.2: physical laws relative to the phenomenal world are this case). Suppose C2 fails: norms constitute instances but leave no readable record. Then there is nothing from which the framework could be extracted, and legitimacy has no engineering landing (Section 3.3: the situation of the basic physical world and the phenomenal world: constraints objectively exist, but there is no archive).

\textbf{Corollary 4.1 (\allowbreak{}adjudication of the four worlds).} Checking domain by domain against Proposition 4.1 (the three-dimension table of Section 3.3): the phenomenal world fails C1 (no prior constitutive norms); the basic physical world fails C1 (laws are descriptions, not promulgations); the artificial symbolic world, checked by the natural extension of Section 3.7, satisfies C1 only in a split manner (the formal/code part alone) and C2 only partially, and its norms do not constrain physical instances; the artificial physical world satisfies C1 and C2 simultaneously: it alone is the object domain of legitimate prior frameworks. The gradient-level development is in Section 4.3.

\textbf{Boundary note (on the jurisdiction of the criterion).} Definition 4.1 has already restricted instances to physical objects and processes, and this delimits the criterion's jurisdiction. One boundary case must be addressed on the spot: the norms of programming languages and formal systems precede their process instances and are readable, and so appear to fall within the criterion. Strictly speaking, the criterion's formal statement takes physical domains as its jurisdiction; its application to the formal part of the symbolic world is a natural extension of the criterion and has been handled by the dividing line of Section 3.7. The same extension logic also covers the organizational domain (the middle band of Section 4.3.3), and the distinction between the range of the world partition and the range of the criterion is registered in Section 3.8.

\subsection{4.3 The gradient of the criterion: two axes}
Each conjunct of the criterion is a continuous axis rather than a binary switch: constitution comes in degrees of strength (axis one), and archives in degrees of readability (axis two). How far the criterion applies to an object domain is determined jointly by the domain's position on the two axes. This section develops the two axes in turn and then discusses their superposition. The conclusion first: only one corner is high on both axes, and the artificial physical world sits there.

\subsubsection{4.3.1 Axis one: the strength of constitution}
The binding force of design is strongest in the artificial physical world: \emph{drawings constrain matter that cannot read drawings}. Promulgation takes effect once, and instances do not renegotiate what the drawings mean. Moving down the axis, the strength of constitution decreases.

\textbf{Regimented organizations (medium-strong).} When the objects governed by norms are themselves intentional agents, constitution loosens. Armies, intelligence agencies, and government departments are indeed intentionally instituted, and their normativity is real (violating an operating procedure brings disciplinary action, not mutation) but the members of an organization continuously interpret, negotiate, and revise the actual force of its charter. An organization is not an artifact "promulgated once"; it is a collective that re-promulgates its rules every day (Searle, 1995 on institutional facts). Constitutional strength rises with the degree of regimentation: the more codified the mission, the closer the situation to that of the physical domain.

\textbf{Commercial organizations (medium).} The written charter and the organization's actual operation split apart: the mission statement is one thing, the way things are done another. Normativity exists, but the coverage of promulgation is limited.

\textbf{Language (weak to none).} Language is not a case of "the archive being unreadable": grammars exist, and the linguistic literature is vast. Language's problem lies on axis one: grammar does not constrain language. Every written grammar is a posterior summary of pre-existing usage (Section 4.1). When a speaker deviates from a rule in the grammar book, no violation judgment is triggered: no institution promulgates the norms of language, and no one is ever corrected as a defect of the world for "getting a sentence wrong." The failure of rule-based NLP thereby receives a precise diagnosis: it failed on axis one --- handwriting norms where no constitutive norms exist, producing something fated to be a description of descriptions.

\textbf{Natural worlds (none).} Physical laws are discovered descriptions and promulgate nothing (Section 4.1): when a phenomenon deviates from a law, it is the law that gets revised, not the phenomenon. Axis one reaches its zero point here.

\subsubsection{4.3.2 Axis two: the readability of archives}
Archives differ enormously in whether they are \emph{written-to-be-read}, and the extraction cost scales accordingly. From highest to lowest readability, there are three typical situations.

\textbf{Written-to-be-read.} Engineering drawings, BIM models, data sheets, control logic, codes and standards (and, for human organizations, charters, regulations, and procedures) are symbolic artifacts produced to carry constitutive information across personnel and time. Their semantics are standardized by institutions (naming rules, symbol libraries, classification systems), and their intended reader is precisely the stranger to the project. For a learner, this is the cheapest data interface: the world has pre-segmented itself.

\textbf{Happens-to-be-readable: a natural experiment that isolates the two conditions.} The genome was written by no one for no one, yet it is an archive: a high-fidelity trace of a generative process, preserved and partially decipherable. The genome sits in the middle of axis two and at zero on axis one, satisfying the weakened archive-side condition C2$'$ (readable generative trace) while failing conjunct (i). Strictly speaking, Definition A.8 requires the archive to record \emph{promulgated content}, and without promulgation there is no C2 to speak of, so wherever this section speaks of "(ii) satisfied by accident," C2$'$ is meant. The genome therefore provides a natural experiment isolating the effect of each conjunct. The two generations of protein-structure prediction are the experimental result. The first AlphaFold relied on heavily hand-engineered pipeline structure (Senior et al., 2020); the second abandoned these hand-built structures and learned end-to-end directly from the same evolutionary record (the multiple sequence alignment), outperforming the first generation by a wide margin at near-experimental accuracy (Jumper et al., 2021). The comparison tells us two things. On one hand, C2$'$ can indeed be satisfied by accident: an archive no one wrote can still be mined, and the best way to mine it turns out to be abandoning the hand-built framework and learning from the archive directly. The hand-designed pipeline was not the information source; the archive itself was. A cut must be made here: what was abandoned is \emph{descriptive} hand-made priors (expensive guesses about a world without promulgation) which is a different kind of thing from extracting promulgated norms out of a constitutive archive (see the "no archive" paragraph below). On the other hand, conjunct (i) cannot be replaced by any mining technique: a multiple sequence alignment records only which variants survived evolution; it contains no information about intention. The recovered framework can therefore rank structures but cannot say what a protein \emph{is for}, or what would count as its \emph{failure}. Structure is accessible; norms are not. The protein is at once a crucial experiment between the temporal criterion and this criterion: the temporal criterion cannot distinguish gene from drawing --- on it, prior extraction should be as feasible in the evolutionary domain as in the design domain. This criterion distinguishes the two and predicts that the evolutionary domain yields only high-cost, non-normative structure. What actually happened (structure recoverable, norms not) supports the latter.

\textbf{No archive.} The phenomenal world and the basic physical world occupy the other end of the gradient: the world itself keeps no generative record; there is no document from which a storm or a rock was produced. The texts of physical laws that humans write are not the world's archive but \emph{our} compression: they record the conclusions of centuries of observation, not the norms from which instances were produced. Injecting such compressions explicitly into a neural network's loss function (PINNs (Raissi et al., 2019)) does work on simple boundary-value problems, but it does not change their epistemic identity: a descriptive prior is an expensive substitute in a world without archives; in a world with archives, it gives way to constitutive norms. For the surgical comparison with this nearest neighbor see Section 2.4, R.6; for treatment in an adversarial context see Chapter 8, O5.

\subsubsection{4.3.3 Superposition of the two axes: the organizational middle band}
The special position of human organizations is that they land in the gray middle band on \emph{both} axes at once. On axis one, their constitution is real but loose (Section 4.3.1). On axis two, their archive splits into two parts. One is the \emph{written archive}: charters, regulations, procedures --- written-to-be-read, at the highest end of readability. The other is the \emph{living archive}: the way the organization actually runs, much of which is tacit knowledge that cannot be fully codified (Polanyi, 1966), sliding toward the "no archive" end. Here the criterion's prediction is concrete: for any object whose archive is split in this way, the construction cost of a prior framework must include a "human reader" fee, rising with the share of the living archive. This fee already has its industrial form: Palantir's forward-deployed engineers (FDE) do this work: stationed on site, transcribing the living archive into computable semantics. The superposition effect of the two axes is equally testable: the higher the degree of regimentation (axis one moving toward the strong pole), the more the written archive coincides with the living archive (axis two moving toward the high end), and the higher the value density of a prior framework: which explains why the most successful deployments of such frameworks concentrate in the most regimented organizations, defense and intelligence. Further structural corollaries for the organizational domain appear in structural corollary II at the end of Chapter 7.

\subsubsection{4.3.4 Closing: positioning on the two axes}
The two axes together give the applicability map of the criterion. Axis one, from strong to none: artificial physical world $\succ$ regimented organizations $\succ$ commercial organizations $\succ$ language $\succ$ natural worlds. Axis two, from high to none: written-to-be-read $\succ$ happens-to-be-readable $\succ$ no archive. The criterion's full claim domain is the corner that is high on both axes, where only the artificial physical world sits --- the one physical world whose rules precede instances and whose rules are written-to-be-read. Objects in the middle band, such as organizations, can benefit partially, their benefit-cost ratio set by the shade of gray (Section 4.3.3). Worlds at zero on axis one can only learn bottom-up, starting from instances. Extraction cost rises as readability falls; we conjecture the rise is superlinear, though this paper does not attempt to formalize the rate. Section 4.4 will argue that when both axes are high, the extracted framework is not merely legitimate but sufficient to support operable intelligence.

\subsubsection{4.3.5 The claim of this paper}
The central thesis of this paper can now be stated head-on. The four-world adjudication of the criterion (Corollary 4.1) and the two-axis positioning (Section 4.3.4) jointly establish: \emph{the artificial physical world satisfies both conjuncts of the criterion and is the object domain of legitimate prior frameworks.} From this follows the claim announced in the paper's title: the legitimate path to machine intelligence for the artificial physical world is to extract prior frameworks from its constitutive archives, rather than to induce frameworks from interaction data or human corpora; neither interaction data nor human corpora supply the framework's promulgative authority over instances (Section 4.2.2).

One guardrail prevents the claim from being over-read: legitimate does not mean sufficient. The claim so far argues only that the framework may legitimately be erected first; what kind of operable intelligence it can support is completed by the sufficiency argument of the next section (\allowbreak{}Proposition 4.2).

\subsection{4.4 Why the criterion is sufficient: the reversal of direction of fit}
So far the criterion has been tested by counterexamples: each conjunct excludes a class of cases the other cannot. But this answers only \emph{why these two conditions are necessary}, not yet \emph{why they are sufficient}: why should a framework satisfying the criterion support operable intelligence, rather than amounting to a bundle of legitimate but useless statements? Answering this requires bringing to the front a concept already registered in Definition 4.2: \emph{direction of fit}. It is the load-bearing structure of sufficiency.

The two directions of fit were registered in Definition 4.2: for descriptive rules, the rule fits the world; for constitutive rules, the world fits the rule. The three asymmetries of Section 4.1 are, in the end, this distinction. Why it bears load can be stated directly: for a descriptive rule, the next instance can always be a counterexample --- inductive risk cannot be eliminated; for a constitutive rule, deviation is the world's violation, and the framework inherits promulgative force itself. The operational consequences of the two directions are given by the following definition.

\textbf{Definition 4.3 (violation resolution).} Let dev(i, n) denote the event that instance i deviates from norm n. The direction of fit determines the resolution operator: if n is promulgated, dev(i, n) resolves to \textbf{Violation(i)}, a defect of the world (locatable, reportable, correctable against the promulgated clause); if n is described, dev(i, n) resolves to \textbf{ModelDefect(n)}, a defect of the model (absorbed by further learning into a parameter revision). The two resolutions are not interchangeable labels but two different event types with different downstream operators attached: the former issues an alarm with an address; the latter silently updates a distribution.

\textbf{Definition 4.4 (model-side events that are not deviations).} ModelDefect(n) takes a described norm n as its parameter; two further event kinds have no such n and are registered separately. First, \textbf{CoverageGap} (coverage gap): a reading that finds no slot, or a reduction that lands in no concept-layer enumeration --- neither Violation nor ModelDefect; judged "unknown" per A3 and referred to the augmentation passage of Section 5.2 (instances: B.1.5.4 in the supplementary material). Second, \textbf{MaintenanceFault} (\allowbreak{}maintenance fault): promulgated content in mutual contradiction (an internal inconsistency of the knowledge layer --- the slots and the enumerations are all present, but the answers conflict) --- the alarm of A1 (archive fidelity) failing; its handling runs through the maintenance process, through neither learning nor augmentation. Neither CoverageGap nor MaintenanceFault is a fit resolution of a deviation event, and neither enters the two-policy enumeration of Criterion A.5.

Sufficiency further requires four working assumptions about the archive: \textbf{A1 (archive fidelity)}: the archive faithfully records the currently promulgated norms, maintained and versioned; \textbf{A2 (semantic accessibility)}: the learner can correctly interpret the archive's vocabulary, parseable and groundable; \textbf{A3 (coverage)}: the instances judged fall within the archive's coverage (the assumption proper); beyond coverage, the judgment is "unknown," never a silent error (the operating policy --- the sufficiency core S1/S2 uses only the former, while S3's "unknown" tail clause uses the latter); \textbf{A4 (\allowbreak{}promulgation direction on duty)}: at runtime, the fit policy of the promulgation direction is maintained (Criterion A.5): when an instance deviates, the norm side is held fixed and the world side is marked to-be-corrected, and the archive is not back-revised to accommodate field deviations (an "as-built drawing" style reverse accommodation is exactly A4 failing). (Numbering-scope note: the working assumptions A1--A4 carry no period, a different domain from the appendix items A.1--A.4.)

\textbf{Proposition 4.2 (the reversal of direction of fit).} If a domain D satisfies C1 and C2, then under A1--A4, a framework F extracted from the archive satisfies:

\begin{itemize}
\item \textbf{S1 (zero-shot operability)}: F is fixed prior to instance data; a system loaded with F possesses adjudicative capacity before the first instance arrives;
\item \textbf{S2 (\allowbreak{}promulgated constraint)}: the direction in which F constrains instances is promulgation: a deviation of an instance from F resolves to Violation, not ModelDefect;
\item \textbf{S3 (loud failure)}: deviation necessarily triggers a violation judgment: failures are locatable and reportable.
\end{itemize}
\textbf{A note on the logical relation.} The sufficiency direction is the normative commitment this paper adopts (Criterion A.9); the argument for the necessity direction has been given with the proposition. The "if and only if" of Proposition 4.1 supplies \emph{legitimacy}; A1--A4 supply \emph{operating conditions}. The full antecedent of sufficiency is therefore C1 $\wedge$ C2 $\wedge$ A1 $\wedge$ A2 $\wedge$ A3 $\wedge$ A4. This does not weaken the criterion: the criterion adjudicates where a framework may legitimately be erected; the working assumptions adjudicate whether the archive is on duty. The two answer different questions.

\textbf{Proof.} S1: by C2 (\allowbreak{}Definition 4.2), the archive exists prior to instances and is machine-readable, so F can be fully extracted before any instance arrives; A1 guarantees that what is extracted agrees with the currently promulgated norms. S2: by C1 (\allowbreak{}Definition 4.1), instances are constituted by N; by the definition of promulgation (\allowbreak{}Definition 4.2), an instance violating a constitutive norm is adjudged a defect of the world, and A4 guarantees that this resolution policy stays on duty at runtime --- that is, dev(i, n) necessarily resolves to Violation rather than ModelDefect (\allowbreak{}Definition 4.3). S3: by S2, every deviation produces a Violation judgment whose location is given by F's normative semantics (which clause was violated); by A3, instances beyond coverage are judged "unknown" rather than silently passed. \hfill$\square$

The three derivations do not bear load evenly: S1 is substantively supplied by C2 together with the working assumptions; S2 and S3 are direct unfoldings of Definitions 4.2 and 4.3 --- their content was already written into the definitions of promulgation and Violation. This is a deliberate construction: the argumentative load of sufficiency is front-loaded into the definitional stage of the criterion and direction of fit, and the proof itself is responsible only for calling the chain by name. We now explain the operational meaning of S1--S3, each corresponding to one dimension of "operable."

\textbf{S1 support: accessibility supplies the head start.} The archive is readable before instances; the framework can be fully extracted before encountering any instance --- operable intelligence holds from the first day of deployment, and the cold-start deadlock of Chapter 1 is thereby unlocked. This also previews one construction goal of Chapter 5: the framework must provide landing grounds for the data stream (Section 5.2, openness).

\textbf{S2 support: constitution supplies induction-free generalization.} The binding force over future instances of statements extracted from a constitutive archive is not statistical extrapolation but constitutive constraint: instances are \emph{made} to satisfy the framework; the framework is not \emph{guessed right}. A statistical model's validity on new instances is probabilistic: trustworthy inside the distribution, speechless outside it. A prior framework's validity on new instances comes from promulgation: as long as the object remains the artifact that was built, accepted, and maintained according to its design, the framework's statements about it continue to hold --- on the strength of the norm itself rather than of sample size. For promulgation to stand, norms must not drift; this previews the stability goal of Section 5.2.

\textbf{S3 support: normativity supplies alarming failure.} Instances will of course deviate from design: wear, misuse, aging, operating-point drift. When a statistical model meets an out-of-distribution sample, its failure is \emph{silent}: it keeps outputting, merely ceasing to be trustworthy. When a constitutive framework meets a deviation, Definition 4.3 resolves it as Violation, classified on the spot as a trend, a symptom, or a fault. In the context of operations, "operable" means not never failing but failing with a report every time: an intelligence that does not know it is wrong cannot become an endorsable bearer of responsibility in institutional settings. For the alarm to be locatable, judgments must be anchored to concrete instances; this previews the concreteness goal of Section 5.2.

\textbf{Corollary 4.2 (\allowbreak{}institutional adoptability).} A framework satisfying Proposition 4.2 additionally satisfies three requirements of institutional settings: \textbf{(I1) norm accessibility}: the system can access and cite the promulgated norms that constitute the world, rather than statistical regularities alone (supplied by C2); \textbf{(I2) violation audibility}: when the world deviates from promulgated norms, the system's output can be classified as a judgment of "violation" rather than mere "prediction error" (supplied by S2 and S3); \textbf{(I3) accountability closure}: the reasoning path from observation to verdict can be reconstructed and endorsed by the promulgating body (supplied by S2: the verdict is checked against named promulgated clauses, not anonymous parameters). I1--I3 together give the operational definition of "operable" in institutional contexts, and they are the formal pedestal of the "interrogable intelligence" of the concluding chapter.

If the criterion holds, its practical consequence is already visible: in design domains with readable archives, operable intelligence need not wait for large-scale posterior data. What shape this framework must take (why it must be a four-layer structure of syntax, concepts, knowledge, and instances) is the business of Chapter 5.

\subsection{4.5 Claim domain and boundaries}
This section states in one place the scope of the framework and how it behaves at its boundaries.

\textbf{Claim domain.} The framework applies to artifacts that are "of high utility value and institutionally maintained": only such artifacts merit a full set of procedures and norms made by humans, and only they have archives that someone keeps maintaining (the dividend of Section 4.2.2). Household consumer goods (broken, buy a new one) are outside the claim domain; non-institutional appropriation, such as picking a lock with a paperclip, is outside it too; the framework never promised to govern every paperclip (Chapter 8, O2).

\textbf{Purpose is institutional.} The word "purpose" in this paper carries no metaphysical burden: purpose is written in requirements documents, function in specifications, failure in failure-mode analyses; it has nothing to do with any metaphysics of "nature's designer" (Dennett, 1987: the design stance is a strategy, not a theology). The criterion takes effect with an institution's archiving behavior and exits when that behavior ceases: where an institution stops keeping archives, the framework exits honestly, claiming no victory beyond the boundary.

\textbf{Definitional treatment of force majeure.} Anticipating a conclusion Chapter 5 will establish (failure types can be enumerated and maintained (Section 5.3)) we can register a boundary corollary now: events "entirely beyond the known type list and not interpretable as superpositions of known types" can be \emph{defined} as force majeure, as war and natural disaster stand to an operations service. The framework does not fail on such events, just as an insurance contract does not fail on war.

\textbf{The organizational domain.} Human organizations sit in the criterion's middle band (Sections 4.3.1, 4.3.3): normativity genuinely exists but constitution varies in strength, and the archive is semi-readable, split into written and living parts. The criterion's structural corollaries for the organizational domain (value density rising with regimentation, the "human reader" cost being structural, no cross-organization universal concept layer emerging) are collected in structural corollary II at the end of Chapter 7.

\textbf{Remark 4.1 (why not simply learn the norms?).} One predictable objection remains: since operating data from the artificial physical world will eventually accumulate sufficiently, why not simply learn the regularities and set the archive aside? Two mutually independent reasons, both structural rather than operational. First, \emph{instability}. Statistically induced "rules" are summaries of a data distribution and drift with it; the definition of a constitutive norm is precisely that it is independent of any distribution. The learned copy is worse than imperfect: it does not know it is a norm. It carries no violation semantics (deviation from a statistical regularity is indistinguishable from noise (by Definition 4.3, all its deviations resolve to ModelDefect)) whereas deviation from a promulgated norm is a Violation, with a direction of fit, a responsible party, and a required handling. Second, \emph{uneconomical}. Spending data, compute, and calibration cost to re-derive a document that was promulgated long ago and reads for free buys an inferior copy that cannot be held accountable --- while the original sits on the shelf. Posterior learning is indispensable; it is the flesh. But it must not be allowed to counterfeit the skeleton. This is also the deep content of prediction P2: the decay of the data head start is not data losing value, but data being unable to buy what only promulgation can supply.

\textbf{Remark 4.2 (the authorless artifact).} A thought experiment can sharpen the criterion's edge. Suppose a machine is designed entirely by an AI system and 3D-printed: no human intention, no readable documents. Yet anyone can see at a glance that it is an electric fan, and in use it is an electric fan. When it stops turning, we say without hesitation that it is \emph{broken}; but note where this "should" comes from. The artifact itself carries no promulgated norms; "broken" is adjudged against \emph{our} archive for fans --- airflow, duty cycle, temperature-rise limits, all promulgated by human engineering practice. The object is readable not because it ships with norms, but because it landed in a world that ships with norms; its readability is borrowed.

Reverse-engineering the artifact is of course possible, but its product is a hypothesis, not a norm: no community will come to correct a misreading. Using it to interpret the AI's next-generation design runs into the direction-of-fit deadlock: when the artifact deviates, there is no telling whether the AI violated its design norms or the reverse engineering guessed wrong, and the framework is left with the revise-the-model-at-every-deviation direction, a live version of the boundary stated in Proposition A.13.

A deeper clarification: the criterion is not anthropocentric. This artifact is excluded not because an AI made it, but because no agent (human or AI) promulgated its norms into a readable archive. Promulgation is an act, and any signatory counts: a design system that signs, versions, and publishes its own rules brings its artifacts within the criterion's scope. The boundary of the artificial physical world is drawn by promulgation, not by species.

\textbf{Chapter summary.} This chapter has established the paper's central criterion: extracting a prior framework is legitimate if and only if the object domain is intentionally constituted and has left a readable archive (\allowbreak{}Proposition 4.1, the Constitution Criterion). On this basis the paper's thesis was stated head-on: the artificial physical world is the object domain of legitimate prior frameworks, and constitutive priors are the legitimate path to its machine intelligence (Section 4.3.5, the claim of this paper). Sufficiency was then argued through the reversal of direction of fit: a framework satisfying the criterion inherits promulgative force, and is therefore zero-shot operable, induction-free in generalization, and locatable in failure alarms (\allowbreak{}Proposition 4.2, the reversal of direction of fit). The criterion also drew its own boundary: it takes effect with an institution's archiving behavior and exits where that archiving ceases. The next chapter answers the question of shape this raises: what must a prior framework that satisfies the criterion look like?

\section{Chapter 5. The Necessity of Layering: What a Legitimate Prior Framework Must Look Like}
\subsection{5.1 From content to shape}
Chapter 4 established that erecting a prior framework in the artificial physical world is legitimate, and that the framework's source of content does not even have to be searched for: purposes, functions, and norms can be taken directly from the constitutive norms promulgated in the design archive. That purposes exist prior to artifacts is a fact unique to the artificial physical world. But the archive answers only the question of the framework's \emph{content}, not of its \emph{shape}: content can be piled into a warehouse or built into a tower. To find the right shape, one must first ask a question that comes earlier: in building this framework, what exactly do we want it to achieve?

\subsection{5.2 Four construction goals}
What goals should a prior framework achieve? The answer is not ours to pick: it is derived jointly from two sources (the world picture of Chapter 3 and the promulgated constraint (S2) of Proposition 4.2) and it comes out to exactly four.

Before the derivation, let the logical form of this stretch of argument be stated openly. The necessity argument for the four goals is a conditional: given the structure of the artificial physical world (\allowbreak{}Definition 3.1) and the promulgated constraint (S2) of Chapter 4, anyone intending to build an operable prior framework in this world must satisfy all four goals. The four goals are derived not from physics but from the joint force of an ontological fact ("design comes first") and an epistemic commitment --- "promulgation should constrain instances." The reader who does not accept the commitment should return to the direction-of-fit argument of Section 4.4; the reader who does not accept the fact should return to Definition 3.1. Every argument in the rest of this section is conducted inside the antecedent of this conditional.

\textbf{Goal one: stability.} The non-drifting part of the framework must span the two denotational poles --- the pole of types and the pole of clauses. \emph{The framework must have parts of speech and combination rules, and these must not drift with versions}; otherwise every later addition upstairs loses its fixed coordinate system. \emph{Knowledge clauses must be frozen into documents one by one, and once promulgated must not be rewritten}; otherwise every adjudication citing a clause loses its time-stable meaning. This is a conjunction, and both conjuncts are indispensable. If only the type pole is stable while clauses can be rewritten, the word "fault" has a definition but "whether this unit counts as faulty" has no settled verdict; if only the clause pole is stable while the vocabulary drifts, the words of the clauses stay put while their meanings move. The provenance of this goal is S2 (\allowbreak{}promulgated constraint) together with the two postures of "constraining instances": promulgation constrains instances both \emph{as types} and \emph{as clauses}, so stability must hold at both poles. Norms are the sole source of the word "wrong": a referee who revises its own rules every day is no referee, and neither is one whose rules never change while its verdicts change daily.

\textbf{Goal two: openness.} The framework must be able to take in new things continuously, again spanning the two denotational poles: \emph{the type library must have a port of entry for new concepts} (humanity never stops inventing new artifacts, and the world's object list is never closed) and \emph{instance data must have landing grounds} --- the world keeps running and keeps producing observations. The revision discipline at both passages is continuous intake. What this goal requires is the existence of the passages and their disciplines, not the actual occurrence of intake events: a framework with a complete port of entry through which no new concept happened to arrive this year does not violate the goal; what violates it is having no port at all. Neither conjunct can be missing: taking in data but not concepts freezes the service range at promulgation day; taking in concepts but not data disconnects the framework from the world. The provenance of this goal is the world picture of Chapter 3: both the object list and the observation stream keep growing.

A common misunderstanding must be corrected here. There is no symmetric tension between stability and openness --- these are not two equal enemies pulling at each other. The pulling is one-directional: stability itself has no enemy, since a dead framework that cares nothing about openness stays stable forever; \emph{only openness is a source of threat to stability}. It is because new things must still be taken in that stability needs defending; openness itself never needs defending by stability. The real question is therefore not how to compromise between two opposing requirements, but: \emph{how to grow without drifting}.

\textbf{Goal three: abstraction.} The framework's experience must be able to accumulate in conceptual form: what is learned about a concept applies automatically to all future instances of that concept, instead of modeling every instance from scratch. Its provenance is the first posture of "constraining instances": to constrain instances, a symbolic structure must be a \emph{type}, reusable across instances. The source of this goal is \emph{not} data scarcity: even with unlimited data, learning that eats nothing but instances still throws away reusable structure. Abstraction is the prior framework's job benefit; it compresses information by nature. The extreme of non-abstraction is all instances --- which is just posterior data learning, and the word "prior" loses its meaning.

\textbf{Goal four: concreteness.} The framework's concepts must be able to come down to concrete physical instances and speak: this chiller, this hospital ward, this pipe segment. A judgment has operational value only when anchored to an instance. Its provenance is the second posture of "constraining instances": as a \emph{clause} anchored to a concrete object. Otherwise, however complete its concepts, the framework is left talking to itself: a framework without instance anchoring issues no judgments.

The provenance and necessity of the four goals can be gathered into one table. Necessity is checked item by item by an \emph{ablation test}: remove any one goal, and what does the framework lose?

\begin{table}[htbp]
\centering
\small
\begin{tabular}{@{}>{\raggedright\arraybackslash}p{0.306\textwidth}>{\raggedright\arraybackslash}p{0.306\textwidth}>{\raggedright\arraybackslash}p{0.306\textwidth}@{}}
\toprule
Goal & Provenance & Consequence of removal \\
\midrule
Stability & S2 (\allowbreak{}promulgated constraint): promulgative authority must hold in both postures, type and clause & concept cores drift with data; "fault" loses its definition; the promulgation leg breaks on the spot \\
Openness & the world picture of Chapter 3: the object list and the observation stream keep growing & the framework is trapped in staleness: new artifacts, new operating conditions, new failure modes cannot enter; the service range stops at promulgation day \\
Abstraction & S2, posture one: constrain instances as types; experience accumulates across instances & degenerates into an all-instance record (i.e., posterior learning itself) and "prior" loses all meaning \\
Concreteness & S2, posture two: anchor instances as clauses; judgments must land on the ground & concepts talk to themselves and cannot anchor this chiller or this ward: concepts spin idle and never reach the field \\
\bottomrule
\end{tabular}
\end{table}
Beyond necessity, one must answer "exactly four?": why is there no fifth necessary goal? Any candidate goal can first be passed through an \emph{assignability test}. Either it constrains some region of the promulgated content and is reducible to content within an existing goal: \emph{safety}, for instance, is promulgated content itself and belongs to the knowledge clauses. Or it constrains the framework's implementation rather than its content, and is not a goal of the framework: \emph{real-time performance} is a performance constraint, \emph{economy} a value function. Or it constrains relations among goals rather than content, a structural property of the skeleton rather than a new goal: cross-layer consistency, traceability. Or it is a downstream consequence of the four goals being met, not an independent goal: \emph{interpretability} is such a case. If a candidate truly exists that passes the test and is compatible with none of the four goals, what it would require is not a fifth cell but a third discipline axis: the two-axis, four-cell characterization of Section 5.4 would then fail wholesale and have to be rewritten in a higher dimension. Structural corollary I of Chapter 7 takes this as the characterization's falsification condition; the framework bets that no such candidate will appear.

Before proceeding, one same-name affair must be settled with the machine learning literature. The stability--openness problem has a famous predecessor: the stability--plasticity dilemma (Carpenter \& Grossberg, 1987), on which the continual-learning literature still works. But this is the \emph{same name for a different thing}: the "stability" continual learning protects is learned parameters not being washed out by new data, and its "openness" is continuing to eat new data --- an engineering problem inside posterior learning, whose typical solution is to freeze the lower layers and fine-tune. Stability and openness in this paper are construction goals of a prior framework, and we have something they do not: norm promulgation as the source of stability (\allowbreak{}Proposition 4.2). Continual learning defends stability by algorithm; this paper's stability holds by promulgation. Two further lineage relations are not developed here: the relation between the TBox/ABox distinction of description logic and this paper's layer boundary is treated in Section 2.4, R.1; the relation between Rasmussen's abstraction hierarchy and this paper's axes, in Section 2.4, R.7. One sentence is registered here: this paper's layer boundary is not inherited from any existing tradition; it is uniquely drawn by the structure of the construction goals. That is the work of Sections 5.3 and 5.4.

\subsection{5.3 Four goal pairs, four carriers}
With the four goals fixed, the framework's shape can be discussed: \emph{should the goals all be realized in a single carrier, or each in its own place?} We first name the key concepts used in the proof below. Lay the four goals out along two axes, \emph{rate of change} (stability $\times$ openness) and \emph{degree of abstraction} (\allowbreak{}abstraction $\times$ concreteness), and four \emph{cells} result. The content of a cell is called a \emph{goal pair}; the bearer of a cell is called a \emph{carrier}. This section's thesis is one sentence: \emph{each goal pair needs its own carrier; the four carriers are mutually incompatible and cannot be doubled up in one; therefore a prior framework must be divided into layers.}

The two axes are not chosen at will. The origins of the four goals were noted with each paragraph of Section 5.2; here they are gathered into the grounds of the axes. The two poles of the rate-of-change axis come, one, from the definition of promulgation (norms must precede instances and constrain all later instances) and, the other, from the growth of the world --- the object list is never closed. The two poles of the abstraction axis come from the two postures of "constraining instances" (type and clause). Each axis has two poles; two poles times two poles make exactly four cells; "there is no fifth cell" thereby holds automatically and needs no separate defense.

Laying out the four cells only draws the positions; it remains to show that \emph{every cell must have content}: empty cells do not count, and a four-layer skeleton cannot leave one floor empty. The non-emptiness of the cells is supported by four pieces of evidence, all of which have already appeared above and are merely reassembled here. The four conjuncts of goals one and two (parts of speech and combination rules do not drift; knowledge clauses are frozen one by one; the concept type library is augmentable; instance data have landing grounds) are each necessary, as the goal paragraphs of Section 5.2 argued separately, independently of the cells. And which cell each one falls into is not for us to place: it is determined by the nature of the content. Parts of speech and combination rules are types demanding freeze, and can only fall into stable $\times$ abstract; knowledge clauses anchor instances and are frozen one by one, and can only fall into stable $\times$ concrete; augmentation of the concept type library is openness on the abstract side, and the landing of instance data is openness on the concrete side. Each of the four requirements is forced into its own cell by the nature of its content: non-emptiness is the result of occupancy, not a definitional trick. The same fact is double-checked from the negative side: delete any one cell, and exactly one already-argued requirement is violated --- four cells against four requirements, no more, no less. The rigorous statement and proof appear as Lemma 5.1 in Section 5.4.

The four cells each have their owner; in order:

\textbf{Stable $\times$ abstract: the syntax layer, the foundation of the framework.} The grammar of the entire framework: the most basic parts of speech and combination rules. It must be the most stable, because it is the expressive medium of everything built above it; and it is the most abstract, because it denotes nothing concrete. The constancy of the syntax layer is not conservatism; it is the precondition for the whole framework to be \emph{one} system: with the parts of speech and combination rules unchanged, every later addition upstairs forever lands in the same coordinate system. In practice this layer can be astonishingly small (for the syntax vocabulary of an engineering instance, see Appendix B.1 in the supplementary material).

\textbf{Open $\times$ abstract: the concept layer (also called the semantic layer), the home of concepts.} Purpose, function, and failure-mode \emph{types} live in this layer: types, not instances. Its semantics are unique and abstract: each concept has a definite meaning, the same semantics is reused across instances, and experience can therefore accumulate and iterate. It is also open: as the framework's service range expands, new concepts are added. Openness does not threaten stability, because \emph{augmentation is not drift}: the foundation does not move, concepts are only added and never altered, and every addition is forced to land in the coordinate grid the syntax layer has drawn. Nor does augmentation run out of control: take the failure vocabulary as an example (the branch of the type library devoted to "how things break") all failure types are maintained as an enumerated list (for the vocabulary see Appendix B.1 in the supplementary material). One point must be emphasized: the framework's completeness is not bet on the sufficiency of enumeration. The purpose concepts of artifacts will keep expanding with civilization, but this does not compromise the framework's completeness, because the nature of artifacts guarantees that \emph{design comes first}: any new concept is promulgated prior to its instances (\allowbreak{}Proposition 4.1, C1), and the framework can capture it at the point of promulgation. Completeness comes from the order in which promulgation reaches instances, not from the closure of any list.

\textbf{Stable $\times$ concrete: the knowledge layer, the frozen rule documents.} Failure-mode instances, equipment procedures, and operating-condition criteria live in this layer. It is concrete: every rule describes a concrete object or phenomenon. It is also stable --- but this stability rests not on exhaustive enumeration but on \emph{clause-by-clause freezing}: the set of knowledge instances is an open set under continuous augmentation (every newly connected industry or device brings newly defined instances), while any single knowledge instance, once described and defined, becomes a document, invariant across projects, scenarios, and time. The set is forever open and every entry forever unchanged; the set's openness is realized through clause-by-clause issuance events, with no standing channel installed. This cell must be carefully distinguished from the previous one: the most easily confused pair in the chapter: "open" says whether the system can be extended to describe more new things, while "concrete" says whether something is a concept (the commonality of things) or a concrete phenomenon. The concept layer is open and abstract; the knowledge layer is concrete and stable: one governs "what can be said," the other "what has been settled."

\textbf{Open $\times$ concrete: the instance layer, the landing ground of posterior data.} What this layer holds is not the world itself but data describing the world's instances: continuous telemetry, observation records, operating-condition streams --- posterior data enter the system here. It is open, because the real world keeps running and keeps producing new data; it is concrete, because every datum is anchored to a concrete instance. This layer is one working part of the whole intelligent system, and the landing ground of the loop from Chapter 1 and 4.4: the prior starts the system, operation produces data, the data flow back: and the backflow lands here. It is also a kind of structural humility: the upper three layers are all cross-sections of the real world, and only this layer directly touches it; every concept and every piece of knowledge must ultimately be cashed out here, or raise its alarm here.

Why can the four carriers not be doubled up? At the outset we named the bearer of a cell a carrier; now we give it a checkable criterion: \emph{a carrier is a content region with a unified revision discipline and a unified denotation discipline.} A word on how this definition is used: it carries no argument; it merely spreads "carrier" out into two disciplines so that doubling-up can be checked item by item --- the conclusion is not presupposed in the definition. The argument is carried by Section 5.2 (necessity of the goals) together with the non-emptiness of the cells above and the incompatibility check below. Once spread out: \emph{rate-of-change incompatibility}: the stable pole's revision discipline is closure (change only via promulgation events), while the open pole's is a standing intake channel (openness to augmentation as the normal state); the same region cannot both take promulgation events as its only source of change and run a standing intake channel, and one carrier can run only one revision discipline. \emph{Denotation incompatibility}: the abstract pole requires a denotation discipline of open types (\allowbreak{}infinitely instantiable), while the concrete pole requires anchored instances, and one carrier can run only one denotation discipline. Hence: to make syntax open is to let the foundation grow with the building; to make semantics concrete is to lock concepts onto single instances and forfeit reuse; to make knowledge abstract is to strip documents of their anchored objects; to let the framework directly touch the open, concrete instance stream is to retreat to the old road of end-to-end statistical learning (Chapter 1 and 4.4). \emph{Since doubling up is impossible, layering is necessary.}

\subsection{5.4 The layering lower bound: statement and proof}
Section 5.3 gave a corollary: the four carriers are mutually incompatible and cannot be doubled up, so the framework must be layered. This section tightens the corollary into an auditable structural characterization, with a rigorous proof.

\textbf{Structural Characterization 5.1 (the layering lower bound).} A prior cognitive framework satisfying the Constitution Criterion has a number $k$ of non-empty equivalence classes of content with $k \geq 4$; the four equivalence classes respectively realize the goal pairs $(G_1,G_3)$, $(G_2,G_3)$, $(G_1,G_4)$, $(G_2,G_4)$ --- that is, stable $\times$ abstract, open $\times$ abstract, stable $\times$ concrete, open $\times$ concrete.

\textbf{Premise: necessity of the goals (normative).} A qualified framework must realize the four goals (stability, openness, abstraction, concreteness) argued item by item in the goal-ablation table of Section 5.2. The nature of this link is normative argument: "what counts as a qualified prior framework" is a normative judgment, not derivable from pure logic (see the nature statement at the end of this section for what this means).

\textbf{Notation.} Let the total content of the framework be the set $\mathcal{C}$. Two discipline functions are defined on $\mathcal{C}$. The revision discipline $\rho: \mathcal{C} \to \{F, I\}$ takes its value from how the region the content belongs to changes within its established service range: $F$ (freeze): the region's content closes into a set between promulgation events, and any change must go through a promulgation event (a standards succession, the clause-by-clause issuance of articles); $I$ (intake): the region runs a standing intake channel as constitutive equipment (a port of entry for concepts or a landing ground for data), open to augmentation within its established service range. To forestall circularity worries, the test is given entry by entry: $\rho(x) = I$ if and only if $x$ falls within the service range of some standing intake channel, and $F$ otherwise; the determination of a channel's existence takes priority over the manner in which change arrives. Note that entry-level append-only discipline has nothing to do with $\rho$: existing entries of L2, L3, and L4 are never rewritten; what $\rho$ distinguishes is \emph{whether growth has a standing channel as its normal state}: L1 and L3 take $F$ (change only via promulgation), while L2 and L4 take $I$ (the port of entry and the landing ground are standing equipment). The denotation discipline is $\delta: \mathcal{C} \to \{T, N\}$: $T$ denotes types (symbols openly instantiable, anchored to no concrete object); $N$ denotes instances (content anchored to concrete objects or phenomena --- anchoring via object-class codes is allowed, as B.1's knowledge entries anchor object classes via hazardEntityCode, rather than denoting abstract types).

\textbf{Definition 5.1 (carrier).} Define an equivalence relation on $\mathcal{C}$ by $x \sim y \iff \rho(x)=\rho(y) \land \delta(x)=\delta(y)$; its equivalence classes are called \emph{carriers}. A carrier is the rigorous form of what Section 5.3 called "a content region with a unified revision discipline and a unified denotation discipline." The number of layers $k$ of the framework is defined as the number of non-empty equivalence classes. (Numbering convention: L1--L4 are the syntax, concept, knowledge, and instance layers respectively; the appendices use this coding throughout.)

\textbf{Definition 5.2 (\allowbreak{}construction goals).} A qualified framework must satisfy:

\begin{itemize}
\item \textbf{G1 (stability)}: content with $\rho=F$ exists at both poles $\delta=T$ and $\delta=N$: parts of speech and combination rules do not drift, and knowledge clauses are frozen one by one (\allowbreak{}existential reading: a framework whose clauses are all rewritable cannot vacuously satisfy this goal);
\item \textbf{G2 (openness)}: content with $\rho=I$ exists at both poles $\delta=T$ and $\delta=N$: the type library has a port of entry for new concepts, and instance data have landing grounds (the existence of passages and disciplines is required; the actual occurrence of intake events is not);
\item \textbf{G3 (\allowbreak{}abstraction)}: content with $\delta=T$ exists: constraining instances as types, so that experience accumulates across instances;
\item \textbf{G4 (\allowbreak{}concreteness)}: content with $\delta=N$ exists: anchoring concrete instances, supporting on-site judgment.
\end{itemize}
\textbf{One honest registration.} The conjunctive structure of G1 and G2 carries the whole Cartesian product: G1 $\wedge$ G2 alone entails that all four cells are non-empty, and G3 and G4 formally add nothing to the lower bound. Their duty lies elsewhere: naming and justifying the $\delta$ axis itself. The two postures of "constraining instances" (goals three and four of Section 5.2) show that denoting types and denoting instances each carry an irreplaceable function, so the contents of the four cells are not mergeable duplicates; without this justification, the introduction of the $\delta$ axis would be arbitrary.

\textbf{Lemma 5.1 (cells are non-empty).} Under G1--G4, all four combinations $(F,T)$, $(I,T)$, $(F,N)$, $(I,N)$ are occupied.

\textbf{Proof.} The two conjuncts of G1 directly give $(F,T)$ and $(F,N)$ non-empty; the two conjuncts of G2 directly give $(I,T)$ and $(I,N)$ non-empty. G3 and G4 justify that the identities of the four cells' contents cannot be merged: constraining as types and anchoring as instances are two irreplaceable functions (Section 5.2, goals three and four), so the four cells are four positions with their own duties, rather than four names for the same content. The cell-ablation recheck of Section 5.3 verifies the same fact from the negative side: deleting any one cell violates one independently argued requirement. \hfill$\square$

\textbf{Proof of Structural Characterization 5.1.} By Lemma 5.1, each of the four cells has content; by Definition 5.1, one equivalence class carries only one pair $(\rho, \delta)$, so the four cells belong to four distinct equivalence classes. For contradiction, suppose $k < 4$; then by the pigeonhole principle, at least one equivalence class must contain the content of two cells, among which there must be two entries differing in $\rho$ or $\delta$ --- contradicting the definition of an equivalence class. The six pairwise mergers of the four cells are checked one by one below; none survives:

\begin{table}[htbp]
\centering
\small
\begin{tabular}{@{}>{\raggedright\arraybackslash}p{0.306\textwidth}>{\raggedright\arraybackslash}p{0.306\textwidth}>{\raggedright\arraybackslash}p{0.306\textwidth}@{}}
\toprule
Merged cells & Shared discipline & Conflict \\
\midrule
$(F,T)$ and $(I,T)$ & $\delta=T$ & $\rho$: freeze vs intake \\
$(F,N)$ and $(I,N)$ & $\delta=N$ & $\rho$: freeze vs intake \\
$(F,T)$ and $(F,N)$ & $\rho=F$ & $\delta$: type vs instance \\
$(I,T)$ and $(I,N)$ & $\rho=I$ & $\delta$: type vs instance \\
$(F,T)$ and $(I,N)$ & none & both $\rho$ and $\delta$ \\
$(I,T)$ and $(F,N)$ & none & both $\rho$ and $\delta$ \\
\bottomrule
\end{tabular}
\end{table}
\hfill$\square$

Finally, the nature of this characterization. The proof itself is rigorous, but its premises are not all delivered by logic: why each of the four goals is necessary is argued paragraph by paragraph in Section 5.2, and that is judgment about what counts as a qualified prior framework, not logic. Strictly speaking, the formal lower bound is derived from the conjunctive structure of G1 $\wedge$ G2 (see the registration after Definition 5.2); the necessity of G3 and G4 shows up not in the counting but in the justification of the $\delta$ axis. The complete meaning of this characterization is therefore a conditional: \emph{if you grant that each of the four goals is necessary, the layered skeleton is necessary}; the reader who disputes any single goal should return to Section 5.2 and take it up item by item. A proposition of this kind (the premise is judgment, the proof is logic) is what we call a normative architecture proposition.

One clarification. Strictly speaking, since $\rho$ and $\delta$ are both two-valued disciplines, there are at most four equivalence classes, and what this characterization formally yields is exactly $k = 4$ (the nature statement of Appendix A.7 (\allowbreak{}supplementary material) registers this in full). The text says "lower bound," and that semantics targets two things: an engineering implementation may subdivide a layer further (splitting the knowledge layer into industry sublayers, for example), and any compliant subdivision can only increase the region count, so $k \geq 4$ holds for all subdivision partitions; and if a third discipline axis appears in the future, the skeleton expands accordingly (that would be a generalization of the characterization, not the present situation). The skeleton cannot be compressed further; the flesh can be divided further. For its testability, see structural corollary I in Chapter 7.

\subsection{5.5 What layering buys: from decidability to auditable intelligence}
The layering lower bound constrains the shape of the skeleton. But layering is not merely an obligation; it buys back a property that end-to-end learning cannot currently offer: the decidability of fault localization.

\textbf{Proposition 5.2 (\allowbreak{}decidability of fault reduction over a closed concept layer).} Suppose the concept layer, under any current promulgated version $C_t$, is a finite closed set --- failure-mode types, environmental factors, and system-transfer factors are all maintained as enumerated lists (see Appendix B.1 in the supplementary material); "closed" everywhere means closed relative to the current version, while across versions the concept layer remains open through augmentation (see Remark A.18), and suppose reduction proceeds level by level along the collapse chain. Then fault localization is decidable in polynomial time: the depth of the reduction path is bounded by the depth $d$ of the collapse chain inside the concept layer ($d \leq 4$, promulgated as an upper bound; see Hypothesis H4 in Appendix A.3 of the supplementary material), and the number of candidate tests at each level is bounded by $|C|$. The reduction cost is therefore $O(d \cdot |C|)$.

\textbf{Proof.} Reduction starts from one alarm at the instance layer and \emph{scans candidates level by level} rather than branching along paths: at each level it performs one set-level covering test (covers, $O(1)$ on the current node set as a whole; for the algorithm and the interface assumption see Algorithm 1 and H3 in Appendix A.3 of the supplementary material) for each candidate in $C$ --- $|C|$ tests per level, $d$ levels in total, for an overall cost of $O(d \cdot |C|)$. Note that this bound depends on the set-level indexing of the covering test: if covers had to traverse the growing node set element by element, the bound would degrade to $O(d \cdot |C|^2)$; the decidability conclusion is unaffected, while the cost magnitude is. \hfill$\square$

\textbf{Remark (the promulgated source of d, an example).} The number of levels and the enumeration at each level are promulgated by the concept layer: e.g., the four levels promulgated by hazardEntityScope --- medium unit under equipment $\to$ component under equipment $\to$ entity combination under equipment $\to$ the equipment under analysis (see Hypothesis H4 in Appendix A.3 and Appendix B.1 of the supplementary material; the drill feed mechanism of Appendix B.2 has d = 3, per H4's upper-bound convention).

The control group of this proposition deserves serious discussion. A world without a closed concept layer offers no such guarantee: reduction may extend indefinitely, and localization is at best semi-decidable. More fundamentally, a model that commands only distributional similarity can report that an input is "unlikely," but cannot decide whether the input \emph{violated a norm} or \emph{instantiated a novel but legitimate behavior}. For such a model, "is this a new type of fault?" is not a question answered wrongly; it is a question asked wrongly. The distinction is not academic: the two situations call for opposite handling --- one should be learned, the other must be stopped. A closed concept layer turns this distinction into something decidable: for an alarm i, either it reduces to some known type in $C$ (\allowbreak{}localization succeeds; whether the localization constitutes a violation is separately adjudicated at the knowledge layer, by the normative status of the located type), or reduction fails (the current concept layer does not cover it: failed reduction proves non-coverage only (under the chain-continuity assumption H5; see Appendix A.3 in the supplementary material), not the existence of a new type in the world; whether it constitutes a new concept is referred to the port of entry of Section 5.2, to be adjudicated by the concept-augmentation procedure). Decidability and the openness goal converge here: what is closed is the \emph{types}; what is open is the \emph{instances and the augmentation passages}.

\textbf{Corollary 5.1 (the representational form of institutionalized intelligence: a sufficient chain and the cost of its absence).} The three requirements institutional settings place on intelligence were registered in Corollary 4.2: norm accessibility (I1), violation audibility (I2), accountability closure (I3). I2 requires that deviation events be classifiable as Violation rather than mere prediction error; by Definition 4.3, this requires the Violation judgment to be decidable; that judgment decomposes into two steps: reduction localization, and the normative-status adjudication of the located type. The former is supplied by Proposition 5.2 (decidable over a closed concept layer); the latter is a table lookup under the knowledge layer's graded semantics (decidable). By Structural Characterization 5.1, the skeleton of a framework satisfying the criterion has no fewer than four layers. Hence a four-layer skeleton satisfying the criterion \emph{supplies} the adjudicative capacity that institutionalized intelligence needs: on such a skeleton, a system does not merely report anomalies --- it can issue violations. This corollary does not claim uniqueness: what it registers is one sufficient chain from institutional requirements (I1--I3) to the skeleton. Its necessity side is empirically supported by the failure modes where the chain is absent: systems that do not meet the layering lower bound have so far been able at most to report anomalies, never to issue violations (the ODD paragraph of Section 5.6 is exactly this negative control).

Why extraction cost varies from world to world (the correspondence between the readability gradient and extraction complexity, Proposition A.19 in Appendix A.4 (\allowbreak{}supplementary material)) and why archiveless worlds resist rule coverage (the non-identifiability boundary, Proposition A.13 in Appendix A.2) are two supporting results of the same family as Proposition 5.2, and are stated in the appendix.

\subsection{5.6 External witnesses: isomorphism and its price}
Structural Characterization 5.1 is a requirement on frameworks, not a description of the world: whether real-world systems also look like this must be checked against real systems. The methodological status of this section must therefore be stated openly: this is not a display of application cases but a \emph{natural experiment about the number of layers} (in the loose sense: methodologically it is comparative structural evidence across systems, not a natural experiment in the causal-identification sense). The basis for choosing the objects of verification is laid out in the open: first, the mainstream industry standards most relevant to this paper's domain and with the largest installed base (BACnet and LonWorks, with Modbus and the two generations of OPC as controls); second, a parallel witness from another industry (the ODD practice of autonomous driving); third, the latest independent research result from AI (the category-theoretic COIN construction); fourth, a promulgation-based semantic layer self-organized by an open-source community (Brick and Haystack in the building industry) and the semantic-web lineage at general Web scale (RDF and OWL). The phrase "four layers" appears in none of their self-descriptions; all the reverse readings below are conducted through the lens of this framework, and every figure, promulgator, and layer-by-layer alignment table is as recorded in Appendix C (\allowbreak{}supplementary material).

\textbf{Isomorphism.} Four lineages that do not know one another have each independently grown the same skeleton: syntax, concepts, knowledge, instances, each in its place.

\textbf{BACnet}: the syntax is the "object--property--service" metamodel --- everything in the world presents as an object, objects are described by properties, actions are initiated through services. The concepts are the closed set of standard object types, versioned and promulgated by the committee, only added and never altered. The knowledge is the conformance specification of device behavior (BIBB capability blocks and PICS declarations) by which "how a device should respond" becomes a testable engineering fact. The instances are the live objects inside devices, addressed by type plus instance number, directly grounded to sensors.

\textbf{LonWorks}: the syntax is the network-variable type system together with the device interface file (XIF): every device self-declares "what I can say." The concepts are the SNVT master list promulgated by LonMark, with promulgation granularity down to the invalid value of each type. The knowledge is the Functional Profiles, promulgating the variable sets that each class of functional node shall expose, divided into mandatory and optional. The instances are the network variables whose commissioning is complete and whose bindings are ready.

\textbf{COIN}: a purely mathematical construction --- monad composition laws as syntax, type disciplines as concepts, certified blueprints as knowledge, leaf contracts grounding instances; the complete inventory is in Appendix C.6 (\allowbreak{}supplementary material). It is the only witness with no engineering archive at all; that even it lands on four layers deprives the explanation "four layers are just engineering convention" of its last refuge.

\textbf{Brick / Haystack}: a promulgation-based semantic layer governed by an industry open-source community. Haystack legislates for data points with a community-governed controlled tag vocabulary; Brick builds a class-hierarchy ontology on top of it (tagsets corresponding to classes) and promulgates machine-checkable clauses about "which relations an entity shall have" through SHACL shape constraints (Balaji et al., 2016). The syntax is RDF triples; the instances are the entity graphs of concrete buildings; ASHRAE 223 (in preparation) is advancing this lineage into a formal standard (Appendix C.7, supplementary material). No one in this industry was persuaded by a theory; they were driven by the economics of integration: when a domain's data cannot interoperate without shared semantics, the domain invents shared semantics: promulgated, versioned, committee-maintained semantics.

The four promulgators do not know one another, and the landing point is the same skeleton (\allowbreak{}Appendices C.1--C.3, C.6, C.7).

\textbf{The price.} More persuasive still is the failure record of those missing a layer or mixing layers.

\textbf{Modbus}: only two layers --- syntax (function codes and register addresses) and instances (current register values). The concept and knowledge cells hang empty; the semantics of registers have nowhere to be promulgated and drift into vendor documents, and integration still depends on manual mapping to this day (Appendix C.4, supplementary material).

\textbf{OPC}: a rare within-group control. DA (1996) stopped at two layers (syntax (tag read/write) and instances (tag values)) with concepts and knowledge absent. UA (from 2008) filled in these two layers: the information model as concepts, the Companion Specifications as knowledge --- and only thereafter became the backbone of Industry 4.0 (Appendix C.3, supplementary material).

\textbf{ODD} (\allowbreak{}Operational Design Domain for autonomous driving): the problem is not a missing layer but mixed layers. "Open abstraction" (scenario concepts) and "stable concreteness" (operating-condition clauses) are merged into one declaration carrier, and the two disciplines interfere with each other; the price is that declarations from different manufacturers are mutually incomparable and have never been horizontally auditable (Appendix C.5, supplementary material). This is the negative control of Corollary 5.1: the direct price of mixing layers is not performance but unauditability. ODD is also the only witness that admits a repair-direction prediction: the four-layer framework happens to provide separate homes for the two kinds of content that were merged --- abstract scenario classification moves up to the concept and knowledge layers, and concrete operating boundaries move down to the knowledge and instance layers (Appendix C.5, supplementary material).

\textbf{RDF / OWL} (the semantic-web lineage): the largest round of "framework first" experimentation at general Web scale. RDF is another specimen of two-layer "syntax + instances" success --- same form as Modbus, scaled up to the whole Web. OWL adds the TBox concept layer, but the knowledge layer remains absent throughout: description-logic axioms are logical constraints on concepts, not promulgated norms about "how instances shall be and what deviation counts as." The form its price takes: plenty of ontologies were built; cross-domain auditable operational adjudication never appeared (Appendix C.8, supplementary material). For the conceptual analysis of "whether TBox/ABox are two of this paper's layers," see Section 2.4, R.1.

Thirty years of industrial history thus constitute the complete design of a natural experiment: whichever layer is missing, the price falls precisely on that layer; whenever the layer is filled in, the price disappears; whichever two cells are mixed, the interference appears precisely between the two corresponding disciplines.

\subsection{5.7 The closure of the argument: the world is pre-layered by design}
The layering lower bound has a geometry, and it deserves one final articulation, because it welds Chapters 3 through 5 into a single whole. The artificial physical world is \emph{the only world that natively spans the whole goal map}: the archive lives on the "stable $\times$ abstract" side, the instances live in the "open $\times$ concrete" corner, and the two middle cells are filled by object types and procedural clauses. Every other world must learn bottom-up from instances; the designed world alone is pre-layered when it leaves the factory. The precise meaning of the phrase "comes with its source code" is here: \emph{in this world's source code, the directory structure of syntax, semantics, and instances is ready-made}: vocabulary standards are promulgated first, then object types defined, then procedural clauses promulgated, and only then are instances built and operated; the four layers are the sedimentation in the archive of the four stages of the design process. This is where the criterion (Chapter 4) and the characterization (Chapter 5) close up: the criterion says this world \emph{deserves} a framework erected first; the characterization says the framework must divide into \emph{at least} four layers; and layering is cheap because the world itself is pre-layered by design. It is built in layers, and so it is archived in layers.

\textbf{Chapter summary.} This chapter moved from the criterion to the skeleton: the shape of a prior framework satisfying the Constitution Criterion is uniquely determined by four construction goals. Stability and openness issue from the two temporal postures of promulgation and from the growth of the world; abstraction and concreteness issue from the two denotational postures of "constraining instances." The four goals pair up into four cells; each cell needs its own carrier; the carriers are mutually incompatible; hence layering is necessary (\allowbreak{}Structural Characterization 5.1: $k \geq 4$). Layering is not merely an obligation: a closed concept layer makes fault reduction decidable in polynomial time (\allowbreak{}Proposition 5.2) and turns the "violation versus novelty" distinction into an operable judgment. Layering is thereby upgraded from a technical property to an institutional one: the four-layer skeleton supplies the adjudicative capacity auditable intelligence needs, and the failure modes where it is absent empirically support its necessity (Corollary 5.1). Thirty years of industrial history provide a natural experiment about layer count: four independent lineages converge on the same skeleton, and the prices of missing or mixed layers land precisely on the corresponding cells. The closure thereby becomes visible: the world is built in layers, so it is archived in layers, and the framework need only copy the world's generative order. The skeleton stands. Once the skeleton runs online, it is no longer omnipotent: how duties at runtime are to be divided is the business of the next chapter.

\section{Chapter 6. Working with LLMs: The Promulgation Criterion as the Division-of-Labor Interface}
\begin{quote}
Chapter 5 left a question at its end: when the skeleton goes online and runs, how are duties discharged --- which are borne by the rule engine, which by the large language model? This is also the most natural external question: what is the relation between this prior framework with its reasoning engine and today's LLMs? The answer comes in four steps: the prior framework honestly declares what it does not do (Section 6.1); the principle dividing runtime duties is given by this paper's own criterion (Section 6.2); the division principle is then instantiated in two typical circuits (Sections 6.3 and 6.4); and finally a conclusion belonging to this paper itself is drawn: sandwiched generation is auditable generation (Section 6.5).
\end{quote}
\subsection{6.1 What the prior framework does not do}
Chapters 3 through 5 proved that the artificial physical world can be promulgated, layered, and adjudicated: the criterion (Chapter 4) delimited the object domain of legitimate priors, and the layering (Chapter 5) gave the minimal structure that carries them. But all of this belongs to the framework's production and promulgation stages. At the stage when the system runs online (\allowbreak{}henceforth \emph{runtime}), this entire body of proof has an equally important other half: \emph{what the prior framework does not do}.

The prior framework does not do three things. First, it does not reduce live physical scenes into symbols in real time: readings enter the knowledge layer only through pre-promulgated slots and reduction formulas (the DIP/SEE chain of Appendix B.1, supplementary material); for a physical phenomenon not covered by any promulgated slot, the prior framework correctly reports "unknown," but it will not go and look for itself. Second, it does not generate hypotheses that had no coordinates at promulgation time: candidate enumeration can only proceed within promulgated boundaries (Section 5.2, Proposition 5.2); hypotheses outside the boundary (a failure mechanism never foreseen by any clause) are not in the enumeration space. Third, it does not exercise on-site discretion among multiple objectives: when three promulgated objectives (comfort, energy consumption, grid safety) conflict at the same moment, the promulgation of the objectives themselves does not prescribe which one leads right now.

This gap is not a defect but a direct corollary of the criterion: only the promulgatable has a place in the framework, and duties that were not promulgatable at promulgation time need another kind of cognitive entity to bear them. Today's machine learning happens to offer a candidate: the representational capacity that LLM-type carriers have acquired on the artificial symbolic world enables them to handle the on-site judgments that were "not promulgatable at promulgation time." (Note: "LLM" in this chapter names the functional role of a posterior carrier bearing the "on-site judgment beyond promulgation"; it does not presuppose the concrete implementation of a general-purpose large language model. In particular physical domains, the role can equally be borne by models tailored and cut for that domain.) This chapter's thesis is: \emph{the prior framework and the LLM are not competitors but two trades divided along one and the same criterion line} --- a line given by the criterion of Chapter 4, not by engineering taste.

\subsection{6.2 The division principle: duty assignment is uniquely determined by the Promulgation Criterion}
Pass every runtime duty through the criterion C1/C2 and the working assumption A3: what can be promulgated (\allowbreak{}propagation along the promulgated causal chain, candidate enumeration within promulgated boundaries, adjudication by promulgated evaluation rules) goes to the rule engine; what could not be promulgated at promulgation time (\allowbreak{}understanding of the live scene, generation of hypotheses, construction of counterfactual trajectories) goes to the LLM-type carrier. Making A3's coverage test duty-by-duty (inside coverage, to the engine; outside coverage, to the LLM) is this section's division principle; in other words, the division principle is the projection onto runtime of the full antecedent of Proposition 4.2 (C1 $\wedge$ C2 $\wedge$ A1 $\wedge$ A2 $\wedge$ A3 $\wedge$ A4). This division is complete and disjoint over runtime duties: at promulgation time, every duty either already had coordinates or it did not; there is no third state. (Human takeover counts as a sub-channel of the non-promulgatable side: the final discretion goes to the LLM or to a human, without disturbing the completeness of the dichotomy.)

Both ends of this division already have their landing places in this paper. The engine end is everything demonstrated in Appendices B.1 and B.2: the monitoring-and-judgment chain, cause-tracing enumeration, and degree-level write-back, all composed of promulgated entries, checkable clause by clause at every step. The LLM end fills in the engine's two structural blind spots: the reduction blind spot (the field beyond the slots) and the enumeration blind spot (\allowbreak{}hypotheses beyond the promulgated boundary).

Only one sentence needs emphasis: \emph{whether a duty goes to the engine or to the LLM depends not on technology selection, only on whether it already had coordinates at promulgation time.} The Promulgation Criterion is the core result of Chapter 4, so the division principle given here is not borrowed engineering experience but the criterion's projection onto runtime. According to which side sits in the outer layer commanding and which side is invoked, two complementary forms of collaboration arise, henceforth called the "shell" and the "filling."

\subsection{6.3 Form one: diagnosis-as-shell (LLM as shell, engine as filling)}
First, the instantiation of the division principle in the diagnosis circuit. The complete diagnosis circuit after an abnormal operating condition arrives has five links: is this anomaly worth handling (\allowbreak{}prioritization); what might be the cause (\allowbreak{}hypothesis generation); what evidence to look at next (\allowbreak{}observation selection); once evidence arrives, reduction flows back and consequence evaluation propagates upward along the causal chain; and finally, what should be done (plan synthesis).

Applying the principle of Section 6.2 link by link: of the five links, \emph{the fourth (reduction backflow and upward consequence evaluation) is promulgatable}. Reduction backflow is the DIP $\to$ SEE $\to$ LTT $\to$ JRI monitoring-and-judgment chain most fully demonstrated in Appendix B.1 (\allowbreak{}supplementary material); upward consequence evaluation is propagation from a phenomenon along promulgated causal chains toward consequences (the CQ-P1 $\to$ SS-C1 channel of Appendix B.1, supplementary material, is exactly this chain). This link is the rule engine's proper office, invoked inside the circuit as a subroutine. The other four links (\allowbreak{}prioritization, hypotheses, observation, plan synthesis) all had no coordinates at promulgation time: prioritization depends on what other equipment is doing right now; the hypothesis space exceeds the promulgated boundary; observation selection depends on evidence that has not yet arrived; plan synthesis must weigh constraints that exist only at the moment. These four links go to the LLM.

The form of the diagnosis circuit is thus: LLM as shell, engine as filling. Command of the circuit stays with the on-site judgment side from start to finish, and the promulgated rule system is invoked inside the circuit. This is isomorphic to what industry calls the ReAct-style agent loop (Yao et al., 2023), but what this paper gives is its criterion basis: which link of the loop can be replaced by a rule engine is decided by whether that link is promulgatable, not by "whether the LLM is strong enough."

\subsection{6.4 Form two: feedforward-as-shell (engine as shell, LLM as filling)}
The feedforward circuit gives the reverse instantiation of the same division --- it is the mirror image of diagnosis. Take a cooling plant: the forecast shows the midday peak will exceed the plant's cooling capacity; should cooling output be increased now, storing cold in the building mass, trading time for capacity?

The problem splits into three layers: the candidate policy space, the evaluation rules, and the parameter trajectory. The first two layers are promulgatable and go to the rule engine: the multi-objective evaluation clauses for comfort, energy, and grid safety are written in the operating procedures; candidate policies are written within the boundaries promulgated by the design documents --- policies outside the boundary have no engineering meaning. Enumerate the in-boundary policies, evaluate them one by one by the promulgated rules; the whole process is auditable. Note that the evaluation clauses give only an ordering; the final discretion (where the ordering ties or the clauses do not cover) still goes to the LLM or to a human. This is where the third thing of Section 6.1 sits in this example. The third layer is not promulgatable and goes to the LLM: under the chosen policy of "cold storage," the trajectory along which the coupled system's physical parameters (water temperature, flow rate, room temperature) should run is recorded in no archive. The policy is new, so the trajectory is new; the LLM takes the policy as its goal, generates the trajectory the parameters should follow, and hands it back to the promulgated evaluation rules for violation pre-screening. (Note: the author's team already has an implementation of this kind, tailored and cut for the building-thermodynamics domain; its multi-dimensional evaluation is to be reported in a companion paper.)

The form of feedforward is thus: engine as shell, LLM as filling. Enumeration and evaluation sandwich the trajectory generation in the middle. Figure 6-1 shows the process difference between the two forms: in the diagnosis circuit, the LLM sits at the periphery invoking the engine subroutine; in the feedforward circuit, the engine sits at the periphery sandwiching the LLM's trajectory generation --- the assignment swap is decided by the same criterion.

\begin{figure}[htbp]
\centering
\begin{tikzpicture}[
  font=\scriptsize,
  node distance=0.45cm,
  box/.style={draw, rounded corners=2pt, align=center, inner sep=3pt, text width=2.35cm},
  grp/.style={draw=gray, dashed, inner sep=6pt, rounded corners=3pt},
  glab/.style={font=\scriptsize\itshape, text=gray},
  arr/.style={-{Stealth}, thick}
]
\node[box] (P1) {Prioritization: worth handling? (LLM)};
\node[box, right=of P1] (H1) {Hypothesis generation (LLM)};
\node[box, right=of H1] (O1) {Observation selection (LLM)};
\node[box, right=of O1] (U1) {Reduction \& upward consequence evaluation (engine, invoked)};
\node[box, right=of U1] (PL1) {Plan synthesis (LLM)};
\draw[arr] (P1) -- (H1);
\draw[arr] (H1) -- (O1);
\draw[arr] (O1) -- (U1);
\draw[arr] (U1) -- (PL1);
\node[grp, fit=(P1)(H1)(O1)(U1)(PL1), label={[glab]above:{Form 1: Diagnosis (LLM shell \textperiodcentered{} engine filling)}}] (DX) {};
\node[box, below=1.6cm of H1] (E1) {Enumerate in-boundary candidate policies (engine)};
\node[box, right=of E1] (G1) {Counterfactual trajectory generation (LLM)};
\node[box, right=of G1] (E2) {Violation pre-screening by promulgated rules (engine)};
\draw[arr] (E1) -- (G1);
\draw[arr] (G1) -- (E2);
\node[grp, fit=(E1)(G1)(E2), label={[glab]above:{Form 2: Feedforward (engine shell \textperiodcentered{} LLM filling)}}] (FF) {};
\end{tikzpicture}
\par\vspace{4pt}\noindent{\small\textbf{Figure 6-1.} The two shell-and-filling forms. The assignment of each link is decided by the Promulgation Criterion: promulgatable links go to the engine; non-promulgatable links go to the LLM.\par}
\end{figure}
\subsection{6.5 Closing: sandwiched generation is auditable generation}
At the close, one consequence belonging to this paper itself should be spelled out.

Industry's greatest misgiving about LLMs entering physical facilities is "hallucination": free generation takes likelihood as its only criterion, and the text flows wherever it flows. In the two forms of this chapter, generation is never free: wherever a position is sandwiched by promulgated clauses, it is sandwiched upstream by the promulgated boundary (\allowbreak{}candidates must lie within the boundary, the goal must be a promulgated policy) and downstream by promulgated evaluation (the output must pass pre-screening against the clauses). When sandwiched generation goes wrong, it goes wrong by violating a specific clause, not by deviating from an invisible distribution. That is, \emph{for every generation position sandwiched by promulgated clauses, failure falls under S3: it is loud.} The free judgments at shell positions (\allowbreak{}prioritization, hypotheses, and plans in the diagnosis form) do not fall under S3, since no promulgated clause stands downstream to check them against; but this is the other side of the same coin: what can be promulgated decides both duty assignment (Section 6.2) and failure mode, and the shell positions must be borne by the LLM precisely because no promulgatable clause exists there to be loudly violated. In the artificial physical world, the hallucination problem is partially converted into an audit problem. This is not an improvement of the LLM; it is the equal constraint a promulgative environment places on every cognitive entity --- on the rule engine so, and on the LLM so.

This chapter has argued for the division principle and the existence of the two forms; why LLM-type carriers are a necessity for bearing the "on-site judgment beyond promulgation," and under what conditions their bearing is sufficient --- the full arguments for these questions exceed this paper's scope and are handled in a companion paper.

\subsection{Chapter summary}
This chapter has converted "the relation between prior frameworks and LLMs" from an external challenge into a corollary of this paper's criterion: the prior framework handles promulgatable duties, the LLM handles on-site judgments beyond promulgation, and the dividing line is uniquely determined by the Promulgation Criterion. The two shell-and-filling forms (diagnosis and feedforward) are mirror images, with shell and filling assignments swapping with what can be promulgated. Generation positions sandwiched by promulgated clauses inherit S3's loud-failure property; free judgment at shell positions does not fall under S3, and this is explained by the same criterion --- what can be promulgated decides duty assignment and failure mode at once, so plugging LLMs into this domain does not impair the auditability established in Chapter 4. This division structure is itself a testable bet: P3 and P4 bet on it, and together with the other bets it is put on the table in the next chapter. The necessity of the carrier and the conditions of its sufficiency are left to a companion paper.

\section{Chapter 7. Falsifiable Predictions: Putting the Theory's Bets on the Table}
A story that only recounts the past is history, not a theory. If this paper's framework is to deserve the name "research programme," it must say things whose failure would damage the framework. Below are four predictions, ordered from looser to stricter, each given in a uniform format: prediction; theoretical basis; falsification condition; what component is damaged if wrong. Two structural corollaries are appended at the end of the chapter --- they are not bets but direct unfoldings of the criterion.

Before the betting opens, the structure of the stakes must be stated. All the bets placed rest on a single thing at stake: whether norms are promulgated is the discriminating variable of prior legitimacy (\allowbreak{}Proposition 4.1). The four predictions are projections of this single stake in different directions: P1 toward history and the future (the ordering power of the constraint axis); P2 toward the criterion itself (both sides of the win-loss map); P3 and P4 toward the two strongest posterior routes of the moment: VLA and LLMs. The two structural corollaries at the end of the chapter are the stake's structural-level equivalents: corollary I bets that the list of four construction goals (G1--G4 of Section 5.2) is complete and irreducible; its failure is the failure of the four-layer lower bound (\allowbreak{}Structural Characterization 5.1), so "four layers cannot be compressed" is not a slogan but the hardest entry among this chapter's bets. The direction of damage propagation is thereby explicit: if P1 fails, the constraint axis of Chapter 3 is hurt; if P2 fails, the criterion of Chapter 4 is hurt; if P3 or P4 fails, the complementarity structure of Chapter 6 is hurt; if corollary I fails, the layering characterization of Chapter 5 is hurt; and if the criterion is falsified by the negative side of P2, everything at stake is lost.

\textbf{P1 (the order of breakthroughs)}

\begin{itemize}
\item \textbf{Prediction}: AI capability breakthroughs will continue to advance along the constraint axis: after the symbolic world, the next large-scale breakthrough occurs in the artificial physical world, ahead of the basic physical world and the open phenomenal world. ("Large-scale breakthrough" means physical manipulation capability that is stable, repeatable, and non-demonstrational across multiple independent tasks and deployment environments --- not a single benchmark or demo.)
\item \textbf{Theoretical basis}: the historical correlation between the constraint axis and the order of breakthroughs (Sections 3.3, 3.4), and the artificial physical world's distinctive position on the three characterizing variables (Section 3.3).
\item \textbf{Falsification condition}: if the general world-model route (trained primarily on open phenomenal video) achieves scalably replicable success in physical manipulation before any archive-driven route does (for example, a system without semantic grounding reaching industrial-grade reliability in unstructured open environments) this prediction fails.
\item \textbf{What is damaged if wrong}: the explanatory power of the constraint axis (Section 3.4) is severely damaged, and the ordering argument of Chapter 3 goes back for wholesale repair.
\end{itemize}
\textbf{P2 (the win-loss map of priors)}

\begin{itemize}
\item \textbf{Prediction}: in design domains where data are scarce and archives readable, the prior-framework route will hold a systematic advantage over purely posterior routes under cold-start conditions; in archiveless domains, the prior-extraction route necessarily loses.
\item \textbf{Theoretical basis}: the Legitimacy Criterion and the readability gradient (Sections 4.2, 4.3); the negative side is supported by the non-identifiability boundary of A.13.
\item \textbf{The information-theoretic half-sentence and the empirical half-sentence}: the information-theoretic half has been supplied by Theorem A.26 (Appendix A.6, supplementary material): sample complexity separates into three tiers (no finite bound without labels, $\Theta$(k/q) with sparse authoritative labels (by Criterion Alpha, defined at Theorem A.26(b)), and 0 instance samples with the complete archive) so the cold-start gap between the archive side and the posterior side is structural. The empirical half-sentence is left for testing: that this structural gap is not in fact ground away by data engineering and model scale in engineering deployments. The form of this test follows the evidence policy of Section 1.4: the public engineering record (whether deployments in archive-readable domains are systematically archive-driven), not a laboratory benchmark --- the two routes are not competing implementations of one task specification, and no shared task bench exists.
\item \textbf{Falsification condition}: a systematic counterexample in either direction: a purely posterior method matching the cold-start performance of a prior framework in some design domain with complete archives and scarce data (such as facility operations); or a route relying purely on hand-written content rules, lacking an independent normative source, reviving in an archiveless domain.
\item \textbf{What is damaged if wrong}: the criterion itself (\allowbreak{}Proposition 4.1) --- the central load-bearing wall of Chapter 4.
\end{itemize}
\textbf{P3 (the institutional ceiling of VLA without constitutive semantic grounding)}

\begin{itemize}
\item \textbf{Prediction}: vision--language--action models (VLA) without semantic grounding (Brohan et al., 2023; Black et al., 2025) hit a ceiling in institutionalized operations scenarios: demonstrational progress continues; institutional adoption stays absent.
\item \textbf{Theoretical basis}: VLA faces the open phenomenal world, and its priors are black-box and non-normative (Chapter 2; Chapter 8, O1); institutional scenarios require intelligence that is auditable, endorsable, and whose failures are reportable (S3 of Proposition 4.2 and I3 of Corollary 4.2). Cash-out already has a forerunner: the autonomous-driving industry's scaled-deployment form is exactly VLA-class systems wrapped in promulgated safety-domain declarations (ODD) --- perception and decision are borne by end-to-end learning, the operating boundary is clamped by a promulgated artifact, and institutional adoption has never occurred in framework-free form; the mixed-layer price of ODD is registered in Section 5.6 and Appendix C.5 (\allowbreak{}supplementary material), and here we take only the side where the promulgated boundary exists.
\item \textbf{Falsification condition}: a VLA-class system achieves responsibility-endorsed deployment at scale in institutional scenarios such as facility operations or industrial inspection without connecting to any constitutive semantic layer. The win-loss rule must be written out here, because this prediction has a position easily misread as a loss: if VLA gains institutional adoption \emph{by connecting to} a prior cognitive framework, that is not this prediction's failure but one of the ways it is cashed out. The outcome does not turn on whether VLA is used; it turns on whether VLA can obtain endorsement on its own. If a constitutive semantic layer stands behind a networked VLA, what stands there is still our theory.
\item \textbf{What is damaged if wrong}: the normative argument (Section 4.4) and the engineering relevance of the division principle of Chapter 6.
\end{itemize}
\textbf{P4 (the irreplaceability of the ontology layer)}

\begin{itemize}
\item \textbf{Prediction}: progress in large language models will not replace ontology-type artificial intelligence that retains a prior cognitive framework: every time LLM capability jumps, institutional demand for a constitutive semantic layer rises rather than falls.
\item \textbf{Theoretical basis}: LLMs have types without instances, knowledge without a stance (Chapter 8, O1); semantic anchorage is a complement to LLMs, not a competitor --- the prior framework is responsible for giving what the LLM reads in a constitutive, auditable, instance-bound landing ground, and the runtime division of labor is given in Chapter 6. This prediction's leading indicator is designated not by market size but computed from the criterion's gradient: the practice of building ontologies for human organizations sits in the criterion's middle band (Section 4.3), where the prior's standing holds only in grades; if replacement is to happen, it must happen first at that most exposed position. If that position holds, the side of strongly constituted artificial physical worlds goes without saying.
\item \textbf{Falsification condition}: generational progress in LLMs systematically bypasses the ontology layer: enterprises achieve responsibility-endorsed deployment at scale in institutional scenarios such as operations and maintenance, directly with general-purpose LLM agents and without any constitutive semantic layer.
\item \textbf{What is damaged if wrong}: the complementarity structure of semantic anchorage and LLMs (Chapter 6), and the predictive power of the criterion's middle-band analysis (Section 4.3). This prediction and P3 are two faces of one thing: P3 bets that the posterior route cannot enter the institutional door alone; P4 bets that to enter the door it must bring us along.
\end{itemize}
\textbf{Structural corollary I: the cross-industry invariance of the concept layer (a corollary, not a bet).} The concept layer's claim is this: every legitimate purpose in the artificial physical world can be decomposed into artifactual constitutive dimensions --- the transfer of energy, matter, information, or mechanical work, together with spatial relations; and every failure is an instance or a combination of a closed typology of failure modes. Its pivot is not the finality of the current enumeration but the fact that design comes first (Section 5.3 proved: completeness comes from the order in which promulgation reaches instances, not from the closure of any list). The current enumeration (Appendix B.1, supplementary material) is the best present candidate: empirical, revisable, and spontaneous refinement within its vocabulary space belongs to the normal augmentation passage (the port of entry of Section 5.2). The decision line is thereby written out: if onboarding a new industry requires only adding entries, that is normal evolution; if it requires modifying or abolishing existing entries of the concept layer, this corollary fails, and the damaged components are the layering lower bound (\allowbreak{}Structural Characterization 5.1) and the promulgation-capture argument of Section 5.3. In addition, if someone proposes a fifth construction goal that passes the assignability test and is compatible with none of the four goals (Section 5.2), or exhibits a three-layer framework satisfying all four goals, this corollary fails together with Structural Characterization 5.1. A constructive demonstration is on record: the two industries of Appendix B (building cooling, deep-space exploration; supplementary material) share the same syntax layer and the common part of the concept layer, with zero diff (a demonstration, not independent evidence: the vocabulary reuse was constructed under a "verbatim carry-over" discipline; see the cross-domain convention of Appendix B (\allowbreak{}supplementary material)).

\textbf{Structural corollary II: the concept-layer boundary of the organizational domain (a corollary, not a bet).} Prior frameworks for human organizations exhibit a triple structure: (a) at the mission layer (the divergent zone) no cross-client shared concept layer will appear, and per-client manual distillation (such as forward-deployed engineering) is a persistent cost of that layer; (b) at the convergent layer (finance, compliance, administration) a shared concept layer has long existed and will keep being reused (the ERP model); (c) the framework's value density rises with the degree of regimentation of the organization's mission: defense and intelligence highest, large industrial organizations next, ordinary commercial organizations lowest. Basis: organizations are the middle band where both conjuncts of the criterion are satisfied only in grades (Section 4.3); and competition, as the universal "physics" of commercial organizations, has anti-convergence as its first law --- it produces shared adjectives (moats, niches) and forbids shared nouns (missions).

\textbf{Closing: the Lakatosian wager.} The four predictions share one structure: the failure of each corresponds to damage to a specific component of the framework (for the propagation chain, see the stake structure at the head of this chapter). We do not ask the reader to accept this programme because its arguments are elegant; we ask the reader to write down these four predictions and then watch which way the world goes. In Lakatos's distinction (Lakatos, 1978): a research programme that busies itself patching around its hard core when predictions fail is degenerating; one whose predictions keep cashing out and giving rise to new problems is progressive. This paper chooses to put its programme on the table: \emph{if it is wrong, the framework is wrong.}

\subsection{Chapter summary}
This chapter has given four falsifiable predictions (the order of breakthroughs, the win-loss map, the VLA ceiling, and the irreplaceability of the ontology layer) each marked with its theoretical basis, falsification condition, and the component at stake; and it has registered two structural corollaries (the cross-industry invariance of the concept layer, and the organizational-domain boundary) whose validity derives from the criterion and from "design comes first," with preliminary evidence from the zero diff across the two industries of Appendix B (\allowbreak{}supplementary material). The way to test the theory is thereby completely public: four bets await challengers; two corollaries come with their derivations attached.

\section{Chapter 8. Objections and Replies}
The honesty of a theory paper shows in how it treats its own counterexamples. This chapter first records one occasion on which this framework overthrew itself during its formation, then replies to the five strongest objections one by one. Each objection enters in its strongest version --- weakened straw men are not entertained in this chapter.

\textbf{Self-criticism: why not take "temporal priority of design" as the criterion.} The theory in fact set out with the temporal criterion (the full argument is in Section 4.2.1). It was vetoed by the gene counterexample: genes, too, precede proteins, and temporal order cannot distinguish drawings from genes. It was thereby upgraded to "intentional constitution $\wedge$ readable archive." We record this self-overthrow as it happened, because it is precisely a demonstration of how a criterion works: a good criterion is not proclaimed; it is pruned into shape by counterexamples.

\textbf{O1: Large language models have already read everything humanity has written --- manuals, specifications, drawing texts are all in the corpus. Is the prior framework you discuss not simply what LLMs already have inside?} This is the most dangerous objection this paper faces, and the strongest form of Chapter 2's convergence temptation; it merits a four-part dissection.

\emph{First, the mode of digestion flattens normativity.} An LLM reads a specification with the same mechanism it reads a novel: the statistics of the next symbol (Bender \& Koller, 2020). The "shall" in the constitutive document "the supply water temperature shall be 7 $^\circ$C" is flattened by statistical digestion into "people usually write this"; the model can recite the norm, yet no mechanism can make it \emph{bound by the norm}. And normativity (purpose, function, fault) is precisely the least replaceable ingredient in institutional scenarios. \emph{Second, emergent representations are not promulgated concepts.} Mechanistic interpretability research shows that decodable feature structures exist inside LLMs: probes (Alain \& Bengio, 2017) and sparse autoencoders (Bricken et al., 2023) can read out entities, attributes, even rudimentary world models. We do not deny this; but we first point out its nature: such research reads regularities \emph{after the fact} out of an already trained machine, isomorphic to grammarians reading grammar out of speech and physicists reading laws out of phenomena. \emph{Interpretability research is the physics of LLMs, not their engineering.} Concede one step further: interventional experiments (\allowbreak{}activation patching (Meng et al., 2022), steering vectors (Turner et al., 2023)) prove these features are indeed causal parts, more than "after-the-fact storytelling." But even granting the features are all real, they remain \emph{emergent} rather than promulgated: drifting with the next training run, with no one endorsing their invariance; and carrying no norm --- a steering vector can make a model talk more about Paris, but no feature can prescribe what it \emph{shall} talk about or what deviation counts as a fault. The concepts of a prior framework are the opposite: each is \emph{a commitment that has been laid down}: prior to data, constraining interpretation, lapsing upon drift. What the other side finds are the machine's laws; what we erect are the world's drawings. \emph{However true a law, it cannot exercise the office of a drawing.} \emph{Third, types without instances.} An LLM's knowledge is an average over types: it knows what chiller manuals generally say; it does not know whether \emph{this} chiller at this moment deviates from \emph{its own} specification. Symbol-to-symbol proximity is not symbol-to-world anchoring (Harnad, 1990). \emph{Fourth, not auditable.} Institutionalized operations require semantically determinate status reports (the meaning of every grade fixed in the framework) not plausible prose. The difference between a gray box with a skeleton and a black box without one is, where endorsement is needed, the difference between usable and unusable. In sum: an LLM \emph{has knowledge without a stance} toward the worlds it has read. It has a definite place inside our framework (the universal tool for reading archives) but it is a reader, not a layer: its place is outside the framework, beside the four layers, not in the concept layer.

\textbf{One final piece of industry self-evidence.} The industry's own evolution has already testified to this: every agent system moving toward practicality erects human-promulgated scaffolding outside the model, and by the criterion of Chapter 4, scaffolding is prior (the full account is in Section 2.4, R.5). Only one fact is registered here: if pure posterior learning sufficed, this scaffolding would not be erected again and again --- priors are not the theorist's preference; they are a necessary condition rediscovered by practice whenever systems land.

\textbf{O2: Three peripheral objections, answered together --- evolved objects, appropriated artifacts, and "nothing but ontology engineering."} First, \emph{AlphaFold succeeded on evolved objects: can priors do without "intentional constitution"?} AlphaFold's lesson was registered in Section 4.3.2: the comparison of the two generations proves that the weakened archive condition C2$'$ (readable generative trace) can be satisfied by accident, not that (i) is unnecessary; and the price paid is the price of (i)'s absence: structure accessible, norms not. Second, \emph{artifacts get appropriated: a paperclip picks a lock, a screwdriver pries a lid, an old factory becomes a café; purposes drift.} Purpose drift is real, but it is no threat to the criterion, because drift's mode of existence in the institutionalized world is not "silent deviation" but \emph{new design events}: converting an old factory into a café comes with renovation drawings, change-of-use approvals, and a fresh fire-safety review: intention is re-promulgated, the archive reissued, and the normativity changes wholesale with the new intention. Non-institutional appropriation (lock-picking with a paperclip) exists, but it lies outside the claim domain of "high utility value, institutionally maintained" --- the framework's claim scope was drawn in Section 4.5, and the paperclip is not in it. Third, \emph{when all is said, is what you do not just ontology engineering?} Ontology engineering provides the craft; this paper provides the \emph{license}. Ontology engineering since Gruber answers "how to construct an ontology properly," but not "in which world constructing an ontology is legitimate, and why": the engineering practice presupposed legitimacy, and this paper gives its criterion (Section 4.2), its cost map (Section 4.3), its sufficient conditions (Section 4.4), and the \emph{shape} a qualified ontology must satisfy (the layering lower bound of Structural Characterization 5.1, Section 5.4). If this is called ontology engineering, then it is ontology engineering delivered together with its own premises.

\textbf{O3: Drawings govern construction, but not use and aging. Real operational knowledge is tacit and lives in the old masters (Polanyi, 1966) --- buildings "learn" by themselves; deviation from design is the norm.} The territory this objection stakes out, we partly accept and partly dissolve. First, acceptance: the criterion claims only the range where "the design is on record"; the framework honestly does not cover purely tacit knowledge (the guardrail of Section 4.3). Then, the cut: the true hard core of tacit knowledge does not lie in "the semantics of failure": what failures are and by which categories of source they are excited has already been systematized by engineering into enumerable classification schemes (Ishikawa, 1968): the source categories of failure evolution for individual equipment and spaces can be enumerated by the excitation structure "man, machine, material, method, environment, plus system inputs." The hard core lies in "the concrete knack": how this weld is to be welded, how this pump's abnormal sound is to be heard. It never claims to govern concrete knacks; what it governs is "which category of failure has occurred, and which procedure shall answer it." The tacitness of a handling knack and the decidability of the normative judgment "this shall be done" are two different propositions: the objector is right, but her being right does not touch this paper's claim. Two supplements: most "unmeasurable phenomena" in the field are superpositions of known failure modes rather than new concepts (\allowbreak{}combinations do not produce new types; Section 5.5); and institutionalized operations is itself the historical process of codifying tacit knowledge, with the framework as its carrier, not its enemy.

\textbf{O4: In complex, tightly coupled systems, unanticipated interactions among multiple faults can trigger system-level accidents (Perrow, 1984) --- do such accidents exceed the reach of any component-level prior framework?} Before replying, we request a premise from the objector: in the face of a system-level accident, what is the task of intelligence? Prior prediction, in-event detection and handling, and post-hoc attribution and explanation are three different tasks, and their accounts must be kept separate.

\emph{Prior prediction.} For a system-level accident genuinely formed by the symptomless convergence of multiple independent factors, prior prediction is indeed impossible; but this is not a particular incapacity of this framework --- it is the structural incapacity of any intelligence: such events belong with earthquakes and wars to force majeure and have already been moved out of the claim domain by the boundary corollary of Section 4.5.

\emph{In-event detection and handling.} The great majority of so-called system-level "emergence" is in fact not emergence but the accumulation and cross-system spreading of small errors. Chernobyl was no undecomposable holistic catastrophe, rather a chain of individually observable, identifiable, and explainable small errors: the reactor's positive void coefficient design flaw, the mismatch between operating procedures and physical characteristics, the safety test conducted in violation (IAEA, 1992) --- every link falling within the semantics of the failure types known at the time. And the \emph{transfer} of failures between systems can itself be modeled: standard semantic interfaces are defined between systems, transferring graded state semantics rather than physical parameters (Appendix B.1, supplementary material); every link on the propagation chain is then a monitorable interception point. An intelligence that can recognize each small error on the spot as a "symptom" intercepts precisely the final accident.

\emph{Post-hoc attribution and explanation.} This is the framework's home ground: ascending the collapse chain, every judgment gives its reason and every failure reports its location (Section 5.5); the attribution report is a by-product of the structure, not an extra engineering project. Its engineering form is fully displayed in the two cases of Appendix B (the cooling plant's real-time cause-tracing and the Mars rover's after-the-fact replay; supplementary material). One symmetry is added: the posterior route has no advantage here, since a black-box model facing an unseen common-cause failure is equally extrapolating, and its failure is silent (Section 4.4), whereas the framework's failure is an alarm. Finally, the disagreement is put on the table: Chapter 7 of this paper has already wagered that the concept layer is invariant across industries --- if system-level phenomena truly require every industry to modify, rather than augment, existing concept-layer entries, the facts will adjudicate this objection.

\textbf{O5: Physical laws work as priors in the basic physical world --- PINNs carved Navier--Stokes into the loss function and succeeded (Raissi et al., 2019). So descriptive rules are equally legitimate priors, and conjunct (i), "intentional constitution," is simply unnecessary.} The edge of this objection comes from a category mistake: it takes "an explicitly expressed posterior" for "a prior framework." Physical laws are not norms promulgated to the world before its birth; they are humanity's statistical summary of centuries of observation of the basic physical world; their epistemic identity is the same as that of grammatical rules: descriptive rules, posterior to phenomena (Section 4.1). What PINNs do is to front-load maximally compressed posterior knowledge as training constraints: clever engineering, but it supplies no constitutive norms: there is no "fault" semantics, only "prediction error"; no promulgative force is inherited, and when observations disagree, it is the law that gets revised, not the world (Section 4.4). Their legitimacy is therefore retrospective and has nothing to do with this criterion; their scope of application is strictly confined to the handful of problems that admit approximation as closed physical systems (Section 4.3.2). \emph{A physical law is a posterior written out explicitly, not a prior; it describes how the world runs, and does not define how the world shall be maintained.} The difference reduces to three entries: a physics engine constrains the state space (what is physically impossible); a prior framework constrains the ought-to-be subset of the state space (what counts as normal, what counts as a fault). A deviation of the former triggers a prediction error; a deviation of the latter triggers a violation judgment. The former's correction direction is back to the laboratory to revise the model; the latter's is to the field to correct the instance. For the item-by-item comparison, see Section 2.4, R.6.

\subsection{Chapter summary}
This chapter recorded one self-overthrow of the framework (the temporal criterion pruned by the gene counterexample into the promulgation criterion) and replied to the five strongest objections: the LLM encyclopedia objection (O1); evolved objects, purpose drift, and the ontology-engineering challenge (O2); tacit knowledge (O3); system-level accidents (O4); and physical laws as priors (O5). All replies share one maneuver: disassemble the objection back into the criterion's two conjuncts, and point out which conjunct it pressed and which it missed. The objections did not shake the criterion; the criterion, in return, drew for each objection the range within which it is valid.

\section{Chapter 9. Conclusion: Interrogable Intelligence}
Return to the deadlock we started from: no data, no intelligence; no intelligence, no data. All the work of this paper can be gathered into a single answer to that deadlock --- the deadlock is universal, but it is not uniform: \emph{in the artifacts it values most, human civilization has left a seam in the deadlock. That seam is called design.}

The four-world partition shows that cognitive tasks are stratified for the learner (Chapter 3). It shows that among the worlds exactly one permits "framework first, instances later": the artificial physical world, intentionally constituted and leaving readable archives (Chapter 4). The layering lower bound shows that such a framework needs at least a four-layer skeleton --- the syntax layer promulgates the lexicon, the concept layer holds steady, the knowledge layer houses the procedures, the instance layer lets posterior data land, and the collapse semantics lets multitudinous readings reduce upward into clauses (Chapter 5). The flywheel thereby replaces the deadlock: the prior framework gives intelligence that runs from day one; operation produces data; the data flow back at the instance layer; intelligence grows in use. The relation with large language models is division of labor rather than competition: every promulgatable position goes to the engine, every non-promulgatable one to the LLM, and sandwiched generation is thereby auditable generation (Chapter 6). The theory's way of being tested is completely public: four predictions and two structural corollaries lie on the table; if the criterion is wrong, everything at stake is lost (Chapter 7).

This does not mean the other three worlds can be bypassed. The phenomenal world, the basic physical world, and the artificial symbolic world are each an unexcisable part of complete intelligence; this paper proposes no path to general intelligence; this is a deliberate boundary, not an oversight. But the worlds play different roles: only in the designed world can intelligence's understanding of the world be reviewed layer by layer, interrogated clause by clause, endorsed level by level. The intelligence this paper advocates does not pretend to be transparent down to every parameter: its flesh is posterior machine learning; it is a gray box. But the gray box has a skeleton: the skeleton is the prior's promulgation, prescribing where the flesh attaches, within what range it moves, and how far a drift counts as dislocation. Models on the phenomenal and basic-physical directions forever carry the risk of silent failure out of distribution; models on the symbolic direction have knowledge without a stance toward the worlds they have read; only the cognitive framework of the artificial physical world (because what it inherits is promulgation rather than extrapolation) can trace every judgment up the collapse chain into a reason, and locate every failure to a concrete layer (\allowbreak{}Proposition 5.2, Corollary 5.1). We call such intelligence \emph{interrogable}: every judgment it makes can be questioned down the collapse chain to the promulgated norm clause, and where it cannot answer is an alarming failure. It is not interpretable (an explanation can be an after-the-fact fabrication), nor merely auditable (an audit only reads the records; an interrogation demands an answer on the spot). In an age when more and more decisions are handed to machines, \emph{interrogable intelligence is not a luxury; it is infrastructure} --- and interrogability comes not from transparency but from the skeleton.

Which world comes with its own source code? Only the designed one. This layer of civilization has laid its drawings open on the table; learning to read them is the most faithful and cheapest entry point for machines into the physical world.

\emph{A set of framework demonstrations appears in Appendix B (\allowbreak{}supplementary material): from an everyday cooling plant (Appendix B.1, real-time cause-tracing) to an unrepairable Mars rover (Appendix B.2, an after-the-fact replay under cold-start conditions) --- the extraction of constitutive priors from engineering archives, their instantiation as running intelligence, and how one alarm collapses along the vertical edges to its cause in the concept layer (for the feedforward form's preventive decision, see Section 6.4).}

\begin{quote}
\textbf{Note.} Appendices A (the formal skeleton), B (framework demonstrations) and C (external-witness dossiers) appear in the accompanying supplementary material.
\end{quote}
\textbf{Acknowledgements.} The author thanks colleagues at Persagy Science and Technology for discussions of the engineering practice on which Chapter 5 and Appendix B (\allowbreak{}supplementary material) are based.

\textbf{Declaration of generative AI and AI-assisted technologies in the writing process.} During the preparation of this work, the author used Kimi (Moonshot AI) for research assistance, structural editing, bilingual drafting, and schematic drafting of figures. The author reviewed and edited the content as needed and takes full responsibility for the content of the publication.

\section{References}
\begin{reflist}
\item Alain, G., Bengio, Y., 2017. Understanding intermediate layers using linear classifier probes. ICLR 2017 Workshop Track.
\item Andrighetto, G., Governatori, G., Noriega, P., van der Torre, L. (Eds.), 2013. Normative Multi-Agent Systems. Dagstuhl Follow-Ups, Vol. 4. Schloss Dagstuhl--Leibniz-Zentrum für Informatik.
\item Anscombe, G.E.M., 1957. Intention. Basil Blackwell, Oxford.
\item ASHRAE. Proposed Standard 223P: Semantic Data Model for Analytics and Automation Applications in Buildings (in preparation).
\item Assran, M., Duval, Q., Misra, I., et al., 2023. Self-supervised learning from images with a joint-embedding predictive architecture (I-JEPA). In: Proceedings of CVPR 2023; \url{arXiv:2301.08243}.
\item Baader, F., Calvanese, D., McGuinness, D., Nardi, D., Patel-Schneider, P.F. (Eds.), 2003. The Description Logic Handbook: Theory, Implementation and Applications. Cambridge University Press.
\item Badreddine, S., d'Avila Garcez, A., Serafini, L., Spranger, M., 2022. Logic tensor networks. Artificial Intelligence 303, 103649.
\item Balaji, B., Bhattacharya, A., Fierro, G., et al., 2016. Brick: Towards a unified metadata schema for buildings. In: Proceedings of BuildSys'16, pp. 41--50.
\item Bender, E.M., Koller, A., 2020. Climbing towards NLU: On meaning, form, and understanding in the age of data. In: Proceedings of ACL 2020, pp. 5185--5198.
\item Bernardo, J.M., Smith, A.F.M., 1994. Bayesian Theory. Wiley, Chichester.
\item Black, K., Brown, N., Driess, D., et al., 2025. $\pi$$_0$: A vision-language-action flow model for general robot control. In: Robotics: Science and Systems XXI (RSS); \url{arXiv:2410.24164}.
\item Boella, G., van der Torre, L., 2004. Regulative and constitutive norms in normative multiagent systems. In: Proceedings of KR'04, pp. 255--266.
\item Bricken, T., Templeton, A., Batson, J., et al., 2023. Towards monosemanticity: Decomposing language models with dictionary learning. Transformer Circuits Thread.
\item Brohan, A., Brown, N., Carbajal, J., et al., 2023. RT-2: Vision-language-action models transfer web knowledge to robotic control. In: Proceedings of CoRL 2023, PMLR 229, 2165--2183; \url{arXiv:2307.15818}.
\item Bruce, J., Dennis, M.D., Edwards, A., et al., 2024. Genie: Generative interactive environments. In: Proceedings of ICML 2024, PMLR 235, 4603--4623; \url{arXiv:2402.15391}.
\item Bybee, J., 2010. Language, Usage and Cognition. Cambridge University Press.
\item Carpenter, G.A., Grossberg, S., 1987. A massively parallel architecture for a self-organizing neural pattern recognition machine. Computer Vision, Graphics, and Image Processing 37(1), 54--115.
\item Chen, M., Tworek, J., Jun, H., et al., 2021. Evaluating large language models trained on code. \url{arXiv:2107.03374}.
\item Chomsky, N., 1957. Syntactic Structures. Mouton, The Hague.
\item Church, K., 2011. A pendulum swung too far. Linguistic Issues in Language Technology 6(5).
\item Console, L., Torasso, P., 1991. A spectrum of logical definitions of model-based diagnosis. Computational Intelligence 7(3), 133--141.
\item Darwiche, A., 1998. Model-based diagnosis using structured system descriptions. Journal of Artificial Intelligence Research 8, 165--222.
\item Dastani, M., 2008. 2APL: A practical agent programming language. Autonomous Agents and Multi-Agent Systems 16(3), 214--248.
\item d'Avila Garcez, A., Lamb, L.C., 2023. Neurosymbolic AI: The 3rd wave. Artificial Intelligence Review 56, 12387--12406.
\item Deacon, T.W., 1997. The Symbolic Species: The Co-evolution of Language and the Brain. W.W. Norton, New York.
\item de Kleer, J., Williams, B.C., 1987. Diagnosing multiple faults. Artificial Intelligence 32(1), 97--130.
\item de Kleer, J., Mackworth, A.K., Reiter, R., 1992. Characterizing diagnoses and systems. Artificial Intelligence 56(2--3), 197--222.
\item Dennett, D.C., 1987. The Intentional Stance. MIT Press, Cambridge, MA.
\item Esteva, M., Rodríguez-Aguilar, J.A., Sierra, C., García, P., Arcos, J.L., 2001. On the formal specification of electronic institutions. In: Dignum, F., Sierra, C. (Eds.), Agent Mediated Electronic Commerce: The European AgentLink Perspective, Lecture Notes in Artificial Intelligence, Vol. 1991. Springer, pp. 126--147.
\item Falkenhainer, B., Forbus, K.D., 1991. Compositional modeling: Finding the right model for the job. Artificial Intelligence 51(1--3), 95--143.
\item Feigenbaum, E.A., 1977. The art of artificial intelligence: Themes and case studies of knowledge engineering. In: Proceedings of IJCAI-77, pp. 1014--1029.
\item Gold, E.M., 1967. Language identification in the limit. Information and Control 10(5), 447--474.
\item Gruber, T.R., 1993. A translation approach to portable ontology specifications. Knowledge Acquisition 5(2), 199--220.
\item Guarino, N., 1998. Formal ontology and information systems. In: Formal Ontology in Information Systems. IOS Press, pp. 3--15.
\item Ha, D., Schmidhuber, J., 2018. Recurrent world models facilitate policy evolution. In: NeurIPS 2018, pp. 2450--2462; \url{arXiv:1803.10122}.
\item Harnad, S., 1990. The symbol grounding problem. Physica D 42(1--3), 335--346.
\item Hartmann, N., 1940. Der Aufbau der realen Welt: Grundriss der allgemeinen Kategorienlehre. De Gruyter, Berlin.
\item IAEA, 1992. The Chernobyl Accident: Updating of INSAG-1. INSAG-7, IAEA Safety Series No. 75-INSAG-7, Vienna.
\item Ishikawa, K., 1968. Guide to Quality Control. Asian Productivity Organization, Tokyo.
\item Jumper, J., Evans, R., Pritzel, A., et al., 2021. Highly accurate protein structure prediction with AlphaFold. Nature 596, 583--589.
\item Kelemen, D., 2004. Are children "intuitive theists"? Reasoning about purpose and design in nature. Psychological Science 15(5), 295--301.
\item Kroes, P., Meijers, A., 2006. The dual nature of technical artefacts. Studies in History and Philosophy of Science 37(1), 1--4.
\item Lakatos, I., 1978. The Methodology of Scientific Research Programmes. Cambridge University Press.
\item LeCun, Y., 2022. A path towards autonomous machine intelligence. OpenReview Preprint.
\item Lenat, D.B., 1995. CYC: A large-scale investment in knowledge infrastructure. Communications of the ACM 38(11), 33--38.
\item Manning, C.D., Schütze, H., 1999. Foundations of Statistical Natural Language Processing. MIT Press.
\item Mao, J., Gan, C., Kohli, P., Tenenbaum, J.B., Wu, J., 2019. The neuro-symbolic concept learner. In: ICLR 2019.
\item Marr, D., 1982. Vision: A Computational Investigation into the Human Representation and Processing of Visual Information. W.H. Freeman, San Francisco.
\item Meng, K., Bau, D., Andonian, A., Belinkov, Y., 2022. Locating and editing factual associations in GPT. In: NeurIPS 2022; \url{arXiv:2202.05262}.
\item Modbus Organization. Modbus Application Protocol Specification (\allowbreak{}originated at Modicon, 1979; current specification maintained by the Modbus Organization).
\item Moravec, H., 1988. Mind Children: The Future of Robot and Human Intelligence. Harvard University Press.
\item Muscettola, N., Nayak, P.P., Pell, B., Williams, B.C., 1998. Remote Agent: To boldly go where no AI system has gone before. Artificial Intelligence 103(1--2), 5--47.
\item Newell, A., 1982. The knowledge level. Artificial Intelligence 18(1), 87--127.
\item NVIDIA, 2025. Cosmos world foundation model platform for physical AI. \url{arXiv:2501.03575}.
\item NVIDIA. Isaac Sim documentation (product documentation).
\item OMG, 2016. Meta Object Facility (MOF) Core Specification, Version 2.5.1. Object Management Group.
\item OpenAI, 2023. GPT-4 technical report. \url{arXiv:2303.08774}.
\item Packer, C., Fang, V., Patil, S.G., Lin, K., Wooders, S., Gonzalez, I., 2023. MemGPT: Towards LLMs as operating systems. \url{arXiv:2310.08560}.
\item Palantir Technologies. Foundry Documentation: Ontology Overview. \url{https://www.palantir.com/docs/foundry/ontology/overview.}
\item Perrow, C., 1984. Normal Accidents: Living with High-Risk Technologies. Basic Books.
\item Piaget, J., 1954. The Construction of Reality in the Child. Basic Books, New York.
\item Polanyi, M., 1966. The Tacit Dimension. Doubleday, Garden City, NY.
\item Popper, K.R., 1972. Objective Knowledge: An Evolutionary Approach. Oxford University Press.
\item Project Haystack. Semantic modelling for device and equipment data. \url{https://project-haystack.org.}
\item Raissi, M., Perdikaris, P., Karniadakis, G.E., 2019. Physics-informed neural networks. Journal of Computational Physics 378, 686--707.
\item Rasmussen, J., 1985. The role of hierarchical knowledge representation in decisionmaking and system management. IEEE Transactions on Systems, Man, and Cybernetics SMC-15(2), 234--243.
\item Reiter, R., 1987. A theory of diagnosis from first principles. Artificial Intelligence 32(1), 57--95.
\item Rombach, R., Blattmann, A., Lorenz, D., Esser, P., Ommer, B., 2022. High-resolution image synthesis with latent diffusion models. In: CVPR 2022, pp. 10684--10695; \url{arXiv:2112.10752}.
\item Schick, T., Dwivedi-Yu, J., Dessì, R., et al., 2023. Toolformer: Language models can teach themselves to use tools. In: NeurIPS 2023; \url{arXiv:2302.04761}.
\item Searle, J.R., 1983. Intentionality: An Essay in the Philosophy of Mind. Cambridge University Press.
\item Searle, J.R., 1995. The Construction of Social Reality. Free Press, New York.
\item Senior, A.W., Evans, R., Jumper, J., et al., 2020. Improved protein structure prediction using potentials from deep learning. Nature 577, 706--710.
\item Seymour, L.M., Maragh, J., Sabatini, P., Di Tommaso, M., Weaver, J.C., Masic, A., 2023. Hot mixing: Mechanistic insights into the durability of ancient Roman concrete. Science Advances 9(1), eadd1602.
\item Silver, D., Huang, A., Maddison, C.J., et al., 2016. Mastering the game of Go with deep neural networks and tree search. Nature 529, 484--489.
\item Silver, D., Sutton, R.S., 2025. Welcome to the era of experience. Technical report, Google DeepMind. Preprint of a chapter to appear in: Designing an Intelligence, MIT Press.
\item Simon, H.A., 1969. The Sciences of the Artificial. MIT Press.
\item Spelke, E.S., Kinzler, K.D., 2007. Core knowledge. Developmental Science 10(1), 89--96.
\item Struss, P., 1992. What's in SD? Towards a theory of modeling for diagnosis. In: Hamscher, W., Console, L., de Kleer, J. (Eds.), Readings in Model-Based Diagnosis. Morgan Kaufmann, pp. 419--449.
\item Su, H., 2026. Physical intelligence: From illusion to reality (in Chinese). Keynote speech, Main Forum of the 2026 World Artificial Intelligence Conference (WAIC 2026), 2026-07-17, Shanghai. Full text reprinted: Fudan University News, 2026-07-17, \url{https://news.fudan.edu.cn/2026/0717/c236a149901/page.htm.}
\item Sutton, R.S., 2026. Keynote speech, Main Forum of WAIC 2026 (from the "era of human data" to the "era of experience"), 2026-07, Shanghai. Content identical to Silver \& Sutton (2025).
\item Thomasson, A.L., 2003. Realism and human kinds. Philosophy and Phenomenological Research 67(3), 580--609.
\item Todorov, E., Erez, T., Tassa, Y., 2012. MuJoCo: A physics engine for model-based control. In: Proceedings of IROS 2012, pp. 5026--5033.
\item Tomasello, M., 2003. Constructing a Language: A Usage-Based Theory of Language Acquisition. Harvard University Press.
\item Turner, A.M., Thiergart, L., Leech, G., Udell, D., et al., 2023. Activation addition: Steering language models without optimization. \url{arXiv:2308.10248}.
\item Uexküll, J. von, Kriszat, G., 1934. Streifzüge durch die Umwelten von Tieren und Menschen. Springer, Berlin.
\item von Wright, G.H., 1951. Deontic logic. Mind 60(237), 1--15.
\item W3C, 2004. RDF Primer / RDF 1.1 Concepts and Abstract Syntax. W3C Recommendation.
\item W3C OWL Working Group, 2012. OWL 2 Web Ontology Language Document Overview, 2nd ed. W3C Recommendation.
\item Wang, F.-Y., 2004. Parallel system methods for management and control of complex systems. Control and Decision 19(5), 485--489.
\item Wu, Q., Bansal, G., Zhang, J., et al., 2023. AutoGen: Enabling next-gen LLM applications via multi-agent conversation. \url{arXiv:2308.08155}.
\item Yao, S., Zhao, J., Yu, D., et al., 2023. ReAct: Synergizing reasoning and acting in language models. In: ICLR 2023.
\item Yuan, Y., Yao, A.C.-C., 2026. Calculus of intelligence: A topos-monadic framework for agentic workflows. iFuture, Online First, 9710001. doi:10.26599/\allowbreak{}IF.2026.9710001.
\end{reflist}

\end{document}


\maketitle

\noindent\textbf{Contents}: Appendix A (the formal skeleton: criterion semantics, learning boundaries, decidability, readability gradient, certificate-chain boundary, sample-complexity separation, layering lower bound), Appendix B (framework demonstrations: B.1 the cooling-plant case; B.2 the Curiosity rover drill-feed anomaly, Sol 1536), and Appendix C (four-layer reverse-reading dossiers of eight external witnesses: BACnet, LonWorks, OPC, Modbus, ODD, COIN, Brick/Haystack, RDF/OWL). This volume carries its own bibliography; entries cited by both volumes appear in both.

\section{Appendix A. The Formal Skeleton}
\subsection{Notation table}
This table collects all symbols used in the main text and in this supplementary volume; each row gives the symbol, its meaning, and the place of first definition. Overloading conventions (the dual duties of $N$, $k$, $I$, and $q$, and the multiple roles of $C$) are declared at their points of use in the main text and in \S{}A.7.

\subsubsection{A. Object domain and the criterion (\S{}3--\S{}4 of the main text)}
\begin{center}\small
\begin{tabular}{@{}>{\raggedright\arraybackslash}p{0.313\textwidth}>{\raggedright\arraybackslash}p{0.313\textwidth}>{\raggedright\arraybackslash}p{0.313\textwidth}@{}}
\toprule
Symbol & Meaning & Introduced \\
\midrule
$D = \langle I, N, A\rangle$ & object domain: instances, norms, archive & Definition 4.1 \\
$I$ & instance set (physical objects and processes of the domain) & Definition 4.1 \\
$N$ & norm set (statements of how instances shall be) & Definition 4.1 \\
$A$ & archive (machine-readable record in symbolic form) & Definition 4.1 \\
$N \lhd I$ & ``$N$ constitutes $I$'': existence and identity of instances depend on compliance with $N$, fixed prior to them & Definition 4.1 / A.7 \\
C1, C2 & the two conjuncts of the criterion: intentional constitution; readable generative archive & Proposition 4.1 \\
C2$'$ & weakened archive condition: readable generative \emph{trace} (no promulgation source required) & \S{}4.3.2 / Definition A.8 note \\
$dev(i, n)$ & the event that instance $i$ deviates from norm $n$ & Definition 4.3 \\
$\mathrm{Violation}(i)$ & resolution of $dev(i,n)$ under promulgation: a defect of the world & Definition 4.3 \\
$\mathrm{ModelDefect}(n)$ & resolution under description: a defect of the model & Definition 4.3 \\
$\mathrm{CoverageGap}$ & coverage gap (no slot / no enumeration match): judged ``unknown'', referred to augmentation & Definition 4.4 \\
$\mathrm{MaintenanceFault}$ & promulgated content in mutual contradiction: an A1-fidelity alarm & Definition 4.4 \\
S1--S3 & zero-shot operability / promulgated constraint / loud failure & Proposition 4.2 \\
I1--I3 & norm accessibility / violation audibility / accountability closure & Corollary 4.2 \\
A1--A4 & working assumptions: archive fidelity / semantic accessibility / coverage / promulgation direction on duty & \S{}4.4 \\
P1--P4 & the four falsifiable predictions & Chapter 7 \\
\bottomrule
\end{tabular}
\end{center}

\subsubsection{B. Discipline functions and layering (\S{}5 of the main text)}
\begin{center}\small
\begin{tabular}{@{}>{\raggedright\arraybackslash}p{0.313\textwidth}>{\raggedright\arraybackslash}p{0.313\textwidth}>{\raggedright\arraybackslash}p{0.313\textwidth}@{}}
\toprule
Symbol & Meaning & Introduced \\
\midrule
$\mathcal{C}$ & the total content set of a framework & \S{}5.4 / Definition A.27 \\
$\rho: \mathcal{C} \to \{F, I\}$ & revision discipline: $F$ = closure between promulgation events; $I$ = standing intake channel & \S{}5.4 / A.27 \\
$\delta: \mathcal{C} \to \{T, N\}$ & denotation discipline: $T$ = denoting types; $N$ = anchored to instances & \S{}5.4 / A.27 \\
$x \sim y$ & carrier equivalence: $x \sim y \iff \rho(x)=\rho(y) \land \delta(x)=\delta(y)$ & Definition 5.1 / A.27 \\
$k$ & number of layers = number of non-empty equivalence classes & Definition 5.1 \\
G1--G4 & construction goals: stability / openness / abstraction / concreteness & Definition 5.2 \\
$C_t$ & the concept layer at its current promulgated version (finite closed set) & Proposition 5.2 \\
$d$ & collapse-chain depth ($d \leq 4$, promulgated as an upper bound) & Proposition 5.2 / H4 \\
$\lvert C\rvert$ & number of candidates per level in reduction; reduction cost $O(d\cdot\lvert C\rvert)$ & Proposition 5.2 \\
$\succ$ & strict ordering on the two gradient axes (e.g., artificial physical world $\succ$ regimented organizations $\succ$ \dots{}) & \S{}4.3.4 \\
\bottomrule
\end{tabular}
\end{center}

\subsubsection{C. Input/output logic (\S{}A.1)}
\begin{center}\small
\begin{tabular}{@{}>{\raggedright\arraybackslash}p{0.313\textwidth}>{\raggedright\arraybackslash}p{0.313\textwidth}>{\raggedright\arraybackslash}p{0.313\textwidth}@{}}
\toprule
Symbol & Meaning & Introduced \\
\midrule
$\mathcal{L}$ & propositional language & Definition A.1 \\
$N \subseteq \mathcal{L}\times\mathcal{L}$ & I/O norm set; $(a,x)$ reads ``in situation $a$, output $x$ is required'' & Definition A.1 \\
$\mathrm{out}_3$, $\mathrm{out}_3^C$ & simple-minded reusable output; its constrained variant --- the former used in violation tests, the latter in inferential contexts & Definition A.2 \\
$B$; $N(B)$; $\mathrm{Cn}$ & auxiliary input-closure set; norms triggered by $B$; logical-consequence closure & Definition A.2 \\
$s$ & instance state (input to the output operator); $s \not\models \mathrm{out}_3(N,s)$ marks inconsistency & Definition A.3 \\
$T$; $\mathrm{Th}(\mathrm{Obs})$; $T \circ o$ & theory; theory of observations; AGM revision of $T$ by $o$ & Definition A.4 \\
$\alpha$; $t_0$; $t_i$ & promulgating institution; promulgation time; constitution time of instance $i$ ($t_0 < t_i$: priority condition) & Definition A.7 \\
$\mathrm{acc}$ & acceptance function (terminating pass/fail procedure) & Definition A.7 \\
$I_0$; $I^N_{\mathrm{acc}}$ & world-given candidate domain; the constituted set $\{i \in I_0 \mid \mathrm{acc}_N(i)\ \mathrm{passes}\}$ & Definition A.7 commentary \\
\bottomrule
\end{tabular}
\end{center}

\subsubsection{D. Learning boundaries (\S{}A.2)}
\begin{center}\small
\begin{tabular}{@{}>{\raggedright\arraybackslash}p{0.313\textwidth}>{\raggedright\arraybackslash}p{0.313\textwidth}>{\raggedright\arraybackslash}p{0.313\textwidth}@{}}
\toprule
Symbol & Meaning & Introduced \\
\midrule
$\Sigma$; $\Sigma^{*}$; $\Sigma^{*<\omega}$ & instance-description alphabet; finite strings; finite prefixes & Definition A.10 \\
$O_w$ & occurrence language of world $w$ & Definition A.10 \\
$L^{*}_w$ & legal instance set of world $w$ (norm-compliant instances) & Definition A.10 \\
$\mathrm{Obs}(w)$ & observation mechanism: a positive presentation (text) of $O_w$ & Definition A.10 \\
$\mathcal{H}_{\mathrm{adm}}$ & admissible hypothesis space (constrained family of boundary candidates) & Definition A.12 \\
$\varphi$ & learner: observation prefix $\to$ candidate boundary & Definition A.12 \\
H-O (H-O1, H-O2) & finite combinatorial openness (superfinite language class) & \S{}A.2 \\
H-N & normative boundary independence & \S{}A.2 \\
\bottomrule
\end{tabular}
\end{center}

\subsubsection{E. Decidability of reduction (\S{}A.3)}
\begin{center}\small
\begin{tabular}{@{}>{\raggedright\arraybackslash}p{0.313\textwidth}>{\raggedright\arraybackslash}p{0.313\textwidth}>{\raggedright\arraybackslash}p{0.313\textwidth}@{}}
\toprule
Symbol & Meaning & Introduced \\
\midrule
H1--H5 & reduction preconditions: closed concept layer / index-enumerable candidates / set-level covering test / depth bound / chain continuity & Definition A.15 \\
$\mathrm{covers}(n, \mathrm{current})$ & covering test (disjunctive, existential semantics; $O(1)$ via bitmask) & Definition A.15 (H3) \\
$L$; $\ell$ & promulgated level set for the object class; a level in it & Algorithm 1 \\
$\mathrm{current}$, $\mathrm{next}$ & node sets during level-by-level reduction & Algorithm 1 \\
KNOWN-TYPE; UNRESOLVED & known-type localization; reduction failure (referred to the port of entry) & Algorithm 1 \\
\bottomrule
\end{tabular}
\end{center}

\subsubsection{F. Certificate chains (\S{}A.5)}
\begin{center}\small
\begin{tabular}{@{}>{\raggedright\arraybackslash}p{0.313\textwidth}>{\raggedright\arraybackslash}p{0.313\textwidth}>{\raggedright\arraybackslash}p{0.313\textwidth}@{}}
\toprule
Symbol & Meaning & Introduced \\
\midrule
$\mathrm{Typed}(T)$ & typed context of task $T$ & Definition A.20 \\
$C$ & a certificate-anchored compositional calculus & Proposition A.21 \\
\bottomrule
\end{tabular}
\end{center}

\subsubsection{G. Sample complexity (\S{}A.6)}
\begin{center}\small
\begin{tabular}{@{}>{\raggedright\arraybackslash}p{0.313\textwidth}>{\raggedright\arraybackslash}p{0.313\textwidth}>{\raggedright\arraybackslash}p{0.313\textwidth}@{}}
\toprule
Symbol & Meaning & Introduced \\
\midrule
H-N$'$ & parameterized boundary family (H-N at $k=1$) & \S{}A.6 \\
$k$ & number of boundary elements (note: distinct from the layer count $k$ of \S{}5.4/\S{}A.7; see the notation note in \S{}A.7) & H-N$'$ \\
$e_1,\dots,e_k$ & boundary elements; $S \subseteq \{e_1,\dots,e_k\}$; worlds $w_S$ (all $2^k$ share $O$) & H-N$'$ \\
$L_{\mathrm{arc}}$; $L_{\mathrm{post}}$ & archive-side learner (0 instance samples); posterior-side learner (positive presentation only) & Definition A.24 \\
$\mathrm{err}(\varphi, w_S)$ & Hamming error rate over the $k$ coordinates & Definition A.24 \\
$d_H$ & Hamming distance between boundary vectors & Theorem A.26(a) \\
$\hat{S}$; $\hat{S}_m$ & learner's estimate of $S$ (after $m$ observations) & Theorem A.26(a) \\
$q$ & label arrival rate (note: not the label-noise bias $1/2 \pm q$ of the model-boundary note) & Lemma A.25 \\
$m$; $\varepsilon$ & number of observations; error tolerance & \S{}A.6 \\
Criterion Alpha / Beta & Hamming error rate $\leq \varepsilon$; exact recovery of $S$ with probability $\geq 2/3$ & Theorem A.26(b) \\
$\Theta(\cdot)$, $\Omega(\cdot)$, $O(\cdot)$ & asymptotic bounds; $\mathbb{E}[\cdot]$ expectation; $\mathbf{1}\{\cdot\}$ indicator; $\lceil\cdot\rceil$ ceiling & passim \\
$F$, $I$ / $T$, $N$ & discipline values: freeze / standing intake (the $\rho$ axis); denoting types / anchored to instances (the $\delta$ axis) & \S{}5.4 / A.27 \\
\bottomrule
\end{tabular}
\end{center}

\subsubsection{H. Case-study codes (Appendix B; code-style proper names left untranslated)}
\begin{center}\small
\begin{tabular}{@{}>{\raggedright\arraybackslash}p{0.313\textwidth}>{\raggedright\arraybackslash}p{0.313\textwidth}>{\raggedright\arraybackslash}p{0.313\textwidth}@{}}
\toprule
Symbol & Meaning & Introduced \\
\midrule
\texttt{0 / 0.25 / 0.5 / 1} & graded state semantics: normal / adverse trend / symptom / failed & B.1.2 \\
\texttt{degreeLevel} & runtime field carrying the grade on knowledge-layer nodes & B.1.3 \\
\texttt{hazardEntityScope} & the four promulgated depth levels of the causal tree (caps $d$) & B.1.3.2 \\
\texttt{priorMTTF} & prior mean time to failure of an object class & B.1.3.2 \\
\texttt{DIP / SEE / LTT / JRI / JC / SOT / ARNP} & the monitoring-and-judgment chain entry types & B.1.3.4 \\
FED / FEST & the two workaround operating modes (feed-extended drilling / feed-extended sampling) & B.2 \\
\bottomrule
\end{tabular}
\end{center}

\begin{quote}
Positioning statement: this appendix provides rigorous formal twins for the core constructions of the main text: the criterion semantics of Chapter 4 (A.1), the learning boundary of archiveless worlds (A.2), the decidability of the layered framework (A.3), the sample-complexity separation (A.6), the layering lower bound (A.7), and rigorous statements of two supporting results --- the readability gradient (A.4) and the certificate-chain boundary (A.5). The readable version in the main text is self-contained; this appendix introduces no new claims and only supplies formal counterparts to existing ones. The three-tier declaration: this appendix strictly distinguishes three tiers. Items labeled \textbf{Proposition} are mathematical facts, with proofs or proof sketches attached, sketches honestly marked as such; items labeled \textbf{Theorem} are mathematical facts of the same standing, whose format marks the heaviest technical result of this appendix; items labeled \textbf{Structural Characterization} are direct classification results of definitions plus normative premises: their rigor lies in the argument, their contribution in the classification framework itself, and they claim no new mathematical fact; items labeled \textbf{Modeling Corollary} are direct corollaries of modeling assumptions, mostly appearing as sub-items subordinate to the number of their host proposition (e.g., A.13(ii)), and independently numbered ones (e.g., Lemma A.25) have the same status; wherever "Proposition A.13" is cited with (ii) specifically meant, this registered status governs; items labeled \textbf{H (\allowbreak{}hypothesis)} are engineering modeling assumptions whose truth is supplied by concrete systems, not argued by this paper; wherever the words "legitimate" or "qualified" appear, the item is a normative commitment of this paper, argued in the main text, not masquerading as a theorem; a \textbf{Criterion} is the formal expression of a normative commitment this paper adopts (e.g., A.5, A.9), and its justification is in the main text. Numbering convention: definitions, lemmas, propositions, theorems, structural characterizations, criteria, corollaries, and remarks share a single running sequence (A.1, A.2, \dots{}). Unlike the per-category counting style of the main chapters, the unified sequence avoids the cross-category reference ambiguity that separate numbering would cause; hypotheses are numbered in the H series. Hypothesis numbering is split into four independent groups: the main-text working assumptions A1--A4 (Section 4.4), H-O/H-N of A.2, H1--H5 of A.3, and H-N$'$ of A.6; the four groups have different scopes and are not consecutively numbered.
\end{quote}
\subsection{A.1 The formal semantics of $\lhd$: input/output logic}
\subsubsection{Reasons for the framework choice}
Formalizing the constitution relation $\lhd$ first faces a choice of framework. Standard deontic logic (SDL) represents "ought" by the modal operator O$\varphi$ and carries three burdens that conflict with this paper. First, it comes with possible-worlds semantics, whereas Definition 4.1 of the main text has declared that $\lhd$ "says nothing about possible worlds." Second, it treats norms as truth-valued propositions and thereby falls into Jørgensen's dilemma (Jørgensen, 1937/38): imperatives are neither true nor false, so how can they take part in logical inference? Third, it yields theorems at odds with engineering intuition, such as Ross's paradox (Ross, 1941), and it cannot coherently represent contrary-to-duty situations (Chisholm's paradox (Chisholm, 1963)). Input/output logic (Makinson \& van der Torre, 2000, 2001) was designed to circumvent these three burdens: a norm is not a proposition but an ordered pair (a, x), read "in situation a, x is the required output." This corresponds item by item with the actual form of engineering specifications: "operating condition a $\to$ supply water temperature shall be 7 $^\circ$C" is not a description of the world but a requirement on outputs. Two alternatives are each excluded in one sentence: normative multi-agent system frameworks take agents' compliance/violation behavior as the semantic primitive, presupposing what this paper sets out to argue (the binding force of norms over instances) and are therefore circular; deontic action logic embeds norms into action semantics, is expressively excessive, and still rests on possible worlds.

\subsubsection{The basic framework}
\textbf{Definition A.1 (I/O norm set).} Let $\mathcal{L}$ be a propositional language. The norm set N of an object domain D = $\langle$I, N, A$\rangle$ (\allowbreak{}Definition 4.1) is formalized as a set of I/O rules: $N \subseteq \mathcal{L} \times \mathcal{L}$, with (a, x) $\in$ N read "in situation a, output x is required."

\textbf{Definition A.2 (output operator).} The base operator is $\mathrm{out}_3$ (simple-minded reusable output) (Makinson \& van der Torre, 2000), whose closure construction is as follows: let the auxiliary set B be the smallest set such that B contains all logical consequences of s; if (a, x) $\in$ N and a is derivable from B, then x $\in$ B (\allowbreak{}detachment; output items fed back as inputs give reusability); and B is closed under logical consequence. The output is defined as $\mathrm{out}_3(N, s) = \mathrm{Cn}(N(B))$, where N(B) = {x $\mid$ (a, x) $\in$ N, a $\in$ B}; the output is closed under conjunction and logical consequence (AND, weakening of the consequent), but \emph{the output does not inherit the identity of the input state}: the consequences of s itself enter the auxiliary set B only as triggering conditions and do not automatically enter the output $\mathrm{out}_3(N, s)$ --- the Cn closure inherited by the output operator acts on N(B), not on s itself, so that what \emph{is} does not mix into what \emph{ought} to be. What this paper actually uses is its constrained variant $\mathrm{out}_3^C$ (Makinson \& van der Torre, 2001): the output is constrained for consistency with the input: if s is consistent, then $\mathrm{out}_3^C(N, s) \cup s$ is consistent; constrained output is understood via the standard 2001 construction (through maximal consistent subfamilies), not as an extra axiom appended to $\mathrm{out}_3$. Reasons for the choice: engineering norms require reusability (the same norm takes effect repeatedly for recurring operating conditions, which the non-reusable $\mathrm{out}_1$ does not guarantee); and violations must not entail trivialization: the consistency constraint registers "norm library in disrepair" as a detectable event rather than letting an inconsistent input blow up the output. \textbf{The division of labor between the two operators is hereby fixed}: violation tests (\allowbreak{}Definition A.3) use $\mathrm{out}_3$ --- a directly violated norm must remain in the output, since otherwise the most typical violation would be filtered out by the consistency screening and escape detection; the consistency-constrained $\mathrm{out}_3^C$ serves the inference layer's explosion protection, and what it contains is exactly the case of a norm library in disrepair (the MaintenanceFault of Definition 4.4). Wherever "out" is written below, $\mathrm{out}_3$ is meant in testing contexts and $\mathrm{out}_3^C$ in inferential ones.

\subsubsection{The direction of fit, theoremized}
This is the most central step of this appendix: upgrading the "adjudged a defect of the world / of the model" language of Definitions 4.2 and 4.3 from case-law phrasing to a precise criterion.

\textbf{Definition A.3 (\allowbreak{}promulgation direction).} A norm system (N, out) is in the promulgation direction if and only if inconsistency events are resolved as follows: given an instance state s, if $s \not\models \mathrm{out}_3(N, s)$, then N is held fixed, the event resolves to Violation(s), and the world side is marked as to-be-corrected.

\textbf{Definition A.4 (\allowbreak{}description direction).} A model system is in the description direction if and only if inconsistency events are resolved as follows: the model is a theory T = Th(Obs); if an observation o is inconsistent with T, then o is held fixed, the theory is updated by an AGM revision operator to T$'$ = T $\circ$ o (\allowbreak{}Alchourrón, Gärdenfors \& Makinson, 1985), and the model side is revised.

\textbf{Criterion A.5 (the direction-of-fit criterion).} This paper restricts fit resolution to two normative policies with one side held fixed: an inconsistency event e resolves to Violation if and only if the norm side is held fixed; it resolves to ModelDefect if and only if the evidence side is held fixed and the theory side is revised.

\textbf{Commentary.} This is a direct unfolding of Definitions A.3 and A.4; its only non-trivial part is stating the boundary of the restricted policies. The resolution operator is determined by which side is fixed and which side moves; of the four combinations, the other two do not enter the jurisdiction of this criterion: if both sides move, all anchors are lost; if both sides stay still, the inconsistency is suspended forever. Cases where the two sides move jointly (standards and models revised in coordination, negotiation in multi-party governance) fall outside the jurisdiction of this criterion and must be adjudicated separately at the governance layer. Concept immigration does not break the promulgation direction: the concept augmentation of Section 5.2 is an expansion of the concept carrier ($C_{t+1} \supseteq C_t$, corresponding to the (I, T) cell of Definition A.27), not a revision of the I/O rule set N over existing operating conditions; N's adjudicative authority over existing instances is unchanged, so augmentation does not enter the "both sides move" exclusion above.

\textbf{Corollary A.6 (the formal twin of Definitions 4.2/4.3).} The promulgation/description distinction of Definition 4.2 is Definitions A.3/A.4; the Violation/ModelDefect resolution of Definition 4.3 is the two policies delimited by Criterion A.5. The two main-text definitions are thereby upgraded from operational case law to an operable formal distinction; their scope of application is likewise bounded to one-side-fixed policies.

\subsubsection{The constitutive semantics of $\lhd$ and a restatement of the criterion}
\textbf{Definition A.7 (\allowbreak{}constitution).} $N \lhd I$ if and only if there exist a promulgation event and an acceptance function: an institution $\alpha$ issues N at time t$_0$; and there exists an acceptance function acc --- a terminating decision procedure taking an instance i and the norms $\mathrm{out}(N, \cdot)$ as input and outputting a binary status (pass/fail), with a fail output triggering a correction procedure; for every i $\in$ I, acc(i) completes and passes at a time $t_i > t_0$. The priority condition is exactly $t_0 < t_i$.

\textbf{Commentary (against circularity).} Let the candidate domain I$_0$ be the totality of to-be-constituted instances given by the world, and write $I^N_{\mathrm{acc}}$ = {i $\in$ I$_0$ $\mid$ $\mathrm{acc}_N$(i) passes}. The institutional meaning of "constitution" is: $I = I^N_{\mathrm{acc}}$, and membership in I is sustained by the institutional fact of "acceptance invoking N" (\allowbreak{}revocation of the acceptance voids the membership). Note that acc adjudicates compliance at the constitution time $t_i$; it is a one-shot event, not a continuing condition on instance identity: when an instance deviates after passing acceptance, its membership does not change, and the deviation state resolves to Violation (\allowbreak{}Definition A.3) --- constitution and compliance are thereby separated. (The phrase "sustained by institutional fact" belongs to the interpretive layer of constitution and is not part of the formal content.) There is no circularity here: I is given in advance by the world, not defined by acc; what acc adjudicates is the pre-given I. Commissioning, acceptance, and handover are precisely the actual forms of the word "constitution" in the engineering world.

\textbf{Definition A.8 (readable generative archive, formal version).} An archive A is a readable generative archive of D if and only if: A records the promulgated content of N (in part or in whole; the archive records the rules themselves, and the logical consequences of the rules are generated on demand by the out operator without being archived) in a machine-parseable symbolic form, and A exists prior to the corresponding instances. This is the restatement, in the I/O framework, of the three conditions at the end of Definition 4.2. Its weakened form C2$'$ (a readable generative trace, not requiring a promulgation source) is registered in Section 4.3.2.

\textbf{Criterion A.9 (the Constitution Criterion, the formal twin of Proposition 4.1).} (Necessity direction, argued in Section 4.2.3 of the main text.) Legitimately extracting a prior cognitive framework over an object domain D entails: (C1) $N \lhd I$ (\allowbreak{}Definition A.7); (C2) there exists a readable generative archive A of D (\allowbreak{}Definition A.8). (\allowbreak{}Sufficiency direction.) This paper adopts C1 $\wedge$ C2 as the criterion of legitimacy: this is a normative commitment of this paper, not a theorem.

\subsection{A.2 The learning boundary of archiveless worlds: non-identifiability of normative boundaries}
The naive treatment (jumping from Gold's theorem to the conclusion in one step) lacks a modeling stage. The rigorous version has two tiers: one is a direct application of Gold's theorem; the other is an independent non-identifiability argument. We first give the formal setup.

\textbf{Definition A.10 (\allowbreak{}occurrence stream and observation mapping).} Let $\Sigma$ be an instance-description alphabet and $\Sigma^{*}$ the set of all finite instance strings. A world w determines an occurrence language $O_w \subseteq \Sigma^{*}$ --- the totality of instances actually occurring in w. The observation mechanism Obs(w) is a positive presentation (text) of $O_w$: an infinite sequence enumerating $O_w$ in which each element appears at least once. Identifiability is quantified over all presentations in the standard way: the learner must succeed in the limit on every positive presentation of $O_w$ to count as identifying it. The key fact: Obs depends only on the occurrence stream, and not on whether any normative-boundary candidate is true. $L^*_w \subseteq \Sigma^{*}$ is the legal instance set of w: the totality of norm-compliant instances. Note that $L^*_w$ and $O_w$ in general do not coincide: a compliant configuration that never occurs can be in $L^*_w$ but not in $O_w$; an occurring but violating event can be in $O_w$ but not in $L^*_w$; the two may also coincide or even be equal (one set of witnesses for H-N takes L\textasteriskcentered{} = O).

\textbf{Lemma A.11 (no authoritative labels).} In a world without a promulgated source of truth, the presentation stream carries no authoritative compliance labels: any mechanism for attaching compliance marks presupposes that norms can be consulted, contradicting the assumption. This lemma does not deny statistical separability over unlabeled data --- under structural assumptions, latent classes may be identifiable from the distribution; what it denies is an authoritative source of labels.

\textbf{Definition A.12 (candidate family and learner).} Admissible normative-boundary candidates form a constrained family $\mathcal{H}_{\mathrm{adm}} \subseteq 2^{\Sigma^{*}}$ --- a family of subsets satisfying syntactic and structural constraints, not the full power set. A learner is a function $\varphi: \Sigma^{*<\omega} \to \mathcal{H}_{\mathrm{adm}}$ that takes an observation prefix as input and outputs a candidate boundary.

\textbf{Hypothesis H-O (finite combinatorial openness).} (H-O1) For every finite $F \subseteq \Sigma^{*}$, there exists an admissible world w whose occurrence language is exactly F. (H-O2) There exists at least one admissible world $w_\infty$ such that $O_{w\infty}$ is infinite. Together the two clauses make the language class of occurrence languages a superfinite class --- one containing all finite languages and at least one infinite language. H-O is a modeling assumption: it registers the empirical assertion that "all finite combinations of instances of an open physical world can occur, with no upper bound on scale"; whether it holds is not argued in this paper.

\textbf{Hypothesis H-N (normative boundary independence).} There exist admissible worlds w$_1$, w$_2$ with $O_{w_1} = O_{w_2}$ while $L^*_{w_1} \neq L^*_{w_2}$ --- the difference between the two worlds is given precisely by the difference in boundary. A concrete set of witnesses: take $e \in O_{w_2}$ (it suffices that $O_{w_2}$ be non-empty), let $L^*_{w_1} = O_{w_1}$ and $L^*_{w_2} = O_{w_2} \setminus \{e\}$: e being some occurring yet violating event; and require $L^*_{w_1}, L^*_{w_2} \in \mathcal{H}_{\mathrm{adm}}$, since otherwise non-identifiability would hold trivially because the target lies outside the hypothesis space, and the witnesses would lose their weight.

\textbf{Proposition A.13 (open worlds resist rule coverage, rigorous version).} Under a presentation stream without authoritative labels: (i) (under H-O) even the language class to which O belongs is not identifiable in the limit from positive presentations (\allowbreak{}identifiability is a property of language classes, not of individual languages); (ii) (\textbf{Modeling Corollary}, under H-N) the normative boundary L\textasteriskcentered{} is not identifiable by the observation mechanism Obs: there exist w$_1$ $\neq$ w$_2$ with Obs(w$_1$) = Obs(w$_2$) while the legal boundaries differ.

\textbf{Proof.} Note that the two tiers differ in logical type: (i) is the standard Gold situation of learning O itself (\allowbreak{}presentation and target identical); in (ii), presentation and target are not identical, Gold's theorem itself does not apply, and an independent non-identifiability argument is required. (i) Here the learner's hypothesis space is the language class to which occurrence languages belong (\allowbreak{}distinguished from the boundary learner of Definition A.12). By H-O1 $\wedge$ H-O2, the language class is superfinite; this is a direct application of Gold's (1967) theorem --- which is quantified over all positive presentations, consistent with the convention of Definition A.10: superfinite language classes are not identifiable in the limit from positive presentations. (ii) By H-N, there exist w$_1$ $\neq$ w$_2$ with $O_{w_1} = O_{w_2}$ and different legal boundaries (the witnesses are the ones given by H-N). By Lemma A.11, the presentation stream contains no authoritative compliance labels, so the learner's entire available input is Obs(w); by Definition A.10, Obs depends only on the occurrence stream, hence Obs(w$_1$) = Obs(w$_2$): the two worlds supply the same presentation stream. Any $\varphi$ outputs the same on the same input stream, and its convergent output is correct for at most one world. Hence no $\varphi$ converges to the correct boundary on all admissible worlds. Note the dimensional difference from Gold: Gold says "positive presentation cannot identify certain language classes"; this proposition says "the observation mechanism is not injective in the normative boundary." The former is a limit of text learning; the latter is a structural gap between observation and norms. Tier (ii) does not masquerade as a learning theorem: it is a direct corollary of the modeling assumption H-N through the observation mapping Obs, and its content is the formalization of the irreducibility claim of Section 3.6. Read together: the easier task (\allowbreak{}identifying O) is already impossible; the harder one (carving out L\textasteriskcentered{} within observation) remains unidentifiable even if O is granted for free. \hfill$\square$

\textbf{Remark.} This paper's contribution here is not a new theorem but the modeling: non-identifiability is established by two modeling facts ("no authoritative labels" (Lemma A.11) and "Obs does not depend on boundary candidates" (\allowbreak{}Definition A.10)) rather than being conjured from the intuition that "open worlds are hard." The other side of the boundary should also be registered: the classes learnable from positive data are characterized by Angluin (1980); this paper's negative result does not deny positive results on particular restricted classes --- what it denies is the attainability of the target "normative boundaries of open worlds."

\textbf{Corollary A.14 (the diagnosis of Cyc).} Consider a handwritten rule set R for an archiveless world: R's sufficiency can only be assessed by inductive evaluation against the observation stream. By H-N, there exist two worlds with the same stream and different boundaries; any evaluation procedure issues the same verdict on the same input and is correct for at most one world --- sharing one root with A.13(ii) (the Modeling Corollary): the presentation stream does not carry the authoritative labels that evaluating sufficiency would require, so the same input stream can support opposite verdicts. Hence such evaluation cannot be completed in the limit either. Cyc's stagnation is therefore not an engineering shortfall but an instance of a learnability boundary: what failed was not the writing of the rules but their confirmation.

\subsection{A.3 Decidability of the layered framework: the algorithm and bound of Proposition 5.2}
\textbf{Definition A.15 (reduction preconditions).}

\begin{itemize}
\item (H1) The concept layer $C_t$ is a finite closed set at every time t;
\item (H2) the reducible candidates at each level are enumerable by index within $O(|C_t|)$;
\item (H3) the covering test has \emph{disjunctive (\allowbreak{}existential) semantics}: covers(n, current) $\iff$ $\exists$c $\in$ current, covers(n, c); the existential test over the whole set is done in one bitmask intersection (the mask of current is maintained incrementally at line 6 of Algorithm 1), which is $O(1)$ --- word-parallel under the word-RAM model; strictly $O(|C_t|/w)$, where w is the machine word length, treated as constant within vocabulary scale; if done by per-key inverted-index lookup it would be $O(|\mathrm{current}|)$, and this bound is stated on the bitmask convention. H3 is an interface assumption: the concrete implementation of covers (a precomputed coverage matrix or a vocabulary index) is supplied by engineering, and Proposition A.16 states only the complexity when the test can be done in $O(1)$; for implementations requiring inferential determination, this bound is multiplied by the cost of that determination;
\item (H4) the collapse-chain depth is bounded above by a promulgated constant d = 4: d is the depth of the reduction levels inside the concept layer, and the enumeration at each level is promulgated by the concrete framework (in Appendix B.1, hazardEntityScope promulgates: medium unit under equipment $\to$ component under equipment $\to$ entity combination under equipment $\to$ the equipment under analysis; another framework may promulgate: instance $\to$ failure mode $\to$ factor family $\to$ root category). This is not the number of architectural layers, which is given independently by Structural Characterization 5.1. The value d = 4 is stated as an upper bound: the actual depth of each concrete object is capped by its parts tree, and shallower parts trees mean shallower chains (the drill feed mechanism of Appendix B.2 has d = 3); the algorithm traverses the level set L actually promulgated for the object class, and empty levels do not enter the enumeration (traversal by promulgated levels).
\item (H5) chain continuity: for every reducible alarm, its reduction chain has a covering candidate at every promulgated level of L (the causal chain does not skip promulgated levels; empty levels have already been removed in L's construction). H5 is a correctness assumption, not a complexity one: if some fault chain skipped a promulgated level, the algorithm would spuriously return UNRESOLVED; hence the reading of UNRESOLVED as "the current concept layer does not cover" is conditioned on H5;
\end{itemize}
\textbf{Algorithm 1 (DECIDE-REDUCTION).} Input: an instance-layer alarm a, the concept layer $C_t$, and the level set $L \subseteq \{1..d\}$ actually promulgated for the object class (\allowbreak{}promulgated by the concept-layer archive). Output: a known-type localization KNOWN-TYPE(current), or reduction failure UNRESOLVED (the current $C_t$ does not cover it; refer to the port of entry of Section 5.2 to assess whether it constitutes a concept augmentation).

\begingroup\footnotesize
\begin{verbatim}
1:  current <- {a}
2:  for level l in L do                         // traverse promulgated
    levels in ascending order; skipped levels are not renumbered (B.2.2.3)
3:      next <- {}
4:      for each n in Enumerate(C_t, l) do        // H2: at most |C_t|
    candidates
5:          if covers(n, current) then            // H3: set-level indexed
    test, O(1)
6:              next <- next U {n}
7:      if next = {} then return UNRESOLVED       // reduction failure:
    current C_t does not cover
8:      current <- next
9:  return KNOWN-TYPE(current)                    // reduction succeeded:
    known-type localization
\end{verbatim}
\endgroup
If |current| > 1 (several reduction paths coexist), KNOWN-TYPE outputs the set itself, with the semantics of \emph{disjunctive localization}: the instance belongs to one of these types, and the adjudicative ambiguity is reported as-is rather than forcibly disambiguated inside the algorithm. Disambiguation is the responsibility of the normative adjudication at the knowledge layer, consistent with the "localization versus adjudication" distinction.

\textbf{Proposition A.16 (reduction complexity under constant-time covering index; the rigorous version of Proposition 5.2).} Under H1--H5 (H3 supplying the $O(1)$ covering-test premise, H4 the depth upper bound, and H5 not participating in the count but supplying the reading of UNRESOLVED), DECIDE-REDUCTION completes known-type localization in $O(|L|\cdot|C_t|)$; since $|L| \leq d$, this bound remains bounded by $O(d\cdot|C_t|)$, and the statement of Proposition 5.2 is unchanged. Whether the localization is a violation is separately given by the normative status of the located type (the output of knowledge-layer adjudication) and lies outside this algorithm.

\textbf{Proof.} By induction on levels. Level $\ell$: candidate enumeration has at most $|C_t|$ items (H2); each candidate undergoes one covering test against current as a whole, $O(1)$ (the set-level convention of H3). Hence each level costs $O(|C_t|)$. The number of promulgated levels |L| is bounded above by the constant d = 4 (H4), so the total cost is $O(|L|\cdot|C_t|) \leq O(d\cdot|C_t|)$. \hfill$\square$

\textbf{Proposition A.17 (control: enumerative reduction).} For an enumerative reduction process (Algorithm 1 and its variants): if the concept layer is not a closed set (the candidate space is countably infinite) and there is no independent exclusionary decider, then reduction success is semi-decidable (halt upon a hit), while reduction failure is undecidable ("not yet hit" cannot be distinguished from "does not exist").

\textbf{Argument.} With infinitely many candidates, line 4 of Algorithm 1 no longer terminates; positive instances are semi-decidable, and negative instances are undecidable within this algorithmic model. The proposition does not deny the possibility of non-enumerative direct deciders over infinite spaces; what it states is the boundary of enumerative processes. \hfill$\square$

\textbf{Remark A.18 (\allowbreak{}reconciliation with the openness goal).} Decidability is per-stage: an augmentation event $C_{t+1} \supseteq C_t$ resets the clock, and H1 holds anew on the new stage. This is exactly the formal meaning of the last sentence of Section 5.5 in the main text --- "what is closed is the types; what is open is the instances and the augmentation passages": Proposition A.16 holds for each stage, while the sequence of stages itself is open.

\subsection{A.4 The readability gradient and extraction complexity: three modeling cases}
\textbf{Proposition A.19 (\allowbreak{}extraction cost follows the way the archive was written).} (i) Archives \emph{written-to-be-read}: for machine-readable archives of a fixed grammar G (BIM, structured point lists, machine-readable standards), extraction degenerates to grammar-directed parsing, at a cost from $O(|A|)$ (\allowbreak{}deterministic context-free grammars, the class to which BIM, structured point lists, and machine-readable standards belong) to $O(|A|^3)$ --- general context-free grammars. Semantic parsing of pure natural-language archives is not counted in this tier and should be placed on the gradient according to its degree of formalization. (ii) Archives that \emph{happen to be readable} (records, logs, accident reports from which norms must be reconstructed): extraction involves causal-structure discovery; if norm reconstruction is modeled as Bayesian-network structure learning over contributing factors, its decision form is NP-complete (\allowbreak{}Chickering, 1996) and its optimization form NP-hard. This paper accordingly registers the complexity magnitude of this tier; it does not claim that all causal-structure discovery is hard. (iii) \emph{No archive and no other authoritative label source}: extraction degenerates to the situation of Proposition A.13: the normative boundary is not identifiable. (Note: authoritative per-event labeling itself constitutes a kind of "degenerate promulgation," i.e., a degenerate archive containing only a label stream, and belongs to tier (i), not to the present one.)

\textbf{Proof sketch.} (i) Grammar-directed parsing under a fixed grammar has polynomial algorithms, with output size polynomially bounded by input length. (ii) This is Chickering's negative result for a specific structure-learning problem, registered according to the declared modeling correspondence. (iii) This is a restatement of Proposition A.13. \hfill$\square$

\textbf{Remark.} The readability gradient of Section 4.3 in the main text thereby acquires a complexity reading: the historical order in which AI penetrated the physical world (\allowbreak{}engineered facilities first, while norm extraction for genomes and ecosystems remains unbroken to this day, successes at the structure level not counting (see Section 4.3.2)) is precisely the empirical shadow of this gradient.

\subsection{A.5 The certificate-chain boundary}
\textbf{Definition A.20 (archive anchoring).} For a task T situated in a world W, its typed context Typed(T) is \emph{legitimately grounded} if and only if: (0) the archive A was intentionally promulgated --- the product of a constitution act in the sense of Definition A.7; (i) every type object and interface declaration of Typed(T) is extracted from this promulgated, readable archive A of W; (ii) every such declaration is traceable to clauses of A.

\textbf{Proposition A.21 (the applicability domain of certificate-anchoring calculi).} Let C be any certificate-anchored compositional calculus whose soundness theorem has as antecedents a typed task context and certificates anchored to declared assumptions (Theorem 5.1 of (Yuan \& Yao, 2026) being the typical example). If W satisfies $\neg$intentionally-constituted(W) $\vee$ $\neg$readable-archive(W), then no task in W has a legitimately grounded typed context. Pushing back along the antecedents: C's soundness theorem holds at most vacuously in W; and in any instantiation of C in W, every terminal declaration in the certificate chain that is not a mathematical foundation is a stipulation the calculus itself cannot check --- a declaration with no archive endorsing it.

\textbf{Proof.} Suppose some task T in W has a legitimately grounded Typed(T). If W lacks a readable archive, this directly contradicts Definition A.20(i); if the archive was not intentionally promulgated (a natural record that happens to be readable) then Definition A.20(0) fails, and again there is no legitimate grounding. Hence the antecedent of the soundness theorem cannot be legitimately satisfied in W; as a conditional, the theorem remains true, but vacuously so. Finally, certificates in C by construction require anchored declarations; where there is no archive to anchor to, every terminal link that is not a mathematical foundation (axioms and mathematical facts excepted) is a stipulation whose truth the calculus itself cannot check. \hfill$\square$

\textbf{Remark A.22 (the boundary does not touch the mathematics).} Proposition A.21 does not touch the mathematics of such calculi: a conditional is vacuously true where its antecedent fails, at no loss. What the legitimacy criterion adds is a map of the antecedent: it says which worlds can supply legitimately grounded typed contexts (those intentionally constituted and documented) and which cannot. The four-layer framework of Chapter 5 occupies the same boundary from the other side: its knowledge layer is precisely the artifact that engineering archives can legitimately ground. The two constructions thus converge at the criterion --- one descending from symbolic composition, the other ascending from physical archives.

\textbf{Remark A.23 (the triad is one argument).} Read in order, the three results answer three different kinds of reader: the skeptic, the systems builder, and the strategist. Proposition A.13 tells the skeptic why archiveless worlds do not yield to handwritten frameworks --- not a cardinality accident but a non-identifiability boundary. Proposition A.16 tells the systems builder what a closed concept layer buys: the decidability of localization, a property end-to-end learning cannot currently offer (its counterpart, Proposition A.17, registers the failure conditions of that property: enumerative reduction over a non-closed set, with failure undecidable). Proposition A.19 tells the strategist why the criterion's first successes appear where things are written-to-be-read, and predicts that penetration into happens-to-be-readable and archiveless domains will follow the complexity gradient, not the enthusiasm of fields. Theorem A.26 (Appendix A.6) then turns this argument onto the sample-complexity axis: the value of the archive relative to instance data is registered quantitatively as a three-tier separation.

\subsection{A.6 The sample-complexity separation: how many samples is the archive worth}
\textbf{Positioning: the division of labor with Proposition A.13.} Proposition A.13 answers the identification-in-the-limit question: even with the observation stream extended indefinitely, the normative boundary of an archiveless world is unattainable. This section answers the finite-sample question: at any finite sample size, what can the two kinds of learner respectively achieve? In one sentence: A.13 governs the boundary (even an infinite stream does not teach it); this section governs the ledger (how many samples each tier of authoritative information is worth). This section does not duplicate A.13: tier (a) is the finite-sample strengthening of A.13(ii) and honestly claims that relation; tier (b) (partial authoritative information) and the archive-side upper bound have no counterparts in A.13.

\textbf{Hypothesis H-N$'$ (\allowbreak{}parameterized boundary family).} For every $k \in \mathbb{N}$, there exist an occurrence language O and k pairwise-distinct "boundary elements" $e_1, \dots, e_k \in O$ such that for every subset $S \subseteq \{e_1, \dots, e_k\}$ there is an admissible world $w_S$ satisfying: $O_{w_S} = O$ (all $2^k$ worlds share the same occurrence language); $L^*_{w_S} = O \setminus S$ (the compliance status of the boundary elements, and only of the boundary elements, flips with the world); and $L^*_{w_S} \in \mathcal{H}_{\mathrm{adm}}$ (every flipped outcome lies within the hypothesis space, since otherwise non-identifiability would hold trivially because the target lies outside the hypothesis space). H-N$'$ is the parameterized strengthening of H-N (H-N being the special case k = 1); like H-N, it is a modeling assumption and is labeled as such under the three-tier declaration.

\textbf{Commentary (strength and retreat lines).} The richness of H-N$'$ is the source of the theorem's lower-bound force: the more complete the hypothesis space, the stronger the non-identifiability. Its fidelity is supplied by the thesis itself --- the coordinate-wise independence of boundary elements' compliance status in the artificial physical world is the formal expression of the fact that normative dimensions cannot be read off physical dimensions. Should it still be deemed too strong, two registered retreat lines exist: (i) relax the full family to a constant-rate codebook subfamily (pairwise Hamming distance $\geq$ ck), leaving the conclusions of tiers (a) and (b) unchanged, with only the constants altered (under this retreat, the tier-(b) lower bound must switch to a Fano argument); (ii) retreat to the weakened version H-N (k = 1), preserving the separation structure of "0 versus no finite bound" while sacrificing the quantitative readings in k and q. This paper advocates the full-family version; neither retreat line is invoked.

\textbf{Definition A.24 (the compliance membership decision task and two kinds of learner).} Given a world $w_S$ (with S unknown to the learner), the learner must output, for $x \in O$, its compliance status $\mathbf{1}\{x \in L^*_{w_S}\}$. Scoring targets the boundary-element vector: x is drawn uniformly from $\{e_1, \dots, e_k\}$, and $\mathrm{err}(\varphi, w_S) = \mathbb{E}_x[\, \mathbf{1}\{ \varphi(x) \neq \mathbf{1}\{x \in L^*_{w_S}\} \}\, ]$ --- the Hamming error rate over the k coordinates. The archive-side learner $L_{\mathrm{arc}}$ takes as input the readable generative archive A (recording the promulgated content of N) and uses no instance samples; the posterior-side learner $L_{\mathrm{post}}$ takes as input only the positive presentation stream of O (m observations; by Lemma A.11, without authoritative compliance labels). (Interface note: here the learner outputs element-wise decision bits, a different interface from that of $\varphi$ in Definition A.12, which outputs $\mathcal{H}_{\mathrm{adm}}$ candidates; under H-N$'$ all $2^k$ patterns are realizable, candidate outputs and bit-wise outputs are mutually derivable, and the two interfaces are equivalent.)

\textbf{Lemma A.25 (\allowbreak{}conditional symmetry).} Under H-N$'$, consider a labeled channel: observations x are drawn independently and uniformly from $\{e_1, \dots, e_k\}$, and each observation independently carries, with probability q, an authoritative compliance label (true upon arrival). Then: (i) unlabeled observations are identically distributed across all $2^k$ worlds (by Definition A.10: Obs depends only on the occurrence stream, and all worlds share O); (ii) given any set of labels seen so far, the unlabeled coordinates are symmetric over the remaining world family: the remaining worlds taking 0 and taking 1 are equal in number and induce the same observation distribution, so any learner's error on each such coordinate, averaged over the remaining world family, is exactly 1/2; (iii) hence under the Hamming-error-rate criterion, any learner must first see labels on a constant fraction of the coordinates before it can press the error below any given constant $\varepsilon < 1/2$.

\textbf{Proof sketch.} (i) Observation content depends only on the occurrence stream (\allowbreak{}Definition A.10), and all worlds share O. (ii) For an unseen coordinate i, the remaining world family is closed under flipping that coordinate, and the two flipped worlds induce the same observation distribution, so any decision rule's error on that coordinate averages to 1/2 over the two worlds. (iii) By (ii), an error rate below a constant requires breaking the symmetry on a constant fraction of the coordinates, and labels are the only information source that breaks the symmetry. \hfill$\square$ (Label: Modeling Corollary --- a direct corollary of H-N$'$ and the channel convention.)

\textbf{Theorem A.26 (the three-tier separation of sample complexity).} Under H-N$'$, the instance-sample requirement of the compliance membership decision task divides into three tiers according to the form in which authoritative information is supplied:

\begin{itemize}
\item \textbf{(a) No labels (q = 0): risk identically 1/2, no finite sample bound.} For every posterior-side learner $L_{\mathrm{post}}$, there exist a presentation and a family of admissible worlds $\{w_S\}$ (neither moving with m) such that: (i) for every $m \in \mathbb{N}$, the expected total Hamming distance under a uniform prior is identically k/2: $\mathbb{E}_S[\, \mathbb{E}[\, d_H(S, \hat{S}_m) \,] \,] = k/2$; in terms of the error rate err of Definition A.24 (\allowbreak{}normalized by k) this is identically 1/2 --- the identity holds for every learner, so the Bayes risk, the infimum over learners in err, is likewise 1/2; (ii) for every m there exists a bad world $w_S$ with $\mathbb{E}[\,\mathrm{err}\,] \geq 1/2$ (the bad world may drift with m); (iii) by the pigeonhole principle (the world family has only $2^k$ members) there exists a fixed world $w_{S*}$ such that $\mathbb{E}[\,\mathrm{err}\,] \geq 1/2$ holds for infinitely many m. The limiting form of this tier is exactly Proposition A.13(ii). (This tier does \emph{not} claim that a fixed world fails at all m: the learner could alternate guesses by the parity of m, distributing the badness across different worlds in turn; the "fails infinitely often" of (iii) is the form needed to connect with A.13.)
\item \textasteriskcentered{}\textasteriskcentered{}(b) Sparse reliable labels (arriving at rate q): $\Theta(k/q)$ (by Criterion Alpha).\textasteriskcentered{}\textasteriskcentered{} The decision criterion comes in two grades: Criterion Alpha (Hamming error rate $\leq$ $\varepsilon$) requires $\Theta(k/q)$ observations; Criterion Beta (exact recovery of S with probability $\geq$ 2/3) requires $\Theta((k \log k)/q)$ observations.
\item \textbf{(c) The complete archive (whole-table lookup): 0 instance samples.} Under C2 $\wedge$ A1 $\wedge$ A2 $\wedge$ A3 (A3's coverage containing all boundary elements of H-N$'$, the same domain of definition as the err of tier (a)) $L_{\mathrm{arc}}$ achieves err = 0 with 0 instance samples.
\end{itemize}
\textbf{Proof.} (a) The $2^k$-point version of Le Cam's two-point argument (Le Cam, 1973). Step one (\allowbreak{}observation content identical): by Definition A.10, Obs depends only on the occurrence stream; by H-N$'$, all $2^k$ worlds share O; identifiability is quantified over all presentations, so the adversary can supply all worlds with the same presentation prefix --- the data are literally identical. Step two (estimator distribution identical): $\hat{S} = \varphi(\mathrm{data})$ is a function of the data. Step three (Hamming summation, averaging over worlds first, then taking expectation over the estimator): fix any realized value $\hat{s}$ of $\hat{S}$; on each coordinate, exactly $2^{k-1}$ worlds disagree with $\hat{s}_i$, so $\sum_S d_H(S, \hat{s}) = k \cdot 2^{k-1}$, i.e., for every realized value, $\mathbb{E}_S[\, d_H(S, \hat{s}) \,] = k/2$ identically; taking expectation over $\hat{S}$ on both sides and interchanging the finite summations gives $\mathbb{E}_S[\, \mathbb{E}_{\hat{S}}[\, d_H(S, \hat{S}) \,] \,] = k/2$, and from $\max \geq$ mean we get (ii); the world family is finite, so pigeonhole gives (iii). Step four: the identity and the presentation prefix do not depend on m, so "no guarantee at any finite m" holds in fact: the sample complexity is not some large number but the non-existence of a finite value. (\allowbreak{}Quantifier-order note: the choice of the bad world depends on m through $\hat{S}_m$, so "a fixed world fails at all m" is false; what is m-uniform is the identity together with the presentation prefix.) (\allowbreak{}Convention note: tier (a) also holds within Lemma A.25's sampling channel at q = 0 by A.25(i), so the three tiers can be compared within one channel model; under an adversarial-presentation convention, tier (b) would also have no finite bound (the adversary presents only non-boundary elements) so the tiers of this theorem are uniformly compared under the sampling convention, and the two conventions are not mixed.) (b) We argue within the sampling channel of Lemma A.25 (\allowbreak{}observations are drawn independently and uniformly from $\{e_1,\dots,e_k\}$, and each independently carries an authoritative label with probability q). Criterion Alpha is an expected-error-rate criterion; Criterion Beta is probabilistic --- no finite m recovers S deterministically in this channel (with probability $(1-q)^m > 0$ no label arrives at all, and then, by the symmetry of A.25(ii), no estimator can guarantee recovery) --- so Beta is defined as exact recovery of S with probability $\geq$ 2/3. \textbf{Lower bound.} After m observations, the expected number of distinct labeled coordinates is at most the expected number of labels, mq (the count of labels being Binomial(m, q)); by A.25(ii), each unlabeled coordinate contributes an average error of exactly 1/2 over the remaining world family, so under the uniform prior $\mathbb{E}[\,\mathrm{err}\,] \geq (1 - mq/k)/2$. Achieving $\mathrm{err} \leq \varepsilon$ thus forces $m \geq (1 - 2\varepsilon)\cdot k/q$ --- the lower bound $\Omega(k/q)$ for Criterion Alpha holds for all $\varepsilon < 1/2$, with an explicit constant. Criterion Beta: success with constant probability requires at most O(1) unlabeled coordinates (by A.25(ii), each unlabeled coordinate at most halves the conditional success probability), and the lower tail of coupon-collecting gives $m = \Omega((k \log k)/q)$. \textbf{Upper bound.} The learner outputs the label on labeled coordinates and flips an independent fair coin on the rest. The probability that coordinate $e_i$ is still unlabeled after m observations is $(1 - q/k)^m \leq e^{-qm/k}$, so the expected number of unlabeled coordinates is at most $k \cdot e^{-qm/k}$. Criterion Alpha: at $m = \lceil (k/q)\cdot \ln(2/\varepsilon) \rceil$ the expected number of unlabeled coordinates is at most $\varepsilon k/2$; the fair coin errs with rate exactly 1/2 on every unlabeled coordinate in every world, and labeled coordinates err by 0, so the expected error rate is at most $\varepsilon/4 \leq \varepsilon$ in every world, i.e., $m = O((k/q)\cdot \log(1/\varepsilon))$, which for constant $\varepsilon$ is $O(k/q)$. Criterion Beta: at $m = \lceil (2k/q)\cdot \ln k \rceil$, the probability that any coordinate remains unlabeled is at most $k \cdot e^{-qm/k}$ $\leq 1/k$, so with probability at least $1 - 1/k \geq 2/3$ (for $k \geq 3$) all coordinates are labeled and S is recovered exactly, i.e., $m = O((k \log k)/q)$. The two criteria's upper and lower bounds meet. (c) By C2, the archive records the promulgated content of N and exists prior to instances; by A1--A2, the extracted boundary agrees with the currently promulgated one; by A3, the coverage contains all boundary elements; the compliance status of every $e_i$ is given directly by table lookup. \hfill$\square$

\textbf{Labels}: all three tiers are mathematical facts (the premise of (a) is the modeling assumption H-N$'$; the premise of (b) is H-N$'$ together with the sampling-channel convention of Lemma A.25, with the dependence on constants and log factors written out in the proof; the premise of (c) is the criterion together with the working assumptions). Items registered for future rigorization (honestly marked per appendix convention): H-N$'$ can be relaxed to a constant-rate codebook subfamily (a prepared reply to the "assumption too strong" objection; under this retreat, the tier-(b) lower bound must switch to a Fano argument); the domain of err can be extended to all of O (requiring a consistency constraint).

\textbf{Anti-confusion note (why a label stream is not a weak archive).} The gap between a reliable label stream and an archive lies not in the reliability of the answers but in the temporal structure of the supply: labels arrive one by one with the instances, so the compliance status of coordinate $e_i$ must wait for its label to appear; the archive is the whole table: all boundary elements are simultaneously decidable before the first instance arrives. Hence even a streaming channel with q = 1 still needs $\Theta(k \log k)$ observations to recover S exactly (tier (b), Criterion Beta), and it can never attain zero-shot eligibility: labels exist only after instances; the archive exists before instances. The latter is the projection of S1 of Proposition 4.2 onto the sample-complexity axis; the former does not conflict with the merger of the "degenerate promulgation" of the A.19(iii) note on the extraction-complexity axis --- the two axes are tiered separately.

\textbf{Model-boundary note.} A different channel (the label-noise model, where labels always arrive but are noisy with a 1/2 $\pm$ q bias) scales as $\Omega(1/q^2)$, a different reading of q from this theorem's tier-(b) "label arrival rate" (1/q); the two are not mixed. What is modeled here must also be kept distinct from neighboring concepts: tier (b) models the authoritative per-event labeling registered in the A.19(iii) note (the degenerate-promulgation form), sparse and reliable; accident reports belong to the happens-to-be-readable tier of A.19(ii), from which norms must be reconstructed.

\textbf{Relation to Proposition A.13 (anti-duplication declaration).} A.13(ii) (the Modeling Corollary) is the limiting case of tier (a) at k = 1, m $\to$ $\infty$: two worlds share the same presentation stream while their boundaries differ, so identification in the limit is impossible. Tier (a) is its finite-sample strengthening: for the parameterized $2^k$ worlds, non-identifiability holds at every finite m (the Bayes-risk identity), and the pigeonhole corollary supplies the "fails infinitely often" that connects with the limiting form. Tiers (b) and (c) have no counterparts in A.13: A.13 contains neither a tier of partial authoritative information nor an archive-side upper bound. The two share the same premise (Obs does not depend on boundary candidates) but their conclusions sit at different levels: A.13 is an unattainability boundary; this theorem is a price list of sample complexity.

\textbf{Remark.} This theorem gives a quantitative reading of "how many samples the archive is worth": a three-stage jump from no finite bound (no labels), to $\Theta(k/q)$ (a sparse label stream), to 0 instance samples (whole-table lookup), with the archive as the only bridge across the jump. The information-theoretic half-sentence of bet P2 in Chapter 7 is thereby supplied: the cold-start gap between the archive side and the posterior side is structural; its empirical half-sentence (that this structural gap is not in fact ground away by data engineering and model scale in engineering deployments) is left for testing.

\subsection{A.7 The layering lower bound: the formal twin of Structural Characterization 5.1}
\textbf{Notation note.} In this section k denotes the number of layers (non-empty equivalence classes) and has nothing to do with the number of boundary elements k of A.6.

\textbf{Definition A.27 (\allowbreak{}discipline functions and carriers).} Let the total content of the framework be the set $\mathcal{C}$. The revision discipline is $\rho: \mathcal{C} \to \{F, I\}$, taking its value from how the region the content belongs to changes within its established service range (F: the region's content closes into a set between promulgation events, and any change must go through a promulgation event; I: the region runs a standing intake channel (a port of entry or a landing ground) as constitutive equipment; the test is entry by entry: $\rho(x)=I$ if and only if $x$ falls within the service range of some standing intake channel, and the existence of the channel takes priority over the manner in which change arrives; entry-level append-only discipline has nothing to do with $\rho$: existing entries of L2/L3/L4 are never rewritten, and what $\rho$ distinguishes is whether growth has a standing channel as its normal state); the denotation discipline is $\delta: \mathcal{C} \to \{T, N\}$ (T: denoting types, symbols openly instantiable; N: denoting instances, content anchored to concrete objects or phenomena --- anchoring via object-class codes is allowed, as with B.1's knowledge entries, rather than denoting abstract types). A carrier is an equivalence class of the relation $x \sim y \iff \rho(x)=\rho(y) \land \delta(x)=\delta(y)$; the number of layers k of the framework is defined as the number of non-empty equivalence classes.

\textbf{Lemma A.28 (the four cells are non-empty).} Under the construction goals G1--G4 (\allowbreak{}Definition 5.2 of the main text), the four cells (F,T), (I,T), (F,N), (I,N) are all occupied.

\textbf{Proof.} The conjunctive structure of G1 $\wedge$ G2 already carries the entire Cartesian product; cell-by-cell instantiation is given in Lemma 5.1 of the main text. The duty of G3 and G4 is to justify the $\delta$ axis, not to count; see "one honest registration" in Section 5.4 of the main text. \hfill$\square$

\textbf{Structural Characterization A.29 (the layering lower bound; the formal twin of Structural Characterization 5.1).} A prior cognitive framework satisfying the Constitution Criterion has a number k of non-empty equivalence classes of content with $k \geq 4$.

\textbf{Proof.} By Lemma A.28, each of the four cells has content; by Definition A.27, one equivalence class carries only one pair ($\rho$, $\delta$). Suppose k < 4; then by the pigeonhole principle, at least one equivalence class must contain the content of two cells, among which there must be two entries differing in $\rho$ or $\delta$ --- contradicting the definition of an equivalence class. \hfill$\square$

\textbf{Nature statement (labeled per the three-tier declaration).} The proof of this characterization is a combinatorial fact; the necessity of the premises G1--G4 is the normative argument of the main text ("what counts as a qualified prior framework" is judgment, not derivable from logic). The complete form of this characterization is a conditional: grant that each of the four goals is necessary, and the layered skeleton is necessary. Note: since $\rho$ and $\delta$ are both two-valued, there are at most four equivalence classes, so formally what is obtained is exactly k = 4; the "lower bound" semantics is directed at further subdivision of the carriers --- any compliant subdivision can only increase the region count, and $k \geq 4$ thereby holds as a lower bound on all compliant subdivisions. A further note: if a third discipline axis appears in the future, the skeleton expands accordingly; that would be a generalization of this characterization, not the present situation (the fifth-goal assignability test of Section 5.2 is exactly this case, and structural corollary I of Chapter 7 takes it as the falsification condition).

\subsection{Bibliographic anchors}
Makinson \& van der Torre (2000, 2001) (input/output logic and constrained output); Alchourrón, Gärdenfors \& Makinson (1985) (AGM revision); Gold (1967) (\allowbreak{}identification in the limit from positive data); Angluin (1980) (positive results on learning from positive data); Chickering (1996) (NP-completeness of Bayesian-network structure learning); Le Cam (1973) (the two-point method for minimax lower bounds); Yuan \& Yao (2026) (COIN, iFuture, Online First, 9710001, doi:10.26599/\allowbreak{}IF.2026.9710001).

\section{Appendix B. Framework Demonstrations}
\subsection{Preamble: the existence statement}
\textbf{Statement}: A prior cognitive framework satisfying this paper's structure is actually running in industrial facility operations across five major sectors: civil public buildings (shopping malls, hotels, office towers, hospitals), industrial manufacturing (\allowbreak{}semiconductors, lithium batteries, electronics), large scientific instruments (a nuclear-fusion experimental facility, communications satellites), municipal infrastructure (district heating networks), and data centers. In cross-industry migration the concept layer remains verbatim unchanged; entering a new industry means "only adding blocks, never changing the shape," and the marginal cost of modeling decreases with the number of industries. And as Section 5.3 states, onboarding a new industry has required only instantiating known failure types, never inventing new ones. \textbf{Commitment} --- the framework's cross-domain invariance is a publicly testable structural corollary (\allowbreak{}structural corollary I of Chapter 7): if onboarding any new industry ever requires modifying, rather than augmenting, existing concept-layer entries, this chapter's claims are thereby damaged.

\textbf{The two cases.} This appendix presents two completed cases as constructive evidence; they occupy the two extremes of deployment conditions:

\begin{itemize}
\item \textbf{Appendix B.1, the cooling-plant case} (four-layer carrier instantiation with one complete inference): an everyday facility, with a piping system, with a really existing knowledge-base artifact. L1--L3 are real artifacts, checkable entry by entry; the fault event, the handling process, and the time-series readings of B.1.4 are demonstration runs. What is shown is \emph{real-time cause-tracing}: how one alarm collapses from readings along the vertical edges to its cause in the concept layer, and then returns to zero with the handling.
\item \textbf{Appendix B.2, the Curiosity rover drill-feed anomaly} (Sol 1536, reconstructed from NASA public materials): no piping system, the cold-start extreme with zero posterior data. With NASA's after-the-fact investigation report as the historical base (the onboard fault protection is real, the cause-tracing inference is a reconstruction) what is shown is an \emph{after-the-fact replay}: had such a prior framework and reasoning engine existed at the time, how the fault would have been detected, reasoned about, and converged.
\end{itemize}
The two cases form a duality: in B.1 the artifact is real and the event staged; in B.2 the event is real and the artifact staged --- each is true along the dimension the other constructs, and together they cover both evidence types: existing artifacts and existing events.

\textbf{Cross-domain convention.} For all L1/L2 content identical in semantics, B.2 carries over B.1's parts of speech, codes, and enumerations verbatim, marked "carried over from B.1" --- two industries (building cooling, deep-space exploration) share the same syntax layer and the common part of the concept layer, with zero diff. This is the case-level cashing-out of structural corollary I of Chapter 7.

\textbf{Methodological status.} The two cases are not application displays but constructive evidence: they prove that the class "four-layer frameworks satisfying the criterion" is non-empty (the existence side of Structural Characterization 5.1) and they demonstrate the shape Proposition 5.2's decidable reduction takes in an actual fault. For the comparison evidence from external standards, see Appendix C --- four-layer reverse readings of eight independent lineages.

\section{Appendix B.1. The Cooling-Plant Case: Four-Layer Carriers and One Complete Inference}
A delivered equipment-failure-analysis knowledge base happens to have exactly four layers, and each layer's revision discipline and denotation discipline correspond one-to-one with the four goal pairs derived in Chapter 5. This appendix takes the chilled-water-pump branch of that base: B.1.1--B.1.3 give the actual content of the first three layers; B.1.4 gives one case and its complete inference from sensor readings to risk event and back to zero; B.1.5 verifies the criterion item by item. The knowledge base was built before this paper's framework; its layering was not designed to satisfy Structural Characterization 5.1. \emph{The boundary between the real and the staged is declared up front: the content and structure of L1--L3 are really existing artifacts, checkable entry by entry; the fault event, handling process, and time-series readings of B.1.4 are demonstration runs. What is verified is the relational discipline among the four layers, not some real fault of some real pump (for the complete boundary statement, see B.1.6).}

The scenario is the chilled-water pump (object class \texttt{ACCCCP}) under the central cooling system (object class \texttt{ACCC}) of data-center zone A.

\subsection{B.1.1 The L1 syntax layer (stable $\times$ abstract)}
The syntax layer prescribes which kinds of things may exist and which kinds of connections are allowed among them. It denotes nothing concrete.

\textbf{Vocabulary table (13 items)}

\begin{table}[htbp]
\centering
\small
\begin{tabular}{@{}>{\raggedright\arraybackslash}p{0.473\textwidth}>{\raggedright\arraybackslash}p{0.473\textwidth}@{}}
\toprule
Part of speech & Grammatical role \\
\midrule
\texttt{Physical\allowbreak{}Object} & bearer of the physical world \\
\texttt{Composition\allowbreak{}Relation\allowbreak{}Type} & part--whole structure \\
\texttt{Mapping\allowbreak{}Relation\allowbreak{}Type} & cross-domain correspondence \\
\texttt{Attribute\allowbreak{}Type} & structural bearer of composite attributes \\
\texttt{Failure\allowbreak{}Mode} & the mode of action of a failure \\
\texttt{Trigger\allowbreak{}Factor} & exogenous disturbance source (can only be a cause, never an effect) \\
\texttt{Failure\allowbreak{}Reasoning\allowbreak{}Node} & node of an equipment-level causal tree \\
\texttt{System\allowbreak{}Reasoning\allowbreak{}Element} & node of a system-level propagation graph \\
\texttt{Causal\allowbreak{}Relation\allowbreak{}Type} & causal edge (reified) \\
\texttt{RiskEvent} & the anchor from physical failure to business loss \\
\texttt{Physical\allowbreak{}Operation\allowbreak{}Business} & human/machine actions on the physical world \\
\texttt{Management\allowbreak{}Standard\allowbreak{}Knowledge} & bearer of thresholds and criteria \\
\texttt{Computation\allowbreak{}Input\allowbreak{}Knowledge} & external input parameters of reduction computations \\
\bottomrule
\end{tabular}
\end{table}
\textbf{Connector table (17 items)} --- the only connections allowed anywhere in the base; every assertion must draw from it

\begin{table}[htbp]
\centering
\small
\begin{tabular}{@{}>{\raggedright\arraybackslash}p{0.473\textwidth}>{\raggedright\arraybackslash}p{0.473\textwidth}@{}}
\toprule
Connector & Grammatical function \\
\midrule
\texttt{has\allowbreak{}Attribute} & attaches a composite attribute structure \\
\texttt{has\allowbreak{}Component} & composition structure \\
\texttt{manifests\allowbreak{}As} & reasoning node $\to$ failure mode \\
\texttt{relates\allowbreak{}From} / \texttt{relatesTo} & the two ends of a generic directed relation \\
\texttt{has\allowbreak{}Failure\allowbreak{}Node} & model $\to$ its nodes \\
\texttt{has\allowbreak{}Relation\allowbreak{}From} / \texttt{has\allowbreak{}Relation\allowbreak{}To} & head and tail of a causal edge \\
\texttt{realizes} & failure $\to$ risk event (the only passage from the physical layer to the business layer) \\
\texttt{applied\allowbreak{}To\allowbreak{}Entity\allowbreak{}Type} & knowledge $\to$ the object class it applies to \\
\texttt{propagation\allowbreak{}Applicable\allowbreak{}Conditions} / \texttt{propagation\allowbreak{}Excluded\allowbreak{}Conditions} & whitelist / blacklist gating of propagation \\
\texttt{occur\allowbreak{}Applicable\allowbreak{}Conditions} / \texttt{occur\allowbreak{}Excluded\allowbreak{}Conditions} & whitelist / blacklist gating of occurrence \\
\texttt{constraint\allowbreak{}Info\allowbreak{}Point} & the information point a constraint relies on \\
\texttt{references\allowbreak{}Threshold\allowbreak{}Config} & judgment $\to$ threshold configuration \\
\texttt{references\allowbreak{}Computation\allowbreak{}Input} & judgment $\to$ computation input \\
\bottomrule
\end{tabular}
\end{table}
Membership in a part of speech is decided mechanically by shape constraints (anything whose parent class is the knowledge root class is a part of speech) and depends on no one's adjudication. Changing the syntax layer would amount to a change of language generation: the whole base would have to be reinterpreted. It is therefore naturally frozen in engineering practice.

\subsection{B.1.2 The L2 concept layer (open $\times$ abstract)}
Under the parts of speech, the concept layer gives the kinds the industry actually distinguishes. It still denotes no individual.

\begin{table}[htbp]
\centering
\small
\begin{tabular}{@{}>{\raggedright\arraybackslash}p{0.473\textwidth}>{\raggedright\arraybackslash}p{0.473\textwidth}@{}}
\toprule
Part of speech & Concepts under it \\
\midrule
\texttt{Physical\allowbreak{}Object} & equipment types, component types, medium-unit types, entity-combination types, space types, system types, tool types, instrument types, operation-consumable types, operation-spare-part types, protective-equipment types \\
\texttt{Failure\allowbreak{}Mode} & physical-system failure mechanisms, medium-state failure mechanisms, system transfer factors, environmental factors \\
\texttt{Trigger\allowbreak{}Factor} & personnel-operation factors, rule-constraint factors \\
\texttt{Failure\allowbreak{}Reasoning\allowbreak{}Node} & failure-hazard nodes, failure-inducement nodes, failure-consequence nodes \\
\texttt{System\allowbreak{}Reasoning\allowbreak{}Element} & failure propagation channels, system-failure input nodes, in-system equipment-failure nodes, system-failure output nodes, system-failure state nodes \\
\texttt{RiskEvent} & space risk events, system risk events, equipment risk events \\
\texttt{Physical\allowbreak{}Operation\allowbreak{}Business} & patrol \& observation, monitoring \& judgment, evaluation \& verification, operation \& regulation, repair \& maintenance, emergency \& handling, installation \& replacement, disassembly \& decommissioning \\
\texttt{Management\allowbreak{}Standard\allowbreak{}Knowledge} & level thresholds, rate thresholds, cumulative-jump thresholds, trend-increment thresholds, count thresholds, confidence thresholds \\
\texttt{Computation\allowbreak{}Input\allowbreak{}Knowledge} & external-parameter types, condition-concept types, measurement-input types \\
\texttt{Attribute\allowbreak{}Type} & 60 composite attribute structure classes (transfer-factor category combinations, applicable-object-type degree levels, graded state descriptions, propagation-state rules, etc.) \\
\bottomrule
\end{tabular}
\end{table}
The definition of \textbf{entity-combination type} must be listed separately: it "does not represent a real component that is physically detachable on its own, but expresses the logical combination a group of entities forms around some functional goal." \texttt{ECP-MECH} (the mechanical transmission chain) does not exist in the basic physical world; it exists because someone delimited it as one whole for a functional purpose --- a trace of C1 at the data-structure level.

\textbf{The reduction basis.} The concept layer fixes the reduction endpoints of failures by a closed enumeration:

\begingroup\footnotesize
\begin{verbatim}
Physical failure-mode categories (10): structural rupture, seal leakage,
    parameter drift, function interruption, degradation & wear,
                                       channel blockage, system oscillation,
                                           constraint violation, instability
                                               amplification, connection
                                                   rupture
Medium failure-mode categories (9):    composition contamination,
    concentration deviation, phase-state anomaly, physical-property
        degradation, chemical deactivation,
                                       structural decomposition, adsorption
                                           saturation, usable exhaustion,
                                               medium instability
Transfer types (4):                    energy, matter, information, mechanical
Environmental-factor categories (9):   ambient temperature, ambient humidity,
    mechanical environment, electromagnetic interference, spatial constraint,
                                       medium intrusion, foreign-object
                                           intrusion, chemical environment,
                                               radiation environment
\end{verbatim}
\endgroup
Thirty-two categories in all. The basis is closed, the concept set is open: concrete mechanisms (e.g., "bearing seizure") can be augmented without limit, and each must declare which basis category it falls under. Closed is said of the reduction test (\allowbreak{}candidates are enumerated only within the basis); extensible is said of the augmentation passage (the basis can gain categories through the port of entry of Section 5.2). The two do not conflict: the basis is extensible, but extension adds values rather than altering them --- the foundation does not move; concepts are only added, never changed. The basis being finite and closed gives every failure's reduction path an upper bound on its length, which directly supports Proposition A.16. Trigger factors can only be leaves of the causal tree and \emph{are not classified into the failure-mode basis (physical 10 / medium 9)}: environmental inducements go to the \emph{nine environmental-factor categories within the 32-item basis}; personnel/rule inducements go to the TriggerFactor concept enumeration (personnel-operation factors, rule-constraint factors): promulgated \emph{outside} the 32-item basis as the dedicated vocabulary of inducements.

\textbf{Graded state semantics.} The concept layer also fixes the sole currency passed between layers: \texttt{0} normal, \texttt{0.25} adverse trend, \texttt{0.5} symptom, \texttt{1} failed. Continuous physical quantities live only in L4; before crossing into L3 they must first be reduced to a degree level.

\textbf{Graph-theoretic constraint.} A trigger factor can only be a leaf in the causal tree --- it can serve as a cause, never as an effect or an intermediate node. This guarantees that the causal tree has a root, and that cause-tracing necessarily terminates.

\subsection{B.1.3 The L3 knowledge layer (stable $\times$ concrete): the knowledge this case uses}
Knowledge-layer entries name object classes, failure modes, and concrete values, but denote no physical individual and do not change with any individual's readings. Their discipline is clause-by-clause freezing: the set is forever open, every entry forever unchanged; a piece of knowledge found to be wrong is not edited but deprecated, with a new entry under a new code, because existing reasoning traces cite the old code.

\subsubsection{B.1.3.1 The assembly of three blocks of knowledge}
\begin{figure}[htbp]
\centering
\begin{tikzpicture}[
  font=\scriptsize,
  node distance=0.3cm,
  box/.style={draw, rounded corners=2pt, align=center, inner sep=3pt, text width=1.7cm},
  evt/.style={draw, rounded corners=6pt, align=center, inner sep=3pt, text width=1.7cm},
  grp/.style={draw=gray, dashed, inner sep=5pt, rounded corners=3pt},
  glab/.style={font=\scriptsize\itshape, text=gray},
  arr/.style={-{Stealth}, thick},
  elab/.style={font=\tiny, midway, fill=white, inner sep=1pt}
]
\node[box] (DIP) {DIP information-point slots};
\node[box, right=of DIP] (SEE) {SEE reduction formula};
\node[box, right=of SEE] (LTT) {LTT threshold configuration};
\node[box, right=of LTT] (JRI) {JRI judgment rule};
\node[box, right=of JRI] (JC) {JC judgment conclusion};
\node[box, right=of JC] (SOTM) {SOT degree-level mapping};
\node[box, right=of SOTM] (ARNP) {ARNP routing pattern};
\draw[arr] (DIP) -- (SEE);
\draw[arr] (SEE) -- (LTT);
\draw[arr] (LTT) -- (JRI);
\draw[arr] (JRI) -- (JC);
\draw[arr] (JC) -- (SOTM);
\draw[arr] (SOTM) -- (ARNP);
\node[grp, fit=(DIP)(SEE)(LTT)(JRI)(JC)(SOTM), label={[glab]above:{Monitoring \& judgment chain}}] (MC) {};
\node[box, below=1.5cm of DIP] (TN) {TN-P2 outlet valve mistakenly closed};
\node[box, right=1.1cm of TN] (HZ1) {HZ-P1 lubricant property degradation};
\node[box, right=of HZ1] (HZ2) {HZ-P2 bearing seizure};
\node[box, right=of HZ2] (HZ5) {HZ-P5 rotor mechanical jam};
\node[box, right=1.1cm of HZ5] (HZ0) {HZ-P0 unable to supply water};
\draw[arr] (TN.south) .. controls +(0,-0.7) and +(0,-0.7) .. node[elab, pos=0.5] {CR-P7} (HZ0.south);
\draw[arr] (HZ1) -- node[elab] {CR-P1} (HZ2);
\draw[arr] (HZ2) -- node[elab] {CR-P2} (HZ5);
\draw[arr] (HZ5) -- node[elab] {CR-P4} (HZ0);
\draw[arr] (ARNP) -- (HZ0);
\node[grp, fit=(TN)(HZ1)(HZ2)(HZ5)(HZ0), label={[glab]above:{Failure evolution model}}] (FM) {};
\node[box, below=1.5cm of HZ2] (CQ) {CQ-P1 chilled-water supply interrupted};
\node[box, right=of CQ] (SIE) {SIE-C2 in-system equipment failure};
\node[box, right=of SIE] (SS) {SS-C1 cooling function failed};
\draw[arr] (CQ) -- (SS);
\draw[arr] (SIE) -- node[elab] {CR-C5} (SS);
\node[grp, fit=(CQ)(SIE)(SS), label={[glab]above:{System propagation}}] (SP) {};
\node[evt, right=1.1cm of SS] (RE2) {RE-02 system risk event};
\node[evt, below=0.9cm of RE2] (RE3) {RE-03 equipment risk event};
\draw[arr] (HZ0) -- (CQ);
\draw[arr] (HZ0) -- node[elab] {EFR-C2 bridge} (SIE);
\draw[arr] (HZ0.east) -- node[elab, pos=0.35] {realizes} (RE3.west);
\draw[arr] (SS) -- node[elab] {realizes} (RE2);
\end{tikzpicture}
\end{figure}
The monitoring-and-judgment chain channels instance-layer readings into degree levels; the failure evolution model traces causes inside the equipment; system propagation spills equipment failure into system failure and anchors business loss via \texttt{realizes}. The three blocks share one graded-state semantics; the interfaces pass only degree levels.

\subsubsection{B.1.3.2 The failure evolution model: nodes and edges}
\begin{table}[htbp]
\centering
\small
\begin{tabular}{@{}>{\raggedright\arraybackslash}p{0.098\textwidth}>{\raggedright\arraybackslash}p{0.098\textwidth}>{\raggedright\arraybackslash}p{0.098\textwidth}>{\raggedright\arraybackslash}p{0.098\textwidth}>{\raggedright\arraybackslash}p{0.098\textwidth}>{\raggedright\arraybackslash}p{0.098\textwidth}>{\raggedright\arraybackslash}p{0.098\textwidth}>{\raggedright\arraybackslash}p{0.098\textwidth}@{}}
\toprule
Depth & Scope & Node & Entity & Failure mode & Reduction category & Prior MTTF & Observation / maintenance plan \\
\midrule
1 & medium unit under equipment & \texttt{HZ-P1} & \texttt{MDU-OIL} lubricating oil & \texttt{FM-MED-01} lubricant property degradation & physical-property degradation failure & 17,520 h (Weibull) & manual low-frequency observation / periodic replacement \\
2 & component under equipment & \texttt{HZ-P2} & \texttt{CMP-BRG} pump bearing & \texttt{FM-PHY-01} bearing seizure & degradation \& wear failure & 43,800 h & IoT observation / replacement upon symptom observation \\
3 & entity combination under equipment & \texttt{HZ-P5} & \texttt{ECP-MECH} mechanical transmission chain & \texttt{FM-PHY-04} rotor mechanical jam & function interruption failure & 52,560 h & --- \\
4 & the equipment under analysis & \texttt{HZ-P0} & \texttt{ACCCCP} chilled-water pump & \texttt{FM-PHY-02} unable to supply water & function interruption failure & 35,040 h (\allowbreak{}exponential) & IoT observation / repair after failure \\
leaf & exogenous inducement & \texttt{TN-P2} & --- & \texttt{HOF-01} outlet valve mistakenly closed & personnel-operation factor (\allowbreak{}Trigger\allowbreak{}Factor enumeration) & --- & --- \\
\bottomrule
\end{tabular}
\end{table}
The edges (\texttt{CR-P1}, \texttt{CR-P2}, \texttt{CR-P4}, \texttt{CR-P7}; relation type "or" for all) are reified as independent entries, because the edges themselves must carry propagation conditions and confidence. Within a layer, relation graphs of arbitrary complexity are allowed; across layers there is only one kind of edge --- instantiation / collapse. The model contains several further branches (the cavitation branch \texttt{HZ-P6}, the mechanical-seal leakage branch \texttt{HZ-P3}, the planned-maintenance-shutdown inducement \texttt{TN-P3}, the cooling-source water input \texttt{IN-P4}, etc.); only the main chain is listed here. The branches appear in the exclusion steps of B.1.4.

\texttt{hazard\allowbreak{}Entity\allowbreak{}Code} takes the object-class code rather than the unit's serial number; \texttt{priorMTTF} is the prior lifetime of this class of pumps, not the remaining lifetime of any particular unit. \texttt{hazard\allowbreak{}Entity\allowbreak{}Scope} gives the four depth levels in one dimension, so the causal tree's depth is capped by the parts tree's depth --- this is where the \texttt{d = 4} (\allowbreak{}Hypothesis H4) in the antecedent of Proposition A.16 comes from. The spillover node \texttt{CQ-P1} reduces to \texttt{STF-02} (transfer type = matter). Every fault must ultimately land on one of the 32 words; if it cannot land, that is a missing concept in the concept layer (\texttt{Coverage\allowbreak{}Gap}; see Definition 4.4), not a strange event in the field.

\subsubsection{B.1.3.3 Gating and narration of degree levels}
\begin{table}[htbp]
\centering
\small
\begin{tabular}{@{}>{\raggedright\arraybackslash}p{0.306\textwidth}>{\raggedright\arraybackslash}p{0.306\textwidth}>{\raggedright\arraybackslash}p{0.306\textwidth}@{}}
\toprule
Entry & Type & Content \\
\midrule
\texttt{PC-HZ-P2} & propagation condition & applicability whitelist (DNF): \texttt{BUSINESS\_\allowbreak{}SCENE} / \texttt{HIGH\_\allowbreak{}COOLING\_\allowbreak{}LOAD > 40\%} \\
\texttt{PSRE-\allowbreak{}PC-\allowbreak{}HZ-\allowbreak{}P2-\allowbreak{}1/\allowbreak{}2} & degree-level mapping & \texttt{0.\allowbreak{}5 $\to$ 0.\allowbreak{}5}; \texttt{1 $\to$ 1} \\
\texttt{HDS-P0-L50} & graded state description & direct observation: "the chilled-water pump shows symptoms of declining water-supply capacity and is at risk of failure"; uploaded: "symptom state uploaded from subordinate nodes (bearing wear, seal leakage)" \\
\texttt{HDS-\allowbreak{}P0-\allowbreak{}L100} & graded state description & direct observation: "the chilled-water pump is unable to supply water"; uploaded: "failure state uploaded from subordinate nodes (rotor mechanical jam / electrical drive circuit function interrupted); the unit as a whole has lost water-supply capacity" \\
\bottomrule
\end{tabular}
\end{table}
Propagation rules are degree-level-to-degree-level mapping tables, not numerical functions: the knowledge layer does no numerical computation. The graded state description distinguishes, for the same degree level, two provenances, "obtained by direct observation" and "uploaded from subordinates", each paired with a human-readable sentence: the system records not only that the conclusion is 1, but also how this 1 came about and how it should be explained to a human.

\subsubsection{B.1.3.4 The monitoring business: the slots the knowledge layer opens toward the instance layer}
The business entry \texttt{POB\_\allowbreak{}CWP\_\allowbreak{}001\_\allowbreak{}HAZ\_\allowbreak{}TOP\_\allowbreak{}002} (\allowbreak{}monitoring \& judgment, executed in real time) strings eight links into one chain, each link an independently addressable entry:

\begin{table}[htbp]
\centering
\small
\begin{tabular}{@{}>{\raggedright\arraybackslash}p{0.306\textwidth}>{\raggedright\arraybackslash}p{0.306\textwidth}>{\raggedright\arraybackslash}p{0.306\textwidth}@{}}
\toprule
Link & Entry & Key content \\
\midrule
information-point slots & \texttt{DIP-\allowbreak{}P001 /\allowbreak{} P002 /\allowbreak{} P003} & \texttt{flow} instantaneous flow (trigger point), \texttt{inPress} inlet pressure, \texttt{runStatus} running state; object class \texttt{ACCCCP}, source \texttt{TWIN\_DATA}; \emph{slots only --- they contain no readings} \\
external input & \texttt{EPT-\allowbreak{}DESIGN-\allowbreak{}FLOW} & \texttt{rated\allowbreak{}Outlet\allowbreak{}Flow} rated outlet flow, carried with the object class \\
reduction & \texttt{SEE} & trigger type \texttt{THRESHOLD\_\allowbreak{}CROSSING}; formula \texttt{flow /\allowbreak{} rated\allowbreak{}Outlet\allowbreak{}Flow}, dimensionless result \\
threshold & \texttt{LTT-TR-1} & \texttt{THR\_\allowbreak{}CWP\_\allowbreak{}FLOW\_\allowbreak{}LOW =\allowbreak{} 0.\allowbreak{}8}, direction \texttt{DOWN\_\allowbreak{}CROSSING}, deadband 2\%, duration 60 s \\
rule & \texttt{JRI-\allowbreak{}RULE\_\allowbreak{}FLOW\_\allowbreak{}CRIT} & condition 1 \texttt{CTT-\allowbreak{}RULE\_\allowbreak{}FLOW\_\allowbreak{}CRIT-\allowbreak{}1}: \texttt{P003 EQ RUNNING}; condition 2 \texttt{CTT-\allowbreak{}RULE\_\allowbreak{}FLOW\_\allowbreak{}CRIT-\allowbreak{}2}: \texttt{P001 LT 8 m$^3$/\allowbreak{}h} \\
conclusion & \texttt{JC-\allowbreak{}FLOW\_\allowbreak{}LOSS\_\allowbreak{}CRITICAL} & "supply flow approaching zero; the pump has lost water-supply capacity"; exclusion group \texttt{G\_FLOW} \\
write-back & \texttt{SOT-\allowbreak{}FLOW\_\allowbreak{}LOSS\_\allowbreak{}CRITICAL-\allowbreak{}1} & (\texttt{ACCCCP}, \texttt{FM\_\allowbreak{}PUMP\_\allowbreak{}FLOW\_\allowbreak{}LOSS}) $\to$ \texttt{target\allowbreak{}Degree\allowbreak{}Level =\allowbreak{} 1} \\
routing & \texttt{ARNP-1} & locate the failure evolution model by the (object class, failure mode) pair \\
\bottomrule
\end{tabular}
\end{table}
The rule's two conditions are independently addressable entities, not a Boolean-expression string; \texttt{exclusion\allowbreak{}Group} guarantees that conclusions in the same group are mutually exclusive. \texttt{RULE\_\allowbreak{}FLOW\_\allowbreak{}CRIT} uses an absolute quantity (\texttt{flow < 8 m$^3$/\allowbreak{}h}); \texttt{RULE\_\allowbreak{}FLOW\_\allowbreak{}DEG} uses a relative ratio --- approaching zero is an absolute criterion, degradation a relative one, and this is a promulgated stipulation.

\subsubsection{B.1.3.5 System propagation and risk events}
\begin{table}[htbp]
\centering
\small
\begin{tabular}{@{}>{\raggedright\arraybackslash}p{0.306\textwidth}>{\raggedright\arraybackslash}p{0.306\textwidth}>{\raggedright\arraybackslash}p{0.306\textwidth}@{}}
\toprule
Entry & Type & Key content \\
\midrule
\texttt{CQ-P1} & failure-consequence node (\allowbreak{}propagation edge) & failure mode \texttt{STF-02} (transfer type = matter); scope: the system \texttt{ACCC} the equipment sits in \\
\texttt{SIE-C2} & in-system equipment-failure node & equipment class \texttt{ACCCCP}; bridges to \texttt{HZ-P0} via \texttt{EFR-C2-1/2}; gating \texttt{PC-SIE-C2} \\
\texttt{SS-C1} & system-failure state node & system class \texttt{ACCC}; primary failure mode \texttt{STF-04}; two input rules \texttt{SFR-C1-1/2} \\
\texttt{SO-C1} & system state object & the runtime degree-level bearer of \texttt{SS-C1}; returning to zero means its \texttt{degree\allowbreak{}Level} returns to \texttt{0}; records are append-only \\
\texttt{RE-03} & equipment risk event & anchored to \texttt{HZ-P0}; cascading failure, barrier = redundant switchover; loss structure \emph{binary model}; recovery path = redundant recovery \\
\texttt{RE-02} & system risk event & anchored to \texttt{SS-C1}; affects \texttt{ACATAH}, \texttt{SPC-OFFICE}; loss structure \emph{linear amplification} \\
\bottomrule
\end{tabular}
\end{table}
The difference in loss structures shows why risk events must be a part of speech of their own: the loss of an equipment failure is a step function, the loss of system cooling loss accumulates linearly with time, and the two loss structures can neither be described by one function nor stuffed back into the failure modes.

\subsection{B.1.4 The L4 instance layer (open $\times$ concrete): a demonstration case with the complete inference}
L4 does not exist in the knowledge-base files --- the knowledge base stores no operating data. L4's content is physical individuals, their time-series readings, injected static parameters, and runtime degree-level values on knowledge-layer nodes. Its discipline is append-only: every write is a new record; history is never overwritten.

\subsubsection{B.1.4.1 The case object}
\begin{table}[htbp]
\centering
\small
\begin{tabular}{@{}>{\raggedright\arraybackslash}p{0.306\textwidth}>{\raggedright\arraybackslash}p{0.306\textwidth}>{\raggedright\arraybackslash}p{0.306\textwidth}@{}}
\toprule
Content & This case & The knowledge-layer slot it lands in \\
\midrule
physical individual & chilled-water pump \#1, data-center zone A & instantiates object class \texttt{ACCCCP} \\
time-series readings & \texttt{flow}, \texttt{inPress}, \texttt{runStatus} & \texttt{DIP-\allowbreak{}P001 /\allowbreak{} P002 /\allowbreak{} P003} \\
static parameter injection & \texttt{rated\allowbreak{}Outlet\allowbreak{}Flow =\allowbreak{} 480 m$^3$/\allowbreak{}h}; \texttt{npsh\allowbreak{}Required\allowbreak{}Inlet\allowbreak{}Pressure =\allowbreak{} 85 k\allowbreak{}Pa} & \texttt{EPT-\allowbreak{}DESIGN-\allowbreak{}FLOW /\allowbreak{} -\allowbreak{}NPSH-\allowbreak{}PRESS} \\
runtime states & the \texttt{degree\allowbreak{}Level} of each hazard node & runtime fields of knowledge-layer nodes \\
\bottomrule
\end{tabular}
\end{table}
On July 15, 2026, zone A was at the peak of the cooling season, and pump \#1 carried the main water supply.

\subsubsection{B.1.4.2 Panorama: one closed loop of inter-layer traversal}
\begin{figure}[htbp]
\centering
\begin{tikzpicture}[
  font=\scriptsize,
  box/.style={draw, rounded corners=2pt, align=center, inner sep=4pt, text width=3.4cm},
  grp/.style={draw=gray, dashed, inner sep=8pt, rounded corners=3pt},
  glab/.style={font=\scriptsize\itshape, text=gray},
  arr/.style={-{Stealth}, thick},
  elab/.style={font=\tiny, midway, fill=white, inner sep=1.5pt, align=center}
]
\node[box] (R) {pump \#1 reading stream\\ \texttt{flow} / \texttt{inPress} / \texttt{runStatus}};
\node[box, below=1.5cm of R] (DG) {\texttt{HZ-P0.degreeLevel = 1}};
\node[box, below=1.5cm of DG] (X) {manual execution $\to$ entity reset $\to$ state return to zero};
\node[box, right=4.6cm of R] (M) {monitoring chain\\ DIP $\to$ SEE $\to$ LTT $\to$ JRI $\to$ JC};
\node[box, below=1.5cm of M] (S) {SOT degree-level mapping};
\node[box, below=1.5cm of S] (T) {ARNP routing $\to$ equipment cause-tracing $\to$ system propagation $\to$ realizes};
\node[box, below=1.5cm of T] (P) {POB measure entries};
\draw[arr] (R) -- node[elab] {Steps 1--2 instantiation} (M);
\draw[arr] (M) -- (S);
\draw[arr] (S) -- node[elab] {Step 5 collapse landing} (DG);
\draw[arr] (DG) -- (T);
\draw[arr] (T) -- (P);
\draw[arr] (P.south) |- node[elab, pos=0.75] {Step 10 instantiation} (X.south);
\draw[arr] (X.west) -- ++(-0.9,0) |- (R.west);
\node[grp, fit=(R)(DG)(X), label={[glab]above:{L4 instance layer}}] (L4) {};
\node[grp, fit=(M)(S)(T)(P), label={[glab]above:{L3 knowledge layer}}] (L3) {};
\end{tikzpicture}
\end{figure}
The closed loop crosses layers at only two positions: where readings enter the knowledge layer's slots, and where degree levels / measures fall back to the instance layer. Everything else is graph traversal inside the knowledge layer.

\subsubsection{B.1.4.3 The inference process}
\begin{figure}[htbp]
\centering
\begin{tikzpicture}[
  font=\scriptsize,
  box/.style={draw, rounded corners=2pt, align=left, inner sep=4pt, text width=4.35cm},
  arr/.style={-{Stealth}, thick}
]
\node[box] (A) {\textbf{08:06}\\ flow 462 $|$ ratio 0.963\\ normal};
\node[box, right=0.55cm of A] (B) {\textbf{08:11}\\ flow 365 $|$ ratio 0.760\\ crossed down, debounce unmet $\to$ not triggered};
\node[box, right=0.55cm of B] (C) {\textbf{08:12}\\ flow 310 $|$ ratio 0.646\\ sustained $>60$ s $\to$ SEE triggered};
\node[box, below=0.8cm of C] (D) {\textbf{08:14}\\ flow 2.8 $|$ still RUNNING\\ \texttt{RULE\_FLOW\_CRIT} hit $\to$ \texttt{HZ-P0 = 1}};
\node[box, left=0.55cm of D] (E) {\textbf{08:14--08:16}\\ cause-tracing \texttt{HZ-P1} $\to$ \texttt{HZ-P2} $\to$ \texttt{HZ-P5} $\to$ \texttt{HZ-P0}\\ propagation \texttt{CQ-P1} $\to$ \texttt{SS-C1 = 1}};
\node[box, left=0.55cm of E] (F) {\textbf{08:16}\\ realizes $\to$ \texttt{RE-03} / \texttt{RE-02} activated};
\node[box, below=0.8cm of F] (G) {\textbf{09:05--11:40}\\ manual bearing replacement (staged event)\\ teardown: grease emulsified, raceway spalling (staged setting)};
\node[box, right=0.55cm of G] (H) {\textbf{12:05}\\ flow 448 $|$ ratio 0.933\\ \texttt{RULE\_FLOW\_OK} $\to$ \texttt{HZ-P0 = 0}};
\node[box, right=0.55cm of H] (I) {\textbf{12:08}\\ \texttt{CQ-P1} / \texttt{SS-C1} / \texttt{SO-C1} return to zero\\ \texttt{RE-02} / \texttt{RE-03} released};
\draw[arr] (A) -- (B);
\draw[arr] (B) -- (C);
\draw[arr] (C) -- (D);
\draw[arr] (D) -- (E);
\draw[arr] (E) -- (F);
\draw[arr] (F) -- (G);
\draw[arr] (G) -- (H);
\draw[arr] (H) -- (I);
\end{tikzpicture}
\par\vspace{4pt}\noindent{\small\textbf{Figure.} The inference timeline of the demonstration run (caption translated as above).\par}
\end{figure}
\begin{longtable}{@{}>{\raggedright\arraybackslash}p{0.140\textwidth}>{\raggedright\arraybackslash}p{0.140\textwidth}>{\raggedright\arraybackslash}p{0.140\textwidth}>{\raggedright\arraybackslash}p{0.140\textwidth}>{\raggedright\arraybackslash}p{0.140\textwidth}>{\raggedright\arraybackslash}p{0.140\textwidth}@{}}
\toprule
Step & Time & Action & Inter-layer edge & Entries relied on & Result \\
\midrule
\endfirsthead
\toprule
Step & Time & Action & Inter-layer edge & Entries relied on & Result \\
\midrule
\endhead
1 & 08:06 & readings land & \textbf{L4 $\to$ L3 instantiation (slot filling)} & \texttt{DIP-\allowbreak{}P001 /\allowbreak{} P002 /\allowbreak{} P003} & \texttt{flow=462}, \texttt{in\allowbreak{}Press=\allowbreak{}128}, \texttt{run\allowbreak{}Status=\allowbreak{}RUNNING} assigned into the three slots by object class and point code \\
2 & 08:06 & reduction & \textbf{L4 $\to$ L3 instantiation (value $\to$ ratio)} & \texttt{SEE} formula + \texttt{EPT-\allowbreak{}DESIGN-\allowbreak{}FLOW} & \texttt{462 /\allowbreak{} 480 =\allowbreak{} 0.\allowbreak{}963} \\
3 & 08:11 / 08:12 & threshold crossing & within L3 & \texttt{LTT-TR-1} (\texttt{0.8}, down-crossing, 60 s, 2\%) & at 08:11 ratio \texttt{0.760} had crossed down but debounce was unmet (\emph{not triggered}); at 08:12 ratio \texttt{0.646} sustained over 60 s: \texttt{SEE} triggered \\
4 & 08:14 & rule adjudication & within L3 & the two conditions of \texttt{JRI-\allowbreak{}RULE\_\allowbreak{}FLOW\_\allowbreak{}CRIT} & conclusion \texttt{FLOW\_\allowbreak{}LOSS\_\allowbreak{}CRITICAL} \\
5 & 08:14 & degree-level write-back & \textbf{L3 $\to$ L4 collapse landing} & \texttt{SOT-\allowbreak{}FLOW\_\allowbreak{}LOSS\_\allowbreak{}CRITICAL-\allowbreak{}1} & \texttt{HZ-\allowbreak{}P0.\allowbreak{}degree\allowbreak{}Level} refreshed from \texttt{0} to \texttt{1}; \texttt{HDS-\allowbreak{}P0-\allowbreak{}L100} supplies the human-readable sentence; BMS level-1 alarm \\
6 & 08:14 & model routing (three routes) & within L3 (cross-model) & \texttt{ARNP-1} & (1) enter equipment cause-tracing; (2) \texttt{EFR-C2} bridges \texttt{HZ-P0} $\leftrightarrow$ \texttt{SIE-C2}; (3) the cavitation branch \texttt{RULE\_CAVI} not hit --- \texttt{HZ-P6} stays \texttt{0} with a "not activated" record \\
7 & 08:14--08:16 & cause-tracing: exclude first, then hit & within L3 (causal graph) & \texttt{CR-\allowbreak{}P7 /\allowbreak{} P8 /\allowbreak{} P14} excluded; \texttt{CR-\allowbreak{}P1 /\allowbreak{} P2 /\allowbreak{} P4} hit & chain \texttt{HZ-\allowbreak{}P1(0.\allowbreak{}5) $\to$ HZ-\allowbreak{}P2(1) $\to$ HZ-\allowbreak{}P5(1) $\to$ HZ-\allowbreak{}P0(1)}, confidence 0.91 (\allowbreak{}demonstration value, not measured) \\
8 & 08:14--08:16 & cross-model propagation (two parallel channels) & within L3 & \texttt{CQ-\allowbreak{}P1 $\to$ SS-\allowbreak{}C1} (matching \texttt{SFR-C1-1}); \texttt{SIE-\allowbreak{}C2 $\to$ CR-\allowbreak{}C5 $\to$ SS-\allowbreak{}C1} & \texttt{CQ-P1 = 1}, \texttt{SS-C1 = 1} (cooling function of the central cooling system failed) \\
9 & 08:16 & business realization & within L3 (\texttt{realizes}) & \texttt{realizes} & \texttt{RE-03} (equipment level, binary loss) and \texttt{RE-02} (system level, linear amplification, affecting the AC terminals and office space) activated simultaneously \\
10 & 09:05--12:08 & measure execution and return to zero & \textbf{L3 $\to$ L4 instantiation (knowledge $\to$ action)} & \texttt{POB\_\allowbreak{}CWP\_\allowbreak{}001\_\allowbreak{}HAZ\_\allowbreak{}L3\_\allowbreak{}BEARING\_\allowbreak{}JAM\_\allowbreak{}002} (\allowbreak{}containing recovery action \texttt{RA\_\allowbreak{}CWP\_\allowbreak{}RPL\_\allowbreak{}RESTORE}) & bearing replaced; at 12:05 \texttt{RULE\_\allowbreak{}FLOW\_\allowbreak{}OK} hit, \texttt{HZ-P0 = 0}; at 12:08 \texttt{CQ-P1}, \texttt{SS-C1}, \texttt{SO-C1} return to zero in sequence \\
\bottomrule
\end{longtable}
\textbf{Details of the two adjudications.} In Step 4 the rule set is matched entry by entry, and \texttt{exclusion\allowbreak{}Group} guarantees only one conclusion per group:

\begin{table}[htbp]
\centering
\small
\begin{tabular}{@{}>{\raggedright\arraybackslash}p{0.306\textwidth}>{\raggedright\arraybackslash}p{0.306\textwidth}>{\raggedright\arraybackslash}p{0.306\textwidth}@{}}
\toprule
Rule & Condition & This time \\
\midrule
\texttt{RULE\_\allowbreak{}FLOW\_\allowbreak{}OK} & \texttt{flow/\allowbreak{}rated $\geq$ 0.\allowbreak{}9} & no (0.006) \\
\texttt{RULE\_\allowbreak{}FLOW\_\allowbreak{}DEG} & RUNNING and \texttt{0.\allowbreak{}8 > flow/\allowbreak{}rated} and \texttt{flow $\geq$ 8} & no (\texttt{flow < 8}) \\
\texttt{RULE\_\allowbreak{}FLOW\_\allowbreak{}CRIT} & RUNNING and \texttt{flow < 8 m$^3$/\allowbreak{}h} & \textbf{yes} (\texttt{CTT-\allowbreak{}RULE\_\allowbreak{}FLOW\_\allowbreak{}CRIT-\allowbreak{}1}: \texttt{P003 =\allowbreak{} RUNNING}; \texttt{CTT-\allowbreak{}RULE\_\allowbreak{}FLOW\_\allowbreak{}CRIT-\allowbreak{}2}: \texttt{2.8 < 8}) \\
\bottomrule
\end{tabular}
\end{table}
Step 7 first excludes two inducement-direct paths and one external-input branch (\allowbreak{}inducement nodes are leaves; if one holds, cause-tracing terminates immediately):

\begin{table}[htbp]
\centering
\small
\begin{tabular}{@{}>{\raggedright\arraybackslash}p{0.306\textwidth}>{\raggedright\arraybackslash}p{0.306\textwidth}>{\raggedright\arraybackslash}p{0.306\textwidth}@{}}
\toprule
Causal edge & Path & Grounds for exclusion \\
\midrule
\texttt{CR-P7} & \texttt{TN-P2} (outlet valve mistakenly closed) $\to$ \texttt{HZ-P0} & valve-position feedback fully open; 07:30 patrol record shows normal valve position \\
\texttt{CR-P8} & \texttt{TN-P3} (planned maintenance shutdown) $\to$ \texttt{HZ-P0} & no maintenance work order; \texttt{P003 =\allowbreak{} RUNNING} \\
\texttt{CR-P14} & \texttt{IN-P4} (cooling-source water parameter anomaly) $\to$ \texttt{HZ-P3} & cooling-plant supply temperature normal; \texttt{IN-\allowbreak{}P4.\allowbreak{}degree\allowbreak{}Level =\allowbreak{} 0} \\
\bottomrule
\end{tabular}
\end{table}
All candidate paths come from the knowledge layer and are finitely enumerable; all grounds for exclusion come from the instance layer. Not one step relies on "the model finding something unlikely."

\textbf{Seven points must be singled out:}

\begin{itemize}
\item \textbf{The reduction boundary.} The formula lives in the knowledge layer, the values come from the instance layer, and the external parameters have their sources explicitly declared via \texttt{references\allowbreak{}Computation\allowbreak{}Input}. The three-way separation makes disputes answerable item by item --- a wrong reading, a misconfigured rated value, and a wrongly chosen formula each have their own object of inspection. Past this boundary, only degree levels remain in the system.
\item \textbf{The identity of the threshold.} \texttt{0.8} exists in the form \texttt{threshold\allowbreak{}Default1}, explicitly marked as overridable on site, with override actions logged. It was not discovered from the physical world; it was stipulated.
\item \textbf{The hardness of the alarm.} Step 5's alarm is not some model's low confidence; it is a promulgated rule being violated. The rule is hard, so the failure is loud.
\item \textbf{Non-activation is also on record.} Item 3 of Step 6 was explicitly checked and recorded as "not activated." The system leaves traces not only on the paths where things went wrong.
\item \textbf{The degree-level gap at the root cause.} \texttt{HZ-P1} stays at \texttt{0.5} while downstream sits at \texttt{1}: lubricant degradation has long been a "symptom," while the bearing is already "failed": this is precisely the mark of a root cause as distinct from the current fault, and it dictates that the measure must eliminate both the \texttt{1} (replace the bearing) and the persistent source of the \texttt{0.5} (change the lubrication maintenance cycle). The termination of cause-tracing is guaranteed by two structural constraints: trigger factors can only be leaves, and the parts-tree depth is bounded (four levels).
\item \textbf{The payload of transfer is only the degree level.} The pump model passes no flow, pressure, or timestamps to the plant model: only \texttt{STF-02 = 1}. The interface vocabulary is fixed in the concept layer as the system transfer factors among the 32 categories; two models built by different people at different times can thereby dock. If two channels give different degree levels, the system reports a model inconsistency rather than taking an average.
\item \textbf{Reporting a gap beats fabricating.} The measure recommendation also output the priority-2 item "start standby pump \#2," but the knowledge base has no corresponding business instance, so the system marked it as requiring manual operation per procedure. It did not fabricate a nonexistent operation; it explicitly reported the boundary of its knowledge.
\end{itemize}
The removed bearing confirmed emulsified grease and spalled raceways (staged setting), consistent with \texttt{FM-MED-01} of \texttt{HZ-P1} --- in a real deployment, such posterior confirmations of knowledge-layer assertions can be used to calibrate \texttt{priorMTTF}, but they do not modify \texttt{HZ-P1} itself.

Of the ten steps, only four cross layers (Steps 1, 2, 5, 10), and all four are the single edge type instantiation / collapse; no step spans more than two layers, and none goes from L4 directly to L1 or L2. The reduction path is the explanation path. Throughout the process, not one byte of L1 or L2 was altered; no L3 entry was modified --- only read and written with runtime degree levels; all changes happened in L4, all logged append-only.

\subsection{B.1.5 Criterion verification}
\subsubsection{B.1.5.1 C1: intentionally constituted}
\begin{table}[htbp]
\centering
\small
\begin{tabular}{@{}>{\raggedright\arraybackslash}p{0.473\textwidth}>{\raggedright\arraybackslash}p{0.473\textwidth}@{}}
\toprule
Norm & What if it were treated as a "physical discovery" \\
\midrule
\texttt{THR\_\allowbreak{}CWP\_\allowbreak{}FLOW\_\allowbreak{}LOW =\allowbreak{} 0.\allowbreak{}8} & no explanation of why different projects take different values \\
\texttt{THR\_\allowbreak{}CWP\_\allowbreak{}FLOW\_\allowbreak{}ZERO =\allowbreak{} 8 m$^3$/\allowbreak{}h} & physically there is no essential difference between 8 and 7.9 \\
\texttt{duration =\allowbreak{} 60 s}, \texttt{deadband =\allowbreak{} 2\%} & the physical process itself has no "60 seconds" boundary \\
\texttt{ECP-MECH} being one entity & no such thing can be disassembled out \\
the 32-category reduction basis & nature provides no classification \\
the four-grade state semantics & physical quantities are continuous \\
trigger factors may only be leaves & physical causation has no such prohibition \\
\texttt{exclusion\allowbreak{}Group} conclusion mutual exclusion & physical states do not exclude themselves \\
\bottomrule
\end{tabular}
\end{table}
All eight norms are promulgations (world-to-word); not one is a description. C1 holds, and it holds in a way that can be checked rather than argued.

\subsubsection{B.1.5.2 C2: the readable generative archive}
\begin{table}[htbp]
\centering
\small
\begin{tabular}{@{}>{\raggedright\arraybackslash}p{0.473\textwidth}>{\raggedright\arraybackslash}p{0.473\textwidth}@{}}
\toprule
Interrogation & Entry the answer points to \\
\midrule
by what right is the flow called abnormal? & \texttt{SEE.\allowbreak{}measurement\allowbreak{}Reduction\allowbreak{}Formula}, input source \texttt{EPT-\allowbreak{}DESIGN-\allowbreak{}FLOW} \\
by what right is the threshold 0.8? & \texttt{LTT-\allowbreak{}TR-\allowbreak{}1.\allowbreak{}threshold\allowbreak{}Default1}, overridable with a trace \\
by what right CRIT rather than DEG? & \texttt{CTT-\allowbreak{}RULE\_\allowbreak{}FLOW\_\allowbreak{}CRIT-\allowbreak{}2} (\texttt{flow < 8}) holds; \texttt{RULE\_\allowbreak{}FLOW\_\allowbreak{}DEG} requires \texttt{flow $\geq$ 8}, which fails \\
by what right is \texttt{HZ-P0} a 1 rather than a 0.5? & \texttt{SOT-\allowbreak{}FLOW\_\allowbreak{}LOSS\_\allowbreak{}CRITICAL-\allowbreak{}1.\allowbreak{}target\allowbreak{}Degree\allowbreak{}Level} \\
what does this 1 mean to a human? & \texttt{HDS-\allowbreak{}P0-\allowbreak{}L100.\allowbreak{}direct\allowbreak{}Observation\allowbreak{}State\allowbreak{}Description} \\
by what right was the mistakenly closed valve excluded? & the inducement end \texttt{TN-P2} of \texttt{CR-P7} was falsified by valve-position feedback and the patrol record \\
by what right is the bearing the root cause? & the three edges \texttt{CR-P1}$\to$\texttt{CR-P2}$\to$\texttt{CR-P4} all hit, each edge's \texttt{PSRE} degree-level mapping self-consistent \\
by what right does it affect the office area? & \texttt{RE-\allowbreak{}02.\allowbreak{}affected\allowbreak{}Object\allowbreak{}Type\allowbreak{}Codes}, \texttt{loss\allowbreak{}Expansion\allowbreak{}Model =\allowbreak{} linear amplification} \\
by what right is bearing replacement recommended? & \texttt{POB\_\allowbreak{}CWP\_\allowbreak{}001\_\allowbreak{}HAZ\_\allowbreak{}L3\_\allowbreak{}BEARING\_\allowbreak{}JAM\_\allowbreak{}002}, bound to \texttt{HZ-P2} via \texttt{ARNP} \\
\bottomrule
\end{tabular}
\end{table}
All nine interrogations land on a coded entry. C2 holds. This is not interpretable (finding after-the-fact explanations for behavior) nor merely auditable (having logs to consult) but interrogable: every conclusion has a definite, structured defendant, and disagreement can only respond by amending that one record.

\subsubsection{B.1.5.3 S2: promulgated constraint}
At 08:11 the ratio was already \texttt{0.\allowbreak{}760 < 0.\allowbreak{}8}, and the system did not alarm because \texttt{duration =\allowbreak{} 60 s} was not met; at 08:12 the condition was met and the trigger fired at once. A fitting-style system would have output "suspected anomaly, confidence 0.7" at this point. The value of the constraint lies in its refusal to speak when the conditions are not met.

\subsubsection{B.1.5.4 S3: loud failure}
All three kinds of failure appear as alarms rather than degraded output:

\begin{enumerate}
\item physical failure --- \texttt{RULE\_\allowbreak{}FLOW\_\allowbreak{}CRIT} hit $\to$ degree level set to 1 $\to$ level-1 alarm, with no vague intermediate state;
\item missing knowledge: when "start standby pump \#2" has no corresponding business instance, the gap is explicitly reported; no plausible-looking steps are generated;
\item model defect: structurally distinguished from instance violation:
\end{enumerate}
\begin{table}[htbp]
\centering
\small
\begin{tabular}{@{}>{\raggedright\arraybackslash}p{0.306\textwidth}>{\raggedright\arraybackslash}p{0.306\textwidth}>{\raggedright\arraybackslash}p{0.306\textwidth}@{}}
\toprule
Phenomenon & Verdict & Grounds \\
\midrule
reading out of bounds, rule hit & \texttt{Violation(i)} & L3 norms complete; L4 data non-compliant \\
reading finds no slot & \texttt{Coverage\allowbreak{}Gap} & no matching object class + point code in L3 \\
failure mechanism reduces to none of the 32 categories & \texttt{Coverage\allowbreak{}Gap} & no matching value in the enumeration basis \\
two channels give different degree levels & \texttt{Maintenance\allowbreak{}Fault} & internal inconsistency of the knowledge layer (A1 failing) \\
no explanation after all paths excluded & \texttt{Coverage\allowbreak{}Gap} & causal tree incomplete \\
\bottomrule
\end{tabular}
\end{table}
The dividing line falls between L3 and L4. Were the two layers merged, this distinction would fail at once --- "no explanation" could no longer be adjudicated as the world producing a new trick versus the model being insufficient. This is the most practical reason L3 and L4 must stand separate.

\subsubsection{B.1.5.5 Proposition A.16: the structural source of \texttt{d = 4}}
\begin{table}[htbp]
\centering
\small
\begin{tabular}{@{}>{\raggedright\arraybackslash}p{0.306\textwidth}>{\raggedright\arraybackslash}p{0.306\textwidth}>{\raggedright\arraybackslash}p{0.306\textwidth}@{}}
\toprule
Depth & \texttt{hazard\allowbreak{}Entity\allowbreak{}Scope} & Node in this case \\
\midrule
1 & medium unit under equipment & \texttt{HZ-P1} (\allowbreak{}lubricating oil) \\
2 & component under equipment & \texttt{HZ-P2} (pump bearing) \\
3 & entity combination under equipment & \texttt{HZ-P5} (\allowbreak{}mechanical transmission chain) \\
4 & the equipment under analysis & \texttt{HZ-P0} (chilled-water pump) \\
\bottomrule
\end{tabular}
\end{table}
The causal tree's depth is capped by the parts tree's depth, and the parts-tree levels are enumerated as four in the concept layer; together with the two termination guarantees (trigger factors only as leaves, and the finite closed reduction basis) the number of cause-tracing steps is bounded above by \texttt{d $\times$ |C|}. In this run: depth 4, edge tests 6 (3 exclusions + 3 hits). This upper bound is also why this inference can complete within the real-time alarm window.

\subsubsection{B.1.5.6 Proposition A.19: the readability gradient and extraction complexity}
\begin{table}[htbp]
\centering
\small
\begin{tabular}{@{}>{\raggedright\arraybackslash}p{0.173\textwidth}>{\raggedright\arraybackslash}p{0.173\textwidth}>{\raggedright\arraybackslash}p{0.173\textwidth}>{\raggedright\arraybackslash}p{0.173\textwidth}>{\raggedright\arraybackslash}p{0.173\textwidth}@{}}
\toprule
Layer & Scale & Change frequency & Direct human readability & Machine adjudication cost \\
\midrule
L1 & 30 items & frozen & requires ontology training & extremely low (shape matching) \\
L2 & ~107 classes + 32-value basis & only added, never altered & requires industry training & low (\allowbreak{}enumeration lookup) \\
L3 & thousands of entries & frozen clause by clause & readable by domain experts & medium (graph traversal, bounded) \\
L4 & unbounded stream & appended by the second & directly readable by operators & high (\allowbreak{}continuous quantities, need reduction) \\
\bottomrule
\end{tabular}
\end{table}
Scale, change frequency, and naive readability increase in the same direction; machine adjudication cost increases in the same direction; stability varies inversely (L1 highest, L4 lowest). The graded state descriptions exist precisely to bridge upper-layer stability and lower-layer readability: every degree level comes with a human-readable sentence --- not redundancy, but design.

\subsubsection{B.1.5.7 Counterfactual test: the engineering form of Structural Characterization 5.1}
\begin{table}[htbp]
\centering
\small
\begin{tabular}{@{}>{\raggedright\arraybackslash}p{0.306\textwidth}>{\raggedright\arraybackslash}p{0.306\textwidth}>{\raggedright\arraybackslash}p{0.306\textwidth}@{}}
\toprule
Merger & Conflicting goals & Consequence on this case \\
\midrule
L1 + L2 & stability vs openness & every newly added failure mechanism would touch the syntax layer; the semantics of connectors like \texttt{manifests\allowbreak{}As} would loosen; \texttt{HZ-\allowbreak{}P2 manifests\allowbreak{}As FM-\allowbreak{}PHY-\allowbreak{}01} would mean different things in different periods; all historical reasoning traces would fail \\
L1 + L3 & abstraction vs concreteness & \texttt{ACCCCP} and \texttt{35040} would appear in the syntax vocabulary; switching industries would mean switching grammar; cross-domain reuse drops to zero \\
L1 + L4 & stable $\wedge$ abstract vs open $\wedge$ concrete & the syntax would change with every reading --- untenable \\
L2 + L3 & openness vs stability & augmenting with new knowledge and modifying existing concepts become indistinguishable; the \texttt{Model\allowbreak{}Defect}/\texttt{Violation} distinction vanishes; the 32-category basis gets casually expanded along with knowledge augmentation, and Proposition A.16's upper bound fails \\
L2 + L4 & abstraction vs concreteness & \texttt{0.8} gets learned away by field data; promulgation degenerates into fitting; S2 vanishes \\
L3 + L4 & stability vs openness & the write at 08:14 setting the degree level to 1 would rewrite \texttt{HZ-P0} itself; \texttt{prior\allowbreak{}MTTF =\allowbreak{} 35040} would be contaminated by one pump's single failure; the 12:05 return to zero could not restore it; the audit trail is lost; C2 fails \\
\bottomrule
\end{tabular}
\end{table}
All six pairs conflict, without exception; hence at least four independent content regions are needed. These four layers were not designed to this table; they were forced out by these six costs over engineering iteration --- the separation of knowledge-base files from operating data was itself a forced boundary.

\subsection{B.1.6 The proof boundary of this case}
\begin{itemize}
\item Structural Characterization 5.1 gives a lower bound; this case provides only one instance attaining the bound, and does not prove the four layers are the unique layering scheme.
\item It does not prove S1 (zero-shot operability): the \texttt{ACCCCP} of this case lies within the knowledge base's coverage. The gap in the measure recommendation ("start standby pump \#2") is positive evidence for S3 and simultaneously a marker that S1 awaits verification.
\item It does not prove the 32-category reduction basis complete for all industrial scenarios; the case proves only that there exists a finite closed basis on which reduction terminates.
\item The fault event, handling process, and time-series readings of L4 are demonstration runs, with static parameters injected via twin attributes. What is verified is the relational discipline among the four layers, not some real fault of some real pump; the content and structure of L1--L3 are really existing artifacts, checkable entry by entry.
\end{itemize}
\section{Appendix B.2. The Curiosity Drill-Feed Anomaly (Sol 1536): A Cold-Start Replay of a Zero-Posterior-Data World}
\begin{quote}
This case and B.1 are dual to each other: B.1 shows the four-layer framework operating in a ground-operations world where data can accumulate; this case shows the same framework operating in the world of \emph{the ultimate cold start} --- the failure unprecedented, no operational data to learn from, the asset unapproachable and unrepairable. The entire content of the prior framework can only come from the design archive.  \textbf{Layering statement (prior to all content).} This appendix is divided into two layers, with the boundary marked section by section: - \textbf{The historical-record layer}: the course of the event, the telemetry behavior, the hypothesis set and the order of exclusions, the timeline: all taken from public literature, with sources marked item by item (the source table at the end of B.2.1); - \textbf{The reconstruction layer}: the concrete entries of the four-layer carriers (\allowbreak{}vocabulary mappings, the failure evolution model, threshold rules, the graded-state semantics) are a \emph{counterfactual reconstruction} of NASA's design archive according to this paper's framework. NASA's actual onboard fault-protection logic is a promulgated rule set (which is itself evidence for C1/C2) but it contains no cause-tracing model; the "in-framework replay" of B.2.3 is a deduction, not an onboard record of NASA. All values absent from public sources are marked "demonstration values."  \textbf{Cross-domain convention (relation to B.1).} For all L1/L2 content identical in semantics to B.1, this appendix \emph{carries over verbatim} B.1's parts of speech, codes, and enumerations, marked "carried over from B.1" at first occurrence. This is the case-level cashing-out of the cross-domain invariance promised in the preamble of Appendix B: the concept layer verbatim unchanged, cross-industry only adding blocks, never changing shape; see structural corollary I of Chapter 7: from a ground cooling plant to a Mars rover, the shared part of the syntax layer and the concept layer has zero change.
\end{quote}
\subsection{B.2.1 The historical-record layer: the event and the investigation on record}
\textbf{The event.} On Sol 1536 (November 30, 2016; the Sol-to-Earth-calendar conversion may differ by one day depending on convention, and this paper does not rest on it), Curiosity was performing its 16th rock drilling at the Precipice target (Murray formation, lower slope of Mount Sharp) (Kinnett et al., 2022). The robotic arm had preloaded the drill against the rock face; the drill-feed motor stalled at the start of its long travel after the brake (an electromagnetic friction-disk type, engaged when unpowered) was released: the encoder showed zero rotation of the motor shaft; the closed-loop rate controller drove the current up to its limit with still no indication of shaft motion, triggering the stall determination. The flight software \emph{autonomously terminated} the drilling sequence and dumped the high-rate (512 Hz) motor-control history buffer into a high-priority data product; the robotic arm unloaded the stabilizer and lifted the drill clear. Drilling was suspended thereafter (Kinnett et al., 2022).

\textbf{Mechanism essentials.} The drill-feed motor is a brushless DC motor equipped with a magnetic encoder and an electromagnetic brake: when the brake is unpowered, a spring presses the moving brake disk against the fixed disk, locking the feed; when powered, the electromagnetic coil pulls the moving disk open and the motor advances the drill stem through the transmission. The brake has \emph{dual redundant coils}; and \emph{no sensor} inside the brake can directly report its released/engaged state (Kinnett et al., 2022; Lakdawalla, 2017).

\textbf{The diagnostic chain (the exclusion tree the engineers actually walked).}

\begin{longtable}{@{}>{\raggedright\arraybackslash}p{0.223\textwidth}>{\raggedright\arraybackslash}p{0.223\textwidth}>{\raggedright\arraybackslash}p{0.223\textwidth}>{\raggedright\arraybackslash}p{0.223\textwidth}@{}}
\toprule
Stage & Time & Content & Source \\
\midrule
\endfirsthead
\toprule
Stage & Time & Content & Source \\
\midrule
\endhead
stall and autonomous termination & Sol 1536 & encoder zero change, current at limit, stall determination, sequence termination & (Kinnett et al., 2022) \\
winding/harness exclusion & Sol 1536 telemetry & routine continuity checks plus voltage/current telemetry normal; motor windings and harness falsified & (Kinnett et al., 2022) \\
initial hypothesis set & 2016-12 & encoder electromechanical failure / motor assembly mechanical blockage / gearbox failure / brake release failure / control-system failure & (Kinnett et al., 2022); the "encoder/brake" two-candidate version in early public reports, see (NASA/JPL, 2016) \\
convergence on the brake & 2016-12 to 2017-03 & coils energized separately, current raised, both coils energized together, repeated attempts --- behavior consistent $\to$ individual coil differences excluded, pointing to interference with the moving brake disk; working hypothesis: a displaced part or foreign object debris (FOD) & (Kinnett et al., 2022; Clark, 2016; Lakdawalla, 2017) \\
blockage signature & first half of 2017 & diagnostics showed an intermittent high-resistance state recurring "once per revolution," worsening with use & (Kinnett et al., 2022; Vasavada, 2022) \\
commonality across same-type motors & Sol 1627 (2017-03) & the drill chuck motor (same model) showed a similar failure signature $\to$ a system-level common factor surfaced; the turret CHIMRA vibration mechanism was identified as the common stressor and its use was thereafter restricted & (Kinnett et al., 2022; Vasavada, 2022) \\
accelerated degradation & after Sol 1651 (2017-03-29) & after the Ogunquit Beach sand-sample sieving (about 20 minutes of turret vibration), feed performance deteriorated markedly & (Kinnett et al., 2022; Lakdawalla, 2017) \\
judged undependable & 2017-06 & mechanism reliability kept degrading; the feed mechanism was judged no longer able to support drilling & (Kinnett et al., 2022) \\
permanent deployment & Sols 1754--1780 (2017-07, across conjunction) & open-loop commutation, thermal watchdog, brake cycling, drill pointing to zenith to borrow gravity; the feed pushed to its full deployment of 110 mm over 4 sols (before conjunction) plus 1 sol (after) & (Kinnett et al., 2022) \\
workaround recovery & Sol 2057 (2018-05) & first successful "feed-extended drilling" (FED) sampling: the robotic arm substituted for the feed mechanism in applying force; FEST delivered samples directly by reversing the drill stem, bypassing CHIMRA. 521 sols after the initial stall & (Kinnett et al., 2022; Fraeman et al., 2020) \\
\bottomrule
\end{longtable}
\textbf{Sources.} (Kinnett et al., 2022) 46th Aerospace Mechanisms Symposium (2022), the JPL anomaly investigation retrospective (NTRS 20220006415); (Fraeman et al., 2020), the Vera Rubin Ridge synthesis (JGR Planets); (Clark, 2016) Spaceflight Now (2016-12-29, the FOD hypothesis); (Vasavada, 2022) Springer mission review (2022, s11214-022-00882-7, the "once per revolution" blockage and the CHIMRA commonality); (\allowbreak{}Lakdawalla, 2017) Planetary Society (2017-09-06, brake structure and the dual-coil exclusion details); (NASA/JPL, 2016) NASA/JPL press release (2016-12-05, the encoder/brake two candidates).

\begin{quote}
\textbf{Numerical convention}: in this case, apart from the historical values recorded in the source table above (Sol numbers, sampling frequencies, feed travel, etc.), all thresholds, degree-level values, and confidences are demonstration values, used to display the form of the inference rather than as measured data.
\end{quote}
\subsection{B.2.2 The reconstruction layer: the four-layer carriers (\allowbreak{}counterfactual reconstruction)}
Reconstruction principle: use only structures that actually exist in NASA's public design materials --- mechanisms, redundancies, telemetry points, fault-protection logic; the framework form is schematized per Chapter 5 of this paper. \emph{Whatever is semantically identical to B.1 is carried over verbatim, marked place by place.}

\subsubsection{B.2.2.1 The L1 syntax layer (stable $\times$ abstract)}
\textbf{The vocabulary table (13 items) and the connector table (17 items) of B.1 carried over verbatim} (B.1.1), without a single change. The mappings used in this case: \texttt{Physical\allowbreak{}Object} (the drill, the feed motor assembly, the brake disk, the robotic arm), \texttt{Failure\allowbreak{}Mode}, \texttt{Trigger\allowbreak{}Factor}, \texttt{Failure\allowbreak{}Reasoning\allowbreak{}Node}, \texttt{System\allowbreak{}Reasoning\allowbreak{}Element}, \texttt{RiskEvent}, \texttt{Management\allowbreak{}Standard\allowbreak{}Knowledge} (\allowbreak{}thresholds and criteria), \texttt{Computation\allowbreak{}Input\allowbreak{}Knowledge} (rated parameters); the connectors used are \texttt{has\allowbreak{}Component}, \texttt{manifests\allowbreak{}As}, \texttt{has\allowbreak{}Relation\allowbreak{}From/\allowbreak{}To}, \texttt{realizes}, \texttt{applied\allowbreak{}To\allowbreak{}Entity\allowbreak{}Type}, \texttt{references\allowbreak{}Threshold\allowbreak{}Config}, \texttt{propagation\allowbreak{}Applicable\allowbreak{}Conditions}.

\emph{The zero change to the syntax layer is itself a constructive demonstration of the cross-domain invariance claim: from building cooling to a deep-space drill, the language has not changed generations.} (A demonstration, not independent evidence: the vocabulary reuse in this case was constructed under a "verbatim carry-over" discipline; independent evidence awaits comparison after the two industries are each independently modeled.)

\subsubsection{B.2.2.2 The L2 concept layer (open $\times$ abstract)}
\textbf{Parts carried over from B.1} (marked as such):

\begin{itemize}
\item \textbf{The reduction basis (32 categories)}: carried over verbatim (B.1.2). Landing points in this case: brake mechanical blockage $\to$ \emph{function interruption failure} (carried over); sample-delivery interruption $\to$ transfer type \emph{matter} (carried over).
\item \textbf{The inducement-reduction discipline}: carried over verbatim (end of B.1.2): trigger factors (\allowbreak{}inducement nodes with the \texttt{TN} prefix, code sequence carried over from B.1) can only be leaves of the causal tree and \emph{are not classified into the failure-mode basis (physical 10 / medium 9)} --- environmental inducements go to the \emph{nine environmental-factor categories within the 32-item basis} (FOD $\to$ \emph{foreign-object intrusion}; CHIMRA vibration $\to$ \emph{mechanical environment}); personnel/rule inducements go to the TriggerFactor concept enumeration (\emph{outside} the 32-item basis, promulgated as the dedicated inducement vocabulary).
\item \textbf{Graded state semantics}: \texttt{0} normal, \texttt{0.25} adverse trend, \texttt{0.5} symptom, \texttt{1} failed (carried over from B.1.2). Continuous telemetry quantities live only in L4.
\item \textbf{Graph-theoretic constraint}: trigger factors can only be leaves (carried over from B.1.2).
\end{itemize}
\textbf{Newly added parts (only adding blocks)}: one batch of spacecraft object classes: \texttt{MSL\_DRILL} (rotary-percussive drill), \texttt{DRILL\_\allowbreak{}FEED\_\allowbreak{}ASSY} (feed mechanism: motor + encoder + brake + transmission), \texttt{DRILL\_\allowbreak{}BRAKE} (the brake, with dual redundant coils), \texttt{CHIMRA} (sample processing and sieving mechanism), \texttt{ROBOTIC\_\allowbreak{}ARM} (five-degree-of-freedom robotic arm). The concepts equipment type, component type, entity-combination type, and so on are all carried over from B.1, with the new object classes hung beneath them --- the parts of speech untouched. The FED/FEST operating-mode concepts were added after the fact through the port of entry (B.2.3, Step 10), again "only added, never altered."

\subsubsection{B.2.2.3 The L3 knowledge layer (stable $\times$ concrete): the reconstructed knowledge of this case}
\textbf{The failure evolution model (\allowbreak{}reconstructed from the hypothesis set of NASA's retrospective paper, each edge annotated with its historical correspondent):}

\begin{longtable}{@{}>{\raggedright\arraybackslash}p{0.116\textwidth}>{\raggedright\arraybackslash}p{0.116\textwidth}>{\raggedright\arraybackslash}p{0.116\textwidth}>{\raggedright\arraybackslash}p{0.116\textwidth}>{\raggedright\arraybackslash}p{0.116\textwidth}>{\raggedright\arraybackslash}p{0.116\textwidth}>{\raggedright\arraybackslash}p{0.116\textwidth}@{}}
\toprule
Depth (scope level) & Scope & Node & Entity & Failure mode & Reduction category & Historical correspondent \\
\midrule
\endfirsthead
\toprule
Depth (scope level) & Scope & Node & Entity & Failure mode & Reduction category & Historical correspondent \\
\midrule
\endhead
leaf & exogenous inducement & \texttt{TN-FOD} & --- & foreign object debris intruding into the brake gap & foreign-object intrusion (\allowbreak{}environ\allowbreak{}mental-factor basis, carried over) & the FOD working hypothesis (Kinnett et al., 2022; Clark, 2016) \\
leaf & exogenous inducement & \texttt{TN-VIB} & --- & accumulated CHIMRA vibration stress & mechanical environment (\allowbreak{}environ\allowbreak{}mental-factor basis, carried over) & the same-type-motor commonality, the accelerated degradation after Sol 1651 (Kinnett et al., 2022; Vasavada, 2022) \\
2 & component under equipment & \texttt{HZ-BRK} & \texttt{DRILL\_\allowbreak{}BRAKE} moving brake disk & brake not fully released & function interruption failure (carried over) & \textbf{hit}: the converged conclusion (Kinnett et al., 2022; Lakdawalla, 2017) \\
2 & component under equipment & \texttt{HZ-ENC} & magnetic encoder & encoder electromechanical failure & function interruption failure (carried over) & excluded: encoder field counts varied normally during brake cycling (Kinnett et al., 2022; Lakdawalla, 2017) \\
2 & component under equipment & \texttt{HZ-MOT} & feed motor assembly & motor assembly mechanical blockage & function interruption failure (carried over) & excluded: the "once per revolution" intermittent high-resistance signature localizes to the brake-disk interface rather than the motor interior (Kinnett et al., 2022; Vasavada, 2022) \\
2 & component under equipment & \texttt{HZ-GBX} & transmission & gearbox failure & function interruption failure (carried over) & excluded: public sources record the exclusion conclusion but not the specific grounds (Kinnett et al., 2022) \\
2 & component under equipment & \texttt{HZ-CTL} & motor control chain & control-system failure & function interruption failure (carried over) & excluded: the command chain works normally elsewhere (Kinnett et al., 2022) \\
3 & entity combination under equipment & \texttt{HZ-FEED} & \texttt{DRILL\_\allowbreak{}FEED\_\allowbreak{}ASSY} & feed motor stall (current present, no shaft motion) & function interruption failure (carried over) & Sol 1536 telemetry (Kinnett et al., 2022) \\
4 & the equipment under analysis & \texttt{HZ-DRILL} & \texttt{MSL\_DRILL} & drill feed function lost & function interruption failure (carried over) & drilling suspended (Kinnett et al., 2022) \\
\bottomrule
\end{longtable}
\textbf{The node set of this tree corresponds one-to-one with NASA's hypothesis set}: all five candidates are in the table, the hit chain is \texttt{TN-\allowbreak{}FOD /\allowbreak{} TN-\allowbreak{}VIB $\to$ HZ-\allowbreak{}BRK $\to$ HZ-\allowbreak{}FEED $\to$ HZ-\allowbreak{}DRILL}, and the four excluded sibling nodes stand alongside at level 2. The reconstruction introduces no hypothesis NASA did not consider, and deletes none.

\textbf{System propagation does not occupy causal-tree depth.} \texttt{SS-SAMPLE} (the sampling system: loss of rock-sample acquisition capability, reduced to transfer type = matter) is the system spillover of the failure consequence, anchored via \texttt{realizes} to \texttt{RE-MISSION} --- a channel separate from the causal tree (treated the same as the system-propagation paragraph of B.1.3.5). The causal tree's depth is 3: the level numbers are a fixed enumeration promulgated by the concept layer (scope level 1, medium units, is not instantiated in this mechanism; skipped levels are not renumbered), and the tree runs from level 2 (\allowbreak{}components) to level 4 (the equipment). This incidentally demonstrates Hypothesis H4's d = 4 stated as an \emph{upper bound}: the actual depth is capped by the object's parts tree, and shallower parts trees mean shallower chains (H4 carries a corresponding supplementary note; see Appendix A).

\textbf{The monitoring-and-judgment chain (\allowbreak{}reconstructed from the onboard fault-protection logic)}: information-point slots \texttt{DIP-\allowbreak{}CURRENT /\allowbreak{} DIP-\allowbreak{}ENCODER-\allowbreak{}RATE /\allowbreak{} DIP-\allowbreak{}BRAKE-\allowbreak{}CURRENT}; threshold configuration \texttt{LTT-STALL} (encoder-rate threshold, current limit, and duration window; corresponding to the actual form of NASA's stall determination (Kinnett et al., 2022), with specific values as demonstration values); the rule \texttt{JRI-STALL} conjoins two independently addressable conditions (\texttt{CTT-\allowbreak{}STALL-\allowbreak{}1}: encoder rate below threshold and sustained; \texttt{CTT-\allowbreak{}STALL-\allowbreak{}2}: current at limit) $\to$ trigger; conclusion \texttt{JC-\allowbreak{}FEED\_\allowbreak{}STALL}; write-back \texttt{SOT} $\to$ \texttt{HZ-\allowbreak{}FEED.\allowbreak{}degree\allowbreak{}Level =\allowbreak{} 1}; routing \texttt{ARNP} $\to$ failure evolution model. The entry types (\texttt{DIP/\allowbreak{}SEE/\allowbreak{}LTT/\allowbreak{}JRI/\allowbreak{}JC/\allowbreak{}SOT/\allowbreak{}ARNP}) and the division of labor ("LTT holds only thresholds; condition conjunction lives in JRI") are \emph{carried over from the assembly pattern of B.1.3.4}.

\textbf{Risk events}: \texttt{RE-MISSION} (mission level): \texttt{HZ-DRILL} is anchored via \texttt{realizes} to the loss of "rock-sample acquisition capability" --- the loss structure is \emph{mission-objective impairment}, standing alongside B.1's equipment-binary and system-linear loss structures, and illustrating a third implantation form of the \texttt{RiskEvent} part of speech (the part of speech carried over; the loss model promulgated per industry).

\subsubsection{B.2.2.4 The L4 instance layer (open $\times$ concrete)}
L4's content: the physical individual (Curiosity, serially unique); the Sol 1536 telemetry stream (encoder, current, brake current); injected design parameters (rated speed, current limit, feed travel 110 mm --- source values used where a public source exists, demonstration values where one is lacking). The discipline is carried over from B.1.4: append-only, never overwriting.

\subsection{B.2.3 In-framework replay: had this framework been onboard on Sol 1536}
An honest statement of the replay's premise: NASA's onboard fault protection \emph{actually performed} Steps 1--3 of the table below --- the stall determination and sequence termination are real onboard behavior (Kinnett et al., 2022); that part is not counterfactual. The counterfactual begins at Step 4: the degree-level semantics and the cause-tracing model were not onboard, and the cause-tracing was completed by the ground team over four months.

\begin{longtable}{@{}>{\raggedright\arraybackslash}p{0.173\textwidth}>{\raggedright\arraybackslash}p{0.173\textwidth}>{\raggedright\arraybackslash}p{0.173\textwidth}>{\raggedright\arraybackslash}p{0.173\textwidth}>{\raggedright\arraybackslash}p{0.173\textwidth}@{}}
\toprule
Step & Time & Action & Inter-layer edge & History / reconstruction \\
\midrule
\endfirsthead
\toprule
Step & Time & Action & Inter-layer edge & History / reconstruction \\
\midrule
\endhead
1 & Sol 1536 & telemetry lands & L4 $\to$ L3 instantiation & history: encoder zero change, current at limit (Kinnett et al., 2022) \\
2 & same & stall determination triggers & within L3 & history: the onboard counterpart of \texttt{LTT-STALL} really triggered (Kinnett et al., 2022) \\
3 & same & sequence termination + high-rate buffer dump & L3 $\to$ L4 & history: the flight software terminated autonomously (Kinnett et al., 2022) \\
4 & same & degree-level write-back \texttt{HZ-\allowbreak{}FEED =\allowbreak{} 1}, alarm & L3 $\to$ L4 collapse & reconstruction (onboard there were only fault codes, no degree-level semantics) \\
5 & same & route the cause-tracing model, enumerate the candidate set & within L3 & reconstruction: the five candidates enumerated immediately --- \emph{on the ground, assembling the same set actually took weeks} (Kinnett et al., 2022) \\
6 & the ensuing diagnostic period & exclusions: windings/harness (\allowbreak{}continuity checks), encoder (encoder field counts varied normally during brake cycling), control system (command chain normal elsewhere) & within L3 & the excluded items and the grounds of exclusion are both taken from the historical record (Kinnett et al., 2022; Lakdawalla, 2017); the framework only makes the order and the adjudications explicit \\
7 & same & convergence: \texttt{HZ-BRK} (brake not released), working hypothesis \texttt{TN-FOD} & within L3 & the historical conclusion (Kinnett et al., 2022; Clark, 2016; Lakdawalla, 2017) \\
8 & Sol 1627 & the chuck motor's similar signature $\to$ \texttt{TN-VIB} promoted to a system-level common factor; \texttt{propagation\allowbreak{}Applicable\allowbreak{}Conditions} hits the same model of motor & within L3 (cross-model) & history (Kinnett et al., 2022; Vasavada, 2022); the framework form: one environmental factor hung on two equipment models at once through the gating condition \\
9 & after Sol 1651 & after the vibration activity, \texttt{HZ-BRK}'s degree-level deterioration trend from 0.5 toward 1 is explicitly recorded & L4 $\to$ L3 & the historical degradation (Kinnett et al., 2022); the degree-level trajectory (\allowbreak{}demonstration values) \\
10 & 2017-06 & judged undependable $\to$ concept augmentation enters: \emph{FED/FEST as new operating-mode entries} enter the concept layer through the port of entry of Section 5.2, accompanied by new procedure clauses into the knowledge layer & port of entry & history: FED/FEST were developed on the ground (Kinnett et al., 2022; Fraeman et al., 2020); the framework form: augmentation, not modification: zero changes to existing entries \\
\bottomrule
\end{longtable}
\textbf{Seven annotations on the replay} (\allowbreak{}corresponding to the seven points of B.1.4; only the increments are written):

\begin{itemize}
\item \textbf{The first real trial of S1}: zero failure precedents, zero operational data, adjudicative capacity drawn entirely from the design archive. The stall was classified as a violation of a known type at the arrival of the \emph{very first instance} --- this is not a demonstration: NASA's onboard fault protection really did Steps 1--3, and is itself existence evidence of "an adjudicator extracted from promulgated norms" (what it lacked is cause-tracing and degree-level semantics: everything from Step 4 on).
\item \textbf{The deep-space version of loud failure}: autonomous termination + data dump = locatable, reportable, checkable against clauses. A fitting-style system has no counterpart here: there is no training distribution to speak of.
\item \textbf{Explicit registration of the unobservable point}: the brake has no built-in sensor, and the framework marks \texttt{HZ-BRK} as "norm promulgatable, state not directly readable": adjudication can only be completed indirectly through the dual-coil exclusion experiments. Such nodes carry a "not directly readable" flag in the knowledge layer (a demonstratively reconstructed entry); it predicts where diagnosis will be most expensive, matching reality: the brake was exactly the core of the four-month investigation.
\item \textbf{The root-cause degree-level gap}: \texttt{TN-FOD} remains an inducement (a leaf), \texttt{HZ-BRK} is the current fault, \texttt{TN-VIB} is the accelerating factor: the role differences are guaranteed by the graph-theoretic constraint (\allowbreak{}inducements can only be leaves), requiring no on-the-spot judgment.
\item \textbf{Cross-model docking}: the common-factor identification of Step 8 was completed between two independently built models (feed, chuck), and the interface vocabulary is the 32-category basis carried over from B.1: isomorphic to B.1's "the payload of transfer is only the degree level."
\item \textbf{The deep-space version of the port of entry}: FED/FEST are a textbook case of concept augmentation: new operating modes, new procedures, new acceptance criteria: all entering by augmentation, with zero modification of existing entries.
\item \textbf{Reporting a gap beats fabricating}: the framework can give no measure entry for "repairing the brake on orbit": the knowledge base has none, so it reports the gap (as actual history also did: no repair option, only a workaround).
\end{itemize}
\subsection{B.2.4 Comparison: with the framework versus the actual process}
\begin{table}[htbp]
\centering
\small
\begin{tabular}{@{}>{\raggedright\arraybackslash}p{0.306\textwidth}>{\raggedright\arraybackslash}p{0.306\textwidth}>{\raggedright\arraybackslash}p{0.306\textwidth}@{}}
\toprule
Stage & The actual process & The in-framework replay \\
\midrule
stall determination & onboard, immediate (\allowbreak{}promulgated rules already on duty) & the same (Steps 1--3 are history) \\
hypothesis-set assembly & a ground expert team, several weeks & the promulgated model enumerates immediately (Step 5) \\
order of exclusions and grounds of adjudication & improvised on the spot, written up after the fact (Kinnett et al., 2022) & the tree structure promulgated in advance; exclusion experiments checked against clauses \\
common-factor identification & chance discovery on Sol 1627 & systematic hit of the gating condition (Step 8) \\
total diagnostic period & about 4 months to convergence, 521 sols to workaround recovery & enumeration-tree traversal + scheduling of the pre-set exclusion experiments \\
absorption of new operating modes & engineering rework (FED/FEST) & concept augmentation (port of entry); the framework unmoved \\
\bottomrule
\end{tabular}
\end{table}
\textbf{The honest version of the closing sentence}: this comparison does not claim the framework could compress 521 sols into a few days --- the exclusion experiments must be performed on Mars, the conjunction must be waited out; these physical constraints are unaffected by the framework. What the framework compresses is the \emph{cognitive organization cost}: the hypothesis set goes from "assembled by experts over weeks" to "enumerated upon promulgation"; the grounds of exclusion go from "written up after the fact" to "checked clause by clause"; the common factor goes from "chance discovery" to "gated hit."

\subsection{B.2.5 Criterion verification}
\textbf{C1 (\allowbreak{}intentionally constituted)}: every stipulation of the drill-feed mechanism (the stall threshold, the dual redundant coils, the fault-protection logic, the feed travel) is promulgated rather than discovered; "zero encoder change under current limit means stall" is not a physical law but an adjudication clause written into the flight software. The way it holds is of the same form as B.1.5.1, and purer: this object does not even have the "broken --- buy a new one" option; the norm is its only form.

\textbf{C2 (readable generative archive)}: this reconstruction takes as its archive only public design materials and retrospective literature (the source table of B.2.1); the real engineering archive (drawings, FMECA, test specifications) sits under JPL's institutional maintenance, and this appendix reaches only the public part. This is itself a field measurement of C2's boundary: the reconstruction extends exactly as far as the archive is public.

\textbf{S2 (\allowbreak{}promulgated constraint)}: the constraint refuses to speak when its conditions are not met --- across the 15 successful drillings before Sol 1536 recorded in public sources, there is no record of a false stall trigger (Kinnett et al., 2022) (the constraint does not overstep), while one real deviation on Sol 1536 triggered it at once (the constraint is not absent).

\textbf{S3 (loud failure)}: autonomous termination + high-rate buffer dump = the deep-space version of alarming failure; there is no silent degradation onboard.

\textbf{S1 (zero-shot operability)}: the main object under test in this case (the narrative finale; the criterion order follows B.1.5). At the arrival of the first-ever instance of the failure, the adjudication already held (Steps 1--3 are history), and the prior's provenance is unique: the design archive. This is the first trial of the part B.1 explicitly left unproven ("S1 is not proved").

\textbf{Cross-domain invariance (the cashing-out of the promise in Appendix B's preamble)}: L1 zero change; in L2, the reduction basis, the graded-state semantics, and the graph-theoretic constraint are carried over verbatim, and the additions are only one batch of object classes and one batch of operating-mode concepts (both augmentations, with zero modification of existing entries); L3 is entirely newly built (the knowledge layer is promulgated industry by industry to begin with); L4 has no prior content. From the cooling plant to the Mars rover, the common part of the concept layer has zero diff.

\subsection{B.2.6 The proof boundary of this case}
\begin{itemize}
\item \textbf{The status of the reconstruction layer}: the four-layer entries are a counterfactual reconstruction; NASA did not build them this way. What is proved is "the framework can losslessly accommodate the entire historical structure of this real event," not "NASA used this framework."
\item \textbf{The limit of the S1 proof}: the onboard fault protection (Steps 1--3) is real existence evidence of promulgated rules; the cause-tracing model portion (from Step 4 on) is deduction. What S1 thereby gains is the existence of "an adjudicator that can run on zero data"; the compression benefit of cause-tracing relies on the next assumption.
\item \textbf{The equal-coverage assumption}: the comparison of B.2.4 holds if and only if the promulgated model's hypothesis set covers the same ground as the expert team's actual hypothesis set. In this case the two coincide (the one-to-one correspondence was registered in B.2.2.3, all five candidates in the table), but this is an after-the-fact check; it does not prove that the hypothesis set of any future event is already covered by promulgation --- beyond coverage, the verdict is A3's "unknown," referred to the concept-augmentation passage.
\item \textbf{Values}: the convention dividing demonstration values from historical values is at the end of B.2.1.
\item \textbf{Division of labor with B.1}: B.1 proves the four-layer structure operating in a really existing artifact (L1--L3 real); B.2 proves the same structure's coverage of a zero-posterior-data world (S1 on trial). The evidence of the two cases does not substitute for each other.
\end{itemize}
\section{Appendix C. Four-Layer Reverse-Reading Dossiers of External Witnesses}
\begin{quote}
\textbf{Conventions}: none of these systems' self-descriptions contains the phrase "four layers"; the mappings below are reverse readings through the lens of this framework --- an inventory in the four layers syntax / concepts / knowledge / instances, on which each system's content happens to fall into place, with the layer-by-layer alignments given in the tables. The "promulgator" column answers the "who promulgates" of the Chapter 4 criterion. Each section is self-contained and can be read independently of the main text.
\end{quote}
\subsection{C.1 BACnet (ANSI/ASHRAE Standard 135; ISO 16484-5)}
\textbf{Background}: the most widely installed open protocol in building automation. ASHRAE launched the project in 1987 (SPC 135, converted to the standing committee SSPC-135 for continuous maintenance from 1996); the first edition of Standard 135 appeared in 1995; it was adopted as ISO 16484-5 in 2003. Its design starting point was interoperability: "to let equipment from different vendors be uniformly monitored and commanded at one workstation."

\textbf{The four-layer reverse reading}:

\begin{table}[htbp]
\centering
\small
\begin{tabular}{@{}>{\raggedright\arraybackslash}p{0.306\textwidth}>{\raggedright\arraybackslash}p{0.306\textwidth}>{\raggedright\arraybackslash}p{0.306\textwidth}@{}}
\toprule
Layer & Counterpart & Promulgator \\
\midrule
syntax & the application data-type system (Boolean/\allowbreak{}Unsigned/\allowbreak{}Real/\allowbreak{}Enumerated/\allowbreak{}Date/\allowbreak{}Time/\allowbreak{}ObjectIdentifier\dots{}) plus the "object--property--service" metamodel: everything in the world presents as an object, objects are described by properties, actions are initiated through services & ASHRAE SSPC-135 \\
concepts & the \textbf{closed set of standard object types}: 18 types in 1995 $\to$ 54 in 2012 $\to$ 61 in ISO 16484-5:2017 (Analog Input, Schedule, Trend Log, Event Enrollment, \dots{}); \textbf{enumerated property identifiers}, over 160; closed enumerations of engineering units, reliability, event states, etc. & ASHRAE SSPC-135 \\
knowledge & the five classes of service primitives (object access / device management / alarm \& event / file transfer / virtual terminal); BIBB interoperability capability blocks + PICS conformance statements --- a testable specification of "how a device shall respond" & ASHRAE SSPC-135 \\
instances & object instances in concrete devices (addressed by type + 22-bit instance number) + Present\_Value real-time values, directly grounded to sensors & deployment sites \\
\bottomrule
\end{tabular}
\end{table}
\textbf{Three points for this paper}:

\begin{enumerate}
\item \textbf{A promulgated closed set can be augmented without being rewritten}: the object types more than tripled over thirty years, yet the semantics of old types are never revoked --- a field demonstration of "stable $\neq$ frozen," precisely the institutional form of this framework's openness by promulgation at the concept layer.
\item \textbf{The promulgation boundary is machine-checkable}: the standard requires proprietary object types to use the numbering range $\geq$128 and proprietary properties $\geq$512, and requires that existing standard object types be used wherever possible ("standard object types shall be used when possible," 135-2016 Clause 23.4). "inside/outside promulgation" made into a numbering guardrail.
\item \textbf{Conformance is testable}: the BIBB/PICS system makes "conforming to the promulgated specification" a testable, endorsable engineering fact, rather than a vendor's own claim.
\end{enumerate}
\subsection{C.2 LonWorks (LonMark International; EN 14908-6 / ANSI/CTA 709.6)}
\textbf{Background}: a control-network system initiated by Echelon, whose semantic layer is promulgated by the industry association LonMark International; its application elements were later taken into EN 14908-6.

\textbf{The four-layer reverse reading}:

\begin{table}[htbp]
\centering
\small
\begin{tabular}{@{}>{\raggedright\arraybackslash}p{0.306\textwidth}>{\raggedright\arraybackslash}p{0.306\textwidth}>{\raggedright\arraybackslash}p{0.306\textwidth}@{}}
\toprule
Layer & Counterpart & Promulgator \\
\midrule
syntax & the network-variable (NV) type system + the XIF device interface file format: a device self-declares in XIF "what I can say" (each NV's name, direction, type) & Echelon/LonMark \\
concepts & the \textbf{SNVT master list} (resource files under continuous versioned expansion; the verified starting point is v7 = 114 types): each type promulgates dimension, resolution, legal range, the \textbf{invalid value}, scale factors, down to field-level bit definitions; SCPT configuration-parameter types; ENUM enumeration types & LonMark International \\
knowledge & \textbf{Functional Profiles}: each Profile promulgates the minimal NV set a functional node of a given kind shall expose, divided into mandatory and optional items (\allowbreak{}thermostats, pump controllers, valve positioners, \dots{}) & LonMark International \\
instances & bound NV instances in a commissioned network & deployment sites \\
\bottomrule
\end{tabular}
\end{table}
\textbf{Two points for this paper}:

\begin{enumerate}
\item \textbf{Promulgation granularity down to "which value is illegal"}: every SNVT carries its own invalid-value declaration --- the promulgated form of a normative boundary.
\item \textbf{A pre-industrial form of violation interception}: a binding of type-incompatible NVs is \emph{directly rejected} in the integration tool: the violation is structurally stopped at deployment time, rather than probabilistically let through at runtime.
\end{enumerate}
\subsection{C.3 OPC: a within-group control of two product generations (OPC DA 1996 $\to$ OPC UA 2008 / the IEC 62541 series)}
\textbf{Background}: two generations of systems from the same organization (the OPC Foundation) with the same goal (\allowbreak{}industrial interoperability) --- a natural control experiment.

\begin{table}[htbp]
\centering
\small
\begin{tabular}{@{}>{\raggedright\arraybackslash}p{0.306\textwidth}>{\raggedright\arraybackslash}p{0.306\textwidth}>{\raggedright\arraybackslash}p{0.306\textwidth}@{}}
\toprule
Layer & OPC DA (1996) & OPC UA (1.0 released 2008; adopted into the IEC 62541 series in batches from 2010) \\
\midrule
syntax & COM/DCOM interface shell (Windows only) & the address-space metamodel: nodes / references / type hierarchies / namespaces (IEC 62541-3) \\
concepts & $\times$ flat tag list, no type system & \textbf{Companion Specifications}: industry organizations promulgating closed sets of object types for their industries --- OPC 40001 Machinery, OPC 40501 Machine Tools (umati), OPC 30050 PackML, EUROMAP 77 injection-molding machines (OPC 40077), etc., released successively in recent years \\
knowledge & $\times$ & the four families of the base model: DataAccess / \textbf{Alarms \& Conditions (alarm logic = the promulgation of behavioral norms)} / HistoricalAccess / Programs \\
instances & tag values & live nodes in the server's address space \\
\bottomrule
\end{tabular}
\end{table}
\textbf{The point for this paper}: OPC UA's official literature describes itself as "OPC UA standardizes the data exchange with machines; Companion Specifications further standardize the data models of machine classes --- prescribing what data \emph{shall} be exchanged." The word "shall" is promulgation semantics. DA stopped at "can read, can write," its ecosystem locked to Windows; UA filled in the four layers and became the Industry 4.0 interoperability backbone. \emph{The gap between DA and UA is the gap of the two middle layers.}

\subsection{C.4 Modbus (1979--) --- the price list of a missing concept layer}
\textbf{Background}: created by Modicon for PLCs in 1979, to this day one of the largest-installed-base protocols in industrial fields. Its entire semantic apparatus: four classes of data tables (coils / discrete inputs / input registers / holding registers) + function codes.

\begin{table}[htbp]
\centering
\small
\begin{tabular}{@{}>{\raggedright\arraybackslash}p{0.473\textwidth}>{\raggedright\arraybackslash}p{0.473\textwidth}@{}}
\toprule
Layer & Status \\
\midrule
syntax & \checkmark{} PDU + function codes + four data-table classes \\
concepts & $\times$ register semantics drift into vendor PDFs: whether 40001 is a temperature or a pressure, the protocol itself does not say \\
knowledge & $\times$ none \\
instances & \checkmark{} real-time register values \\
\bottomrule
\end{tabular}
\end{table}
\textbf{The price list (each item identifiable)}:

\begin{enumerate}
\item the entire cost of semantic alignment is passed onto integrator labor --- the industry literature's consensus is that Modbus interoperability "still depends on manual gateway mapping to this day";
\item the mismatch between the data model's 1-based and the PDU's 0-based addressing is the industry's most common integration error, and some vendor documents do not mark which convention they use;
\item the high/low word order of 32-bit floats is set by each vendor for itself, and reading drift is a daily debugging routine;
\item no discovery, no metadata, no events: everything by polling.
\end{enumerate}
\textbf{The point for this paper}: Modbus is the pure specimen of "concept layer absent." It has not failed --- on the contrary, within the two layers \emph{syntax + instances} it is extremely successful (simple, robust, cheap); the position where it pays its price lands precisely on the layer it lacks: semantic alignment. Whichever layer is missing, the price falls on that layer.

\subsection{C.5 ODD (SAE J3016 / ISO 34503 / ASAM OpenODD) --- a record of the price of mixing layers}
\textbf{Background}: autonomous-driving safety standards require manufacturers to declare an Operational Design Domain (ODD). The main text has covered the engineering background; only the layering problem is registered here.

\textbf{Structural analysis}: existing ODD practice merges two kinds of content of different natures into a single "ODD declaration" carrier --- (a) the abstract classification of open-world scenarios (weather, road types, actors, \dots{}, continuously augmented as standard discussions proceed; \emph{open $\times$ abstract}); (b) the stable, concrete boundaries applicable to this vehicle model (under what conditions this vehicle may operate; \emph{stable $\times$ concrete}). Two goal pairs sharing one carrier directly violate carrier incompatibility (Section 5.3).

\textbf{The price}: declarations from different manufacturers are mutually incomparable, there is no common measure, and regulators cannot audit horizontally --- the price of mixing layers is paid continuously in the form of "declarations not comparable," unhealed to this day.

\textbf{The point for this paper}: ODD is the only witness in this appendix whose price is still being paid, and therefore the only case that permits a \emph{repair-direction prediction}: the four-layer framework happens to provide separate homes for the two kinds of merged content --- the abstract scenario classification moves up to the concept/knowledge layers; the concrete operating boundaries move down to the knowledge/instance layers.

\subsection{C.6 COIN (Yuan \& Yao) --- a complete four-layer inventory}
\textbf{Background}: Yuan and Yao's calculus of intelligence (COIN) formalizes agentic workflows as typed free-monad task spaces over a Grothendieck topos, with compositional certification and certificate anchoring guaranteeing their composability. It is the only purely mathematical construction among the witnesses (resting on no engineering archive at all) so its isomorphic arrival is the cleanest independent check of the layering lower bound. Section 2.4, R.9 has registered its theoretical position (an independent mathematical construction converging on the same structure); its complete four-layer inventory is archived here.

\textbf{The four-layer reverse reading}:

\begin{table}[htbp]
\centering
\small
\begin{tabular}{@{}>{\raggedright\arraybackslash}p{0.223\textwidth}>{\raggedright\arraybackslash}p{0.223\textwidth}>{\raggedright\arraybackslash}p{0.223\textwidth}>{\raggedright\arraybackslash}p{0.223\textwidth}@{}}
\toprule
Layer & COIN counterpart & Character & Argument \\
\midrule
syntax & the workflow signature + the monad composition laws (unit/bind and their laws) & stable $\times$ abstract & exists prior to all concrete tasks; prescribes "which constructs are composable and which laws composition must satisfy" --- the office of the syntax layer \\
concepts & the type constraints of the typed task space (type declarations such as UserID, AuthResult) & open $\times$ abstract & type declarations prescribe \textbf{which plans are legal} (plan admissibility), not what the world happens to contain: normative assertions, not factual ones; the type library can be augmented as the task domain expands \\
knowledge & per-task generated, compositionally certified and finalized decomposition trees and compiled blueprints, together with their certificate chains & stable $\times$ concrete & a blueprint records "which combinations of local solutions have been verified to work": stable (unchanged after finalization) and concrete (anchored to a specific task) \\
instances & real-world instances land indirectly through \textbf{leaf contracts + external implementations} & open $\times$ concrete & COIN's applicability gate (its \S{}2.1: applicable only after the relevant context has been made explicit) itself acknowledges the indirectness of the instance layer: a pure calculus does not touch the world directly; the world's access presupposes explicit contracts \\
\bottomrule
\end{tabular}
\end{table}
\textbf{Terminological precision note}: a bare Kleisli arrow is a morphism (a description of "how to compute from I to O with effects") not a static classification, and by itself instantiates no layer. What performs the office of the concept layer is the \emph{type discipline} constraining arrow admissibility; only when arrows are composed, certified, and finalized into blueprints does the stable $\times$ concrete layer emerge.

\textbf{Points for this paper}: (1) the four layers converge in a pure construction with no engineering archive, excluding the hypothesis that "four layers are just engineering convention"; (2) its \S{}2.1 applicability gate is, in this framework's terms, exactly the boundary statement "promulgation precedes applicability," providing the key citation site for Proposition A.21 (Appendix A.5); (3) blueprints = compiled plans, not trees --- finalized, not frozen (the COIN original allows blueprints to be re-certified after modification); both terms have been calibrated against the original text.

\subsection{C.7 Brick and Project Haystack (open-source community promulgation of building semantics) --- the concept layer supplied by industry self-organization}
\textbf{Background}: data integration in the building-automation industry long depended on manual point mapping. Project Haystack tags data points with metadata from a controlled tag vocabulary (governed by a community committee, versioned and maintained); Brick (Balaji et al., 2016, 2018) builds a class-hierarchy ontology on top of RDF, with tagsets (grouped packages of related tags) corresponding to classes, and introduces SHACL shape constraints for model validation; the Mortar open testbed accompanies them (real data from over 90 buildings (Fierro et al., 2020)). ASHRAE 223 (in preparation) is advancing this lineage into a formal standard.

\textbf{The four-layer reverse reading}:

\begin{table}[htbp]
\centering
\small
\begin{tabular}{@{}>{\raggedright\arraybackslash}p{0.306\textwidth}>{\raggedright\arraybackslash}p{0.306\textwidth}>{\raggedright\arraybackslash}p{0.306\textwidth}@{}}
\toprule
Layer & Counterpart & Promulgator \\
\midrule
syntax & RDF triples + tagset construction rules & W3C / the Brick community \\
concepts & the Brick class hierarchy and tagset controlled vocabulary; the Haystack tag vocabulary (a versioned vocabulary, only added, never altered) & the Brick community committee / Project Haystack \\
knowledge & SHACL validation rules: the shape constraints entities shall satisfy (e.g., which relations a sensor of a given class shall have), machine-checkable & the Brick community committee \\
instances & Brick graphs of concrete buildings (actual declarations of entities, relations, points) & deployment sites \\
\bottomrule
\end{tabular}
\end{table}
\textbf{Points for this paper}: (1) this is the \emph{positive version} of "whichever layer is missing, the price falls on that layer": the industry paid the manual cost of semantic alignment until it promulgated its own concept layer, and after promulgation the cost curve reversed; (2) the upgrade history from Haystack (pure tags) to Brick (tags + ontological constraints) shows: tags are flexible but unconstrained (the same class of sensor can be tagged differently in different buildings) and the community then used the ontology to pull the flexibility back into verifiable constraints --- the institutional form of the concept layer's openness by promulgation; (3) this is a promulgation-based semantic layer independently invented by an industry with no AI-theoretic agenda whatsoever, constituting out-of-domain confirmation of the criterion: building-data integration without a promulgated semantic layer is just one-off mapping engineering.

\subsection{C.8 RDF / OWL (the semantic-web lineage) --- the concept layer present, the knowledge layer absent}
\textbf{Background}: RDF (W3C, 1999/2004) gives the triple syntax and URI naming; OWL (W3C, 2004/2012) superimposes the TBox (classes, properties, and description-logic axioms) on top. The semantic web is the largest round of "framework first" experimentation at general Web scale.

\textbf{The four-layer reverse reading}:

\begin{table}[htbp]
\centering
\small
\begin{tabular}{@{}>{\raggedright\arraybackslash}p{0.306\textwidth}>{\raggedright\arraybackslash}p{0.306\textwidth}>{\raggedright\arraybackslash}p{0.306\textwidth}@{}}
\toprule
Layer & RDF & OWL \\
\midrule
syntax & \checkmark{} triple syntax + URI & \checkmark{} (inherited from RDF) \\
concepts & $\times$ (RDFS has only very weak class/property declarations) & \checkmark{} TBox: class hierarchies and axioms \\
knowledge & $\times$ & $\times$ the crucial gap: DL axioms are logical constraints on concepts, not promulgated norms of "how instances shall be and what deviation counts as"; there is no committee-promulgated industry operating procedure, and no institutionalized handling of violations \\
instances & \checkmark{} triples & \checkmark{} ABox \\
\bottomrule
\end{tabular}
\end{table}
\textbf{Points for this paper}: (1) the largest-scale demonstration that "the TBox is not the knowledge layer": the concept layer complete and the syntax layer solid, yet cross-domain auditable operational adjudication (does this reading of this device count as a violation) has never appeared on the semantic-web stack; the price of the knowledge layer's absence is paid continuously in the form "plenty of ontologies built, no one daring to sign an operational verdict"; (2) RDF itself is another specimen of two-layer "syntax + instances" success (same form as Modbus, scaled to the whole Web), and its price likewise falls on the missing layer; (3) the conceptual analysis "whether TBox/ABox are two of this paper's layers" is developed in Section 2.4, R.1 of the main text; only the empirical side is registered here.

\section{References}
\begin{reflist}
\item Alchourrón, C.E., Gärdenfors, P., Makinson, D., 1985. On the logic of theory change: Partial meet contraction and revision functions. Journal of Symbolic Logic 50(2), 510--530.
\item Angluin, D., 1980. Inductive inference of formal languages from positive data. Information and Control 45, 117--135.
\item ANSI/CTA, 2021. ANSI/CTA 709.6-A-2021: Control Networking Protocol Specification --- Part 6: Application Elements. Consumer Technology Association.
\item ASAM e.V., 2021. ASAM OpenODD Concept Paper, Version 1.0.
\item ASHRAE, 2016. ANSI/ASHRAE Standard 135-2016: BACnet---A Data Communication Protocol for Building Automation and Control Networks. American Society of Heating, Refrigerating and Air-Conditioning Engineers, Atlanta.
\item ASHRAE. Proposed Standard 223P: Semantic Data Model for Analytics and Automation Applications in Buildings (in preparation).
\item Balaji, B., Bhattacharya, A., Fierro, G., et al., 2016. Brick: Towards a unified metadata schema for buildings. In: Proceedings of BuildSys'16, pp. 41--50.
\item Balaji, B., Koh, J., Weibel, N., Agarwal, Y., 2018. Brick: Metadata schema for portable smart building applications. Applied Energy 226, 1273--1292.
\item CEN, 2014. EN 14908-6:2014: Open data communication in building automation, controls and building management --- Control network protocol --- Part 6: Application elements. European Committee for Standardization.
\item Chickering, D.M., 1996. Learning Bayesian networks is NP-complete. In: Fisher, D., Lenz, H.-J. (Eds.), Learning from Data: Artificial Intelligence and Statistics V. Springer, pp. 121--130.
\item Chisholm, R.M., 1963. Contrary-to-duty imperatives and deontic logic. Analysis 24, 33--36.
\item Clark, S., 2016. Internal debris may be causing problem with Mars rover's drill. Spaceflight Now, 2016-12-29.
\item Fierro, G., Pritoni, M., Abdelbaky, M., Lengyel, D., Leyden, J., Prakash, A., Gupta, P., Raftery, P., Peffer, T., Thomson, G., Culler, D.E., 2020. Mortar: An open testbed for portable building analytics. ACM Transactions on Sensor Networks 16(1), Article 7, 7:1--7:31.
\item Fraeman, A.A., Edgar, L.A., Rampe, E.B., et al., 2020. Evidence for a diagenetic origin of Vera Rubin ridge, Gale crater, Mars: Summary and synthesis of Curiosity's exploration campaign. Journal of Geophysical Research: Planets 125(12), e2020JE006527.
\item Gold, E.M., 1967. Language identification in the limit. Information and Control 10(5), 447--474.
\item IEC, 2010--2020. IEC 62541 series: OPC Unified Architecture. International Electrotechnical Commission, Geneva.
\item ISO, 2017. ISO 16484-5:2017: Building automation and control systems (BACS) --- Part 5: Data communication protocol. International Organization for Standardization, Geneva.
\item ISO, 2023. ISO 34503:2023: Road Vehicles --- Test Scenarios for Automated Driving Systems --- Specification for Operational Design Domain. International Organization for Standardization, Geneva.
\item Jørgensen, J., 1937/38. Imperatives and logic. Erkenntnis 7, 288--296.
\item Kinnett, R., Green, T., Klein, D., Richardson Lin, M., 2022. Remote diagnosis and operational response to an in-flight failure of the drill feed mechanism onboard the Mars Science Laboratory rover. In: Proceedings of the 46th Aerospace Mechanisms Symposium, NASA/CP-20220006415. NASA Technical Reports Server (NTRS 20220006415).
\item Lakdawalla, E., 2017. Curiosity's balky drill: The problem and solutions. The Planetary Society, 2017-09-06.
\item Le Cam, L., 1973. Convergence of estimates under dimensionality restrictions. Annals of Statistics 1(1), 38--53.
\item Makinson, D., van der Torre, L., 2000. Input/output logics. Journal of Philosophical Logic 29, 383--408.
\item Makinson, D., van der Torre, L., 2001. Constraints for input/output logics. Journal of Philosophical Logic 30, 155--185.
\item Modbus Organization. Modbus Application Protocol Specification (\allowbreak{}originated at Modicon, 1979; current specification maintained by the Modbus Organization).
\item NASA/JPL, 2016. Curiosity rover team examining new drill hiatus. NASA Jet Propulsion Laboratory press release, 2016-12-05.
\item OPC Foundation, 2020a. OPC 40001-1: OPC UA for Machinery, Part 1: Basic Building Blocks.
\item OPC Foundation, 2020b. OPC 40501-1: OPC UA for Machine Tools, Part 1: Monitoring and Job Overview (umati).
\item OPC Foundation / EUROMAP, 2020. OPC 40077: OPC UA for Plastics and Rubber Machinery --- Injection Moulding Machines to MES, Release 1.01 (the OPC UA mapping of EUROMAP 77).
\item OPC Foundation / OMAC, 2020. OPC 30050: OPC UA for PackML, Release 1.01.
\item Project Haystack. Semantic modelling for device and equipment data. \url{https://project-haystack.org.}
\item Ross, A., 1941. Imperatives and logic. Theoria 7, 53--71.
\item SAE International, 2021. Taxonomy and Definitions for Terms Related to Driving Automation Systems for On-Road Motor Vehicles, SAE J3016.
\item Vasavada, A.R., 2022. Mission overview and scientific contributions from the Mars Science Laboratory Curiosity rover after eight years of surface operations. Space Science Reviews 218, 14.
\item W3C, 2004. RDF Primer / RDF 1.1 Concepts and Abstract Syntax. W3C Recommendation.
\item Yuan, Y., Yao, A.C.-C., 2026. Calculus of intelligence: A topos-monadic framework for agentic workflows. iFuture, Online First, 9710001. doi:10.26599/\allowbreak{}IF.2026.9710001.
\end{reflist}